\documentclass[acmtog,nonacm]{acmart}
\AtBeginDocument{%
  }

\usepackage{subcaption}
\usepackage{algorithm}
\usepackage{algorithmic}
\usepackage{graphicx}
\usepackage{amsmath}
\usepackage{booktabs}
\usepackage{etoolbox}
\usepackage{graphicx,calc}
\usepackage{enumitem}
\usepackage{url}

\usepackage{tabularray}
\usepackage{nicematrix}
\usepackage{colortbl}
\usepackage{multirow}
\usepackage{float}
\usepackage{graphbox}
\usepackage{diagbox}
\usepackage{stackengine}
\usepackage{afterpage}

\begin{document}

\title{Style-based Clustering of Visual Artworks and the Play of Neural Style-Representations}

\author{Abhishek Dangeti}
\email{abhishek.dangeti@tcs.com}
\author{Pavan Gajula}
\email{pavanbhargav.gajula@tcs.com}
\author{Vivek Srivastava}
\email{srivastava.vivek2@tcs.com}
\author{Vikram Jamwal}
\email{vikram.jamwal@tcs.com}
\affiliation{%
  \institution{TCS Research}
  \country{INDIA}
}

\begin{abstract}

Clustering artworks based on style can have many potential real-world applications like art recommendations, style-based search and retrieval, and the study of artistic style evolution of an artist or in an artwork corpus. 
We introduce and deliberate over the notion of \textit{style-based clustering of visual artworks}. We argue that clustering artworks based on style is largely an unaddressed problem. 
We explore and devise different neural feature representations - from the style-classification, style-transfer to large language vision models -  that can be then used for style-based clustering. Our objective is to assess the relative effectiveness of these devised style-based clustering approaches through qualitative and quantitative analysis by applying them to multiple artwork corpora and  curated synthetically styled datasets. Besides providing a broad framework for style-based clustering and evaluation, our  analysis provides some key novel insights on feature representations, architectures and implications for style-based clustering.

\end{abstract}



\begin{teaserfigure}

\fbox{     

\includegraphics[width=\textwidth]{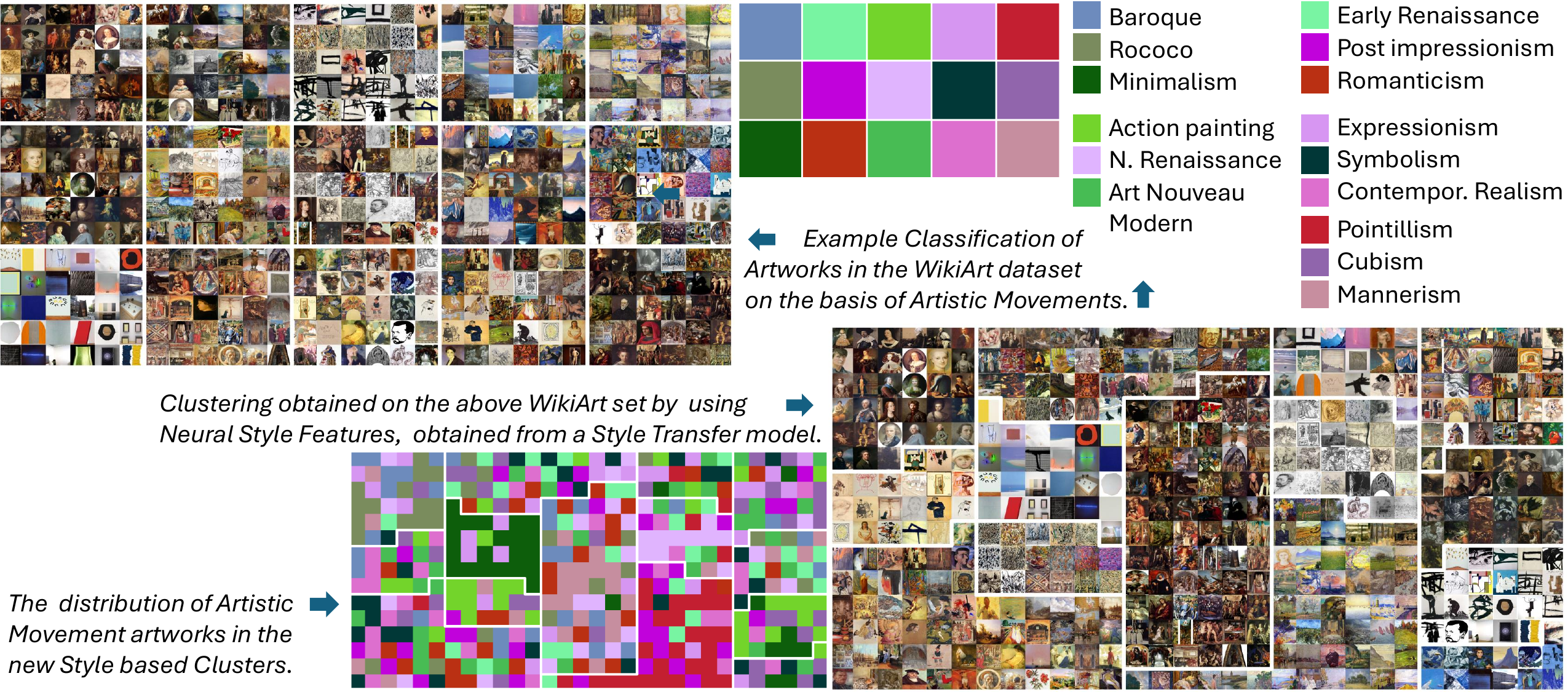}
}

\caption{\textbf{Style-based Clustering of Visual Artworks.} We present an artistic movement based classification of artworks in the WikiArt dataset. Alongside, we show the style-based clusters obtained with the neural style feature $F_{Styleshot}$. We show the distribution of artistic movement in the obtained clusters through color-coded representation. (Please zoom in for finer details.).} 
\label{fig: main_example}


\end{teaserfigure}

\maketitle

\section {Introduction}
\begin{figure*}[h!]

 \centering
\includegraphics[width = \textwidth]{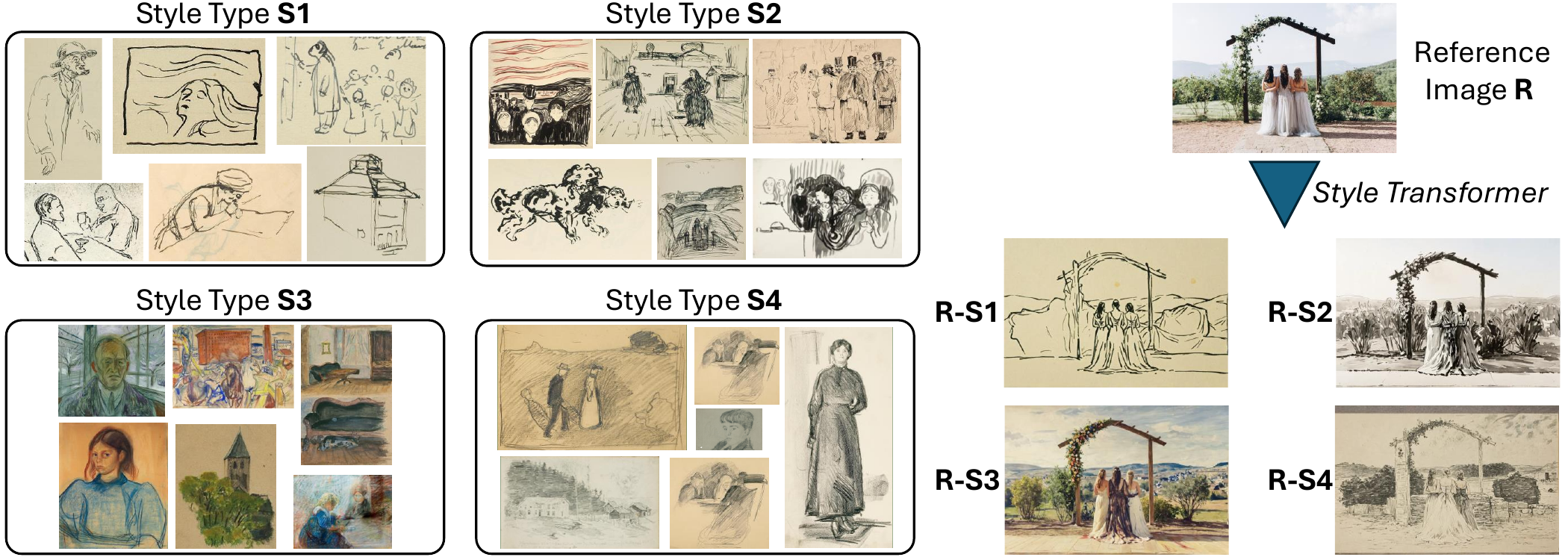}

  \caption{\textbf{Style palettes}: A few set of clusters (S1 to S4) obtained from the Edvard Munch Archive by leveraging $F_{StyleShot}$. On the right (R-S1 to R-S4), we show a few examples of style-transfer obtained using the artworks from the respective style palettes with a style-transfer model.}
  \label{fig:style-palette}
\end{figure*}
\textit{Style} of an artwork or an artist is an essential aspect of art as it has a bearing on the artist's identity, expression (emotional resonance, communication), cultural grounding, and aesthetics. The digitization of visual artworks through platforms such as WikiArt \cite{wikiart-web} and the Munch Museum's digital archive \cite{munch-web} renders them accessible to a global audience. It also facilitates a deeper study of various aspects (including the style) of these artworks through modern AI methods. 

 We argue that clustering artworks according to style, particularly through computer-assisted methods, is of significant importance and also presents some exciting opportunities. Both humans and machines can benefit from style knowledge through clustering.

   \textbf {1. Collection determines the style:}  The essence of style is defined through a collection. The style of an artist (or of an era) is rarely identified through a singular piece of artwork but is usually understood through a collection of an artist's (or an era's) works. Such clusterings are typically done by expert art curators, connoisseurs, and historians. Employing computer-assisted methods to create such clusterings can offer many benefits. They allow us to operate over larger collections and discover new relationships within them while deploying unsupervised or semi-supervised methods.

  \textbf{2. Discovering new styles}: Historically, the artworks produced by various artists have been categorized into (or labeled as) various visual art movements and styles such as Renaissance, Classicism, Cubism, Expressionism, Abstract, Baroque, Modern, Futurism, etc., based on the medium and philosophy of expression. However, after a careful study, one can observe a higher diversity of styles, sub-styles, or categories that can be more fine-granular than those covered by these art movements. Moreover, individual artists do not draw in a singular style; their art evolves depending upon their exposure, circumstances, and changing tastes. For example, in Figure \ref{fig:style-palette}, we observe examples of different artistic styles of Edvard Munch in his collection of sketches and watercolor paintings \cite{munch-web}.

  \textbf{3. Understanding Style Evolution or Building Style-based Narratives}: This can have numerous commercial applications. 
\textit{Style-based browsing} of an artwork archive can help you understand how the artworks from a collection are related to each other stylistically. This can lead to the discovery and explication of the evolution of style and different style-based structural relationships within an artist's work or across different periods of art development. 

  \textbf {4. Creating a Style Transfer Palette}: Various computer methods help to capture the style of an artist or a painting and then transfer it to a new piece of work. Even though the process of style transfer may raise many ethical questions, the development of these techniques can benefit the art industry. For example, an artist can monetize her style, or a museum can create interactive art experiences where people can experience or get a glimpse of how that artist would have drawn in her ear by observing the drawing through a style-trained robotic artist. An example development of \textit{Style-palette} for the artist Edvard Munch based on the collection of his sketch works and the different style-transfers performed on a reference image using the styles from the palette are shown in the figure \ref{fig:style-palette}.

 \textbf {5. Art Appreciation and Learning:}  The ability to explore a style through a cluster of similarly styled artworks provides a more nuanced view of the art style. It helps an aspiring artist to appreciate and learn the intricacies of style effectively by being exposed to multiple examples of a particular style.

Despite its importance, the field is not very well explored. 
Previously, several attempts have been made to analyze and understand artistic styles through methods such as style-based classification of artworks. A majority of the current artwork classification methods \cite{imran2023artistic,nunez2022classification,agarwal2015genre} classify artworks based on art movements as described above. Some methods such as \cite{aladin} have attempted to create more fine-granular artistic datasets through crowd-sourcing methods. However, due to the lack of publically available labeled fine-granular style-specific data, and the subjective nature of style,  the artistic style-based clustering of artworks is a largely unexplored task. Neural representations have tried to capture Style. Mostly applied to Sytle classification, Style Transfer, and to some extent Style-based retrieval. 
There has been no study that studies, how well these representations for work for Style Clustering. Alternatively, by using pattern recognition methods like unsupervised clustering, we can hope to identify artworks with similar style characteristics. This approach can aid in gaining a deeper understanding of an artist's various stylistic expressions and also allow for a more accessible analysis of artistic style evolution. Only a handful of studies, for example, \cite{castellano2020deep, Castellano_2022} have delved into unsupervised clustering of artworks. However, these approaches are generic and do not disentangle the content and style information to be useful specifically for style-level clustering of the artworks.

In this paper, we comprehensively study the problem of \textit{Style Based Clustering of Artworks}. Specifically, we ask the following questions:
 
 \begin{itemize}
      \item \textbf{RQ1:} Do we need specialized feature representation models for style-based clustering of visual artworks?
        \item \textbf{RQ2:} How effective are the respective neural style representations in achieving clustering?
        \item \textbf{RQ3:} What is the impact of underlying clustering architecture on style-based clustering?
        \item \textbf{RQ4:}Do we have neural representations for different style definitions? Does the same neural representation work well for various definitions of style?
        \item \textbf{RQ5:} What are the structural relationships in the styles present in an Art Corpus? Can style-based clustering help uncover it?
\end{itemize}

To explore these questions, we design an evaluation framework. 

        \begin{itemize}
            \item \textbf{Deep Learning Features:} We identify, pick up, or design(develop) neural features that are promising for style  capture from four different types of deep learning models (i) General Deep Learning Image Classification models (ii) Style-trained Image Classification models, (iii) Style Transfer Models, and  (iv)Large (Visio)Language Models 
            \item \textbf{Clustering Models:}We apply clustering on two representative architectures (i) K-Means, and (ii) DEC. The latter modifies the representations as it strives to build better clusters
            \item \textbf{Datasets:} We create datasets that can give us some concrete definition of Style.
            \item \textbf{Evaluation Metrics} We pick up metrics that we believe will give a good indication of how well a representation or an architecture performs the style-based clustering of artworks.
        \end{itemize}

Our paper contributes in the following manner:
 
\begin{enumerate}

 \item Ours is the first work to explicitly consider the problem of \textit{style-based clustering} in general, and applied to Artworks in particular.

 \item We present and provide solutions for the under-explored problem of artistic style-based clustering of artworks. 

  \item We devise and improvise various style-based features from varied sources such as artwork style classification, style transfer, and our custom style-trained network models, and use them as neural feature representations for artwork clustering.
 \item We devise new language-based style feature representations obtained through captioning and annotating large vision language models for artwork clustering.

 \item In the absence of style-based datasets, we curate datasets to create style-based ground truth clusters based on the base style images.
 
\item  We provide a novel evaluation framework to gauge the effectiveness of style-based clustering of artworks.
 \end{enumerate}

\begin{figure*}[ht!]
    \centering
    \begin{subfigure}[t]{0.5\textwidth}
        \centering
        \includegraphics[width=.9\columnwidth]{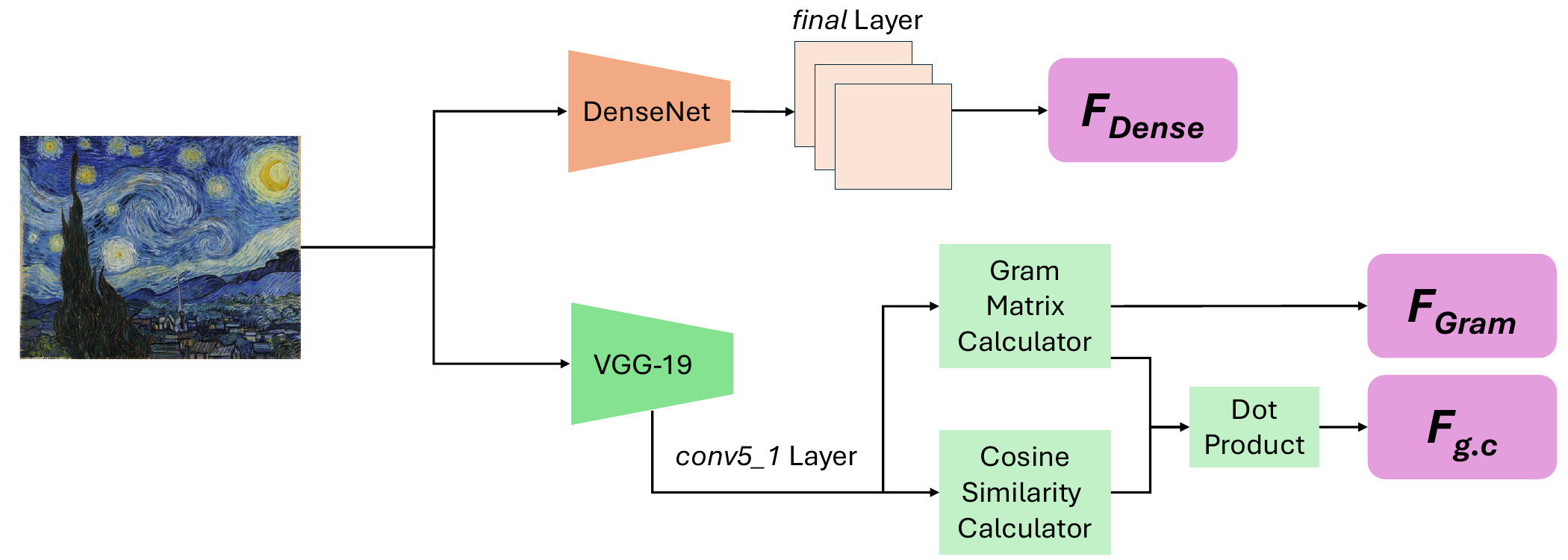}
        \caption{}
    \end{subfigure}%
    ~ 
    \begin{subfigure}[t]{0.5\textwidth}
        \centering
        \includegraphics[width=.9\columnwidth]{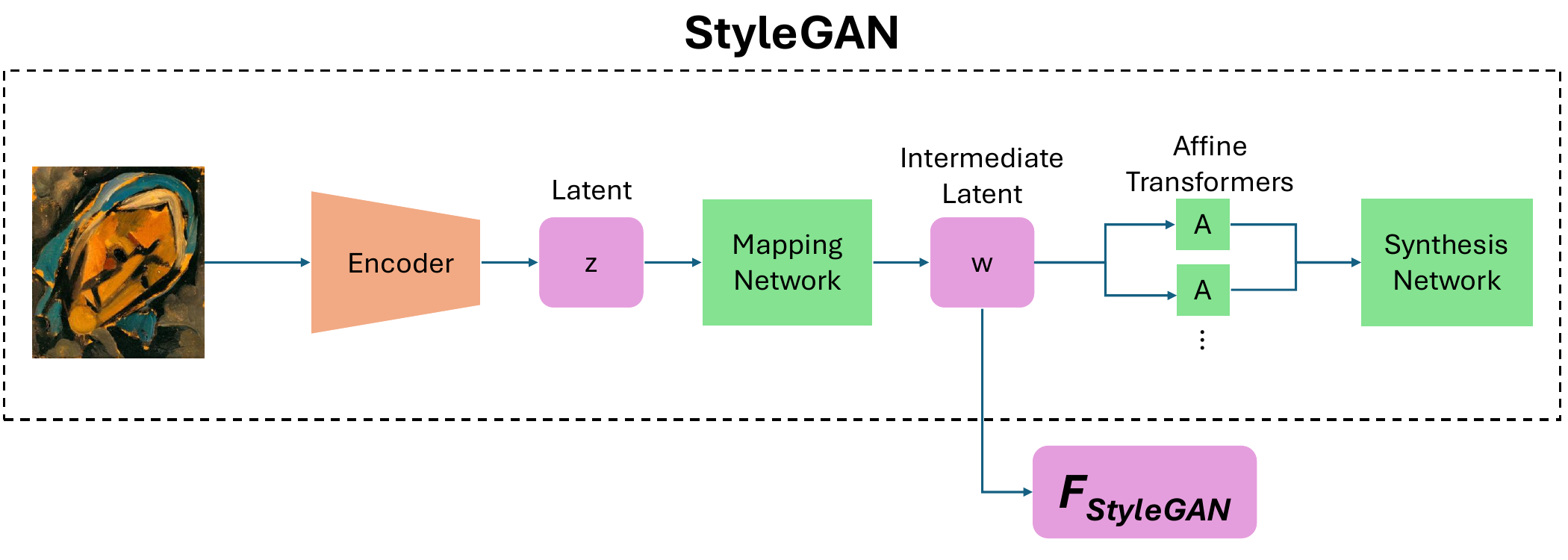}
        \caption{}
    \end{subfigure}
    \caption{Architecture for extracting two different artwork features: (a) $F_{CNN}$ and (b) $F_{StyleGAN}$.}
    \label{fig:feat_ext_nn}
\end{figure*}

\section{Related Work}

While ours is the first work to comprehensively study style-based clustering in artworks and the play of neural style representations, in this section, we discuss the related approaches in deep learning as applied to the \textbf{Style} in artworks.

\subsection{Style discovery and categorization}
Style discovery and categorization has been an active area of study with innovative approaches proposed for style-based classification, retrieval, etc. Several methods have explored artworks-classification based on handcrafted features such as color \cite{yelizaveta2005analysis} and brushstroke \cite{li2011rhythmic}. Works such as \cite{lombardi2005classification} consider various types of handcrafted features such as line, texture, color and light to achieve style classification of artworks. Recently, neural network-based architectures such as Convolution Neural Networks (CNNs) have been used to extract features from the artworks which are further used in artwork classification tasks \cite{cetinic2018fine}. Works such as \cite{deng2020exploring} try to learn the artist specific representations of the artworks to achieve artist-based classification of the artworks. Majority of the artwork classification models employ supervised learning and are limited to predicting the artist \cite{jangtjik2016artist,hua2020artist} or the popular art movement in the history \cite{el2021gcnboost}. These methods are not directly useful for identifying new or unknown artistic styles. Most of the artwork retrieval methods leverage content similarity to retrieve the artwork from a collection. For instance, in \cite{carneiro2012artistic}, monochromatic painting images are retrieved using a query consisting of a combination of classes or keywords, whereas \cite{castellano2021visual} fine-tunes a pre-trained CNN to retrieve paintings with similar artistic motifs given a textual user query. In contrast to retrieving artwork from textual queries, retrieving paintings from the given image(s) is also explored \cite{crowley2015face,shrivastava2011data, bam,img-style-emb}. However, none of these approaches present an in-depth study on style-based clustering of artworks as discussed in this paper.

\subsection{Style transfer} Stylized image generation through neural style-transfer techniques has been an active area of study \cite{li2017demystifying,shen2018neural,chen2017stylebank,virtusio2021neural}. Traditionally, these techniques relied on features extracted from pre-trained deep neural networks \cite{gatys2016image,jing2019neural} to merge content and style. Recently, techniques such as diffusion models and generative adversarial networks have facilitated models to generate high-quality images or transform given images in the style of artwork(s) or an artist \cite{yang2023diffusion,jing2019neural, hong2023aespa,sohn2023styledrop}. For instance, ZipLoRA \cite{shah2025ziplora} merges two individual LoRAs trained for style and content by learning mixing coefficients for their respective columns and generates stylized images using a text-conditioned diffusion model. Such style transfer techniques enable the artists to diversify their style palette which further underlines the importance of cataloging style through style-based clustering of artworks.

\subsection{Image and artwork clustering}
Over the years, several novel clustering algorithms have been proposed in computer vision spanning across supervised, semi-supervised, and unsupervised \cite{portela2014semi,van2020scan,chang2017deep,haeusser2019associative} approaches. \cite{manual-clustering} have explored manually clustering artworks based on visual similarity. They consider both content and style while clustering based on visual similarity. Recently, \cite{castellano2021visual, Castellano_2022} have explored the problem of unsupervised clustering of artworks leveraging the artwork features extracted with deep convolutional neural networks such as DenseNet \cite{huang2017densely}. However, these approaches do not specifically focus on style-based clustering. Furthermore, works such as \cite{GULTEPE201813} focuses on extracting features from artworks using K-Means feature learning \cite{UFLK} and utilize spectral clustering \cite{NIPS2001_801272ee} to cluster the artworks based on art movements. However, these works lacks in-depth discussion and exploration on the play of neural style representations in the context of style-based clustering of visual artworks.

\section{Style Feature Exploration}
\label{sec: feat_explo}

The essential first component in style-based clustering is the choice of neural features that can help us identify the style in an artwork.  We traverse the broad spectrum of style-specific notions in neural networks to extract and devise feature representations that can help us achieve style-based clustering of visual artworks. In general, we explore the following four different types of style-based neural features:

\begin{enumerate}
    \item \textbf{Convolutional Neural Network (CNN) based Style Features} wherein we explore CNN-based features traditionally used in image classification.
    \item \textbf{Style-transfer based Style Features } wherein we explore style features extracted from StyleGAN and other SOA style transfer techniques used for style-based image transformations.
    \item \textbf{Textual Style Features} wherein we propose using descriptive language-based style definitions extracted from artworks using vision language models(VLM).
    \item \textbf{Special Style-trained Image Model based Features } in which we extract features from image models that are specially trained for style classification.
\end{enumerate}
 
\subsection{Convolutional Neural Network based Style Features ($F_{CNN}$)} 
\label{sec:nn_feat}

 Convolutional Neural networks (CNN) have been used in previous works to extract style-based features for various style-based tasks such as style classification \cite{stclass16,stclass18}, style-transfer \cite{gatysnst}, style-based retrieval \cite{cnn-style-retrieval}. We examine two types of $F_{CNN}$ features for clustering based on style.

\subsubsection{DenseNet Features ($F_{Dense}$)}:
Some previous approaches for clustering visual artworks \cite{Castellano_2022} utilize the last layer of DenseNet \cite{huang2018densely} to extract features from the artworks and incorporate them into the DEC model.  The last layer would contain rich information about artwork as each layer of the DenseNet is connected to all its previous layers. 
In our exploration, we utilize $F_{Dense}$ features to check whether these features can also be effective in style-based clustering for visual artworks. Similar to \cite{Castellano_2022}, after obtaining the features for each artwork in the dimensions 1024$\times$7$\times$7, we apply global average pooling (GAP) \cite{gpa} to obtain 1024-feature vector for each artwork.

\subsubsection{Gram Matrices based Style Features ($F_{Gram}$ and $F_{g \cdot c}$)}: Works such as \cite{stclass16,stclass18} utilize the style features introduced by \cite{gatysnst} for style based classification of artworks. They explore different combinations of features extracted from different layers of CNN, as well as different mathematical correlations and combinations of these features for style classification. They observe that the gram matrices of features obtained from $conv5\_1$ ($F_{Gram}$) yield one of the best results in the classification task. They also observe that the dot product of $F_{gram}$ and cosine similarity of the features extracted from $conv5\_1$ ($F_{g \cdot c}$) also produces good results. In our work, we use $F_{Gram}$ and $F_{g \cdot c}$ and explore their impact on unsupervised style-based clustering. We reduce the dimensions of $F_{Gram}$ and $F_{g \cdot c}$ from 512$\times$512 to 512 for each artwork using GAP. The architecture for extracting $F_{CNN}$ features can be seen in Figure \ref{fig:feat_ext_nn} (a).

\subsection{Style-transfer based Style Features ($F_{ST}$)} 
\label{sec:st_feat}

\begin{figure}[!t]
  \begin{subfigure}[t]{\columnwidth}
    \centering
    \includegraphics[width=\textwidth]{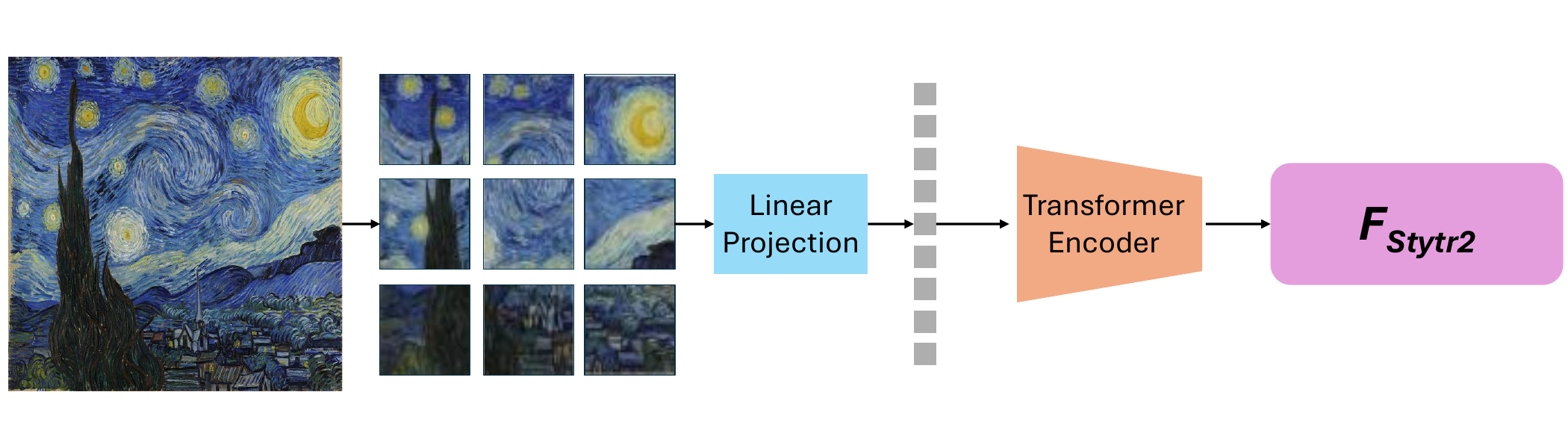}
    \caption{Stytr$^2$}
  \end{subfigure}
   \begin{subfigure}[t]{\columnwidth}
    \centering
    \includegraphics[width=\textwidth]{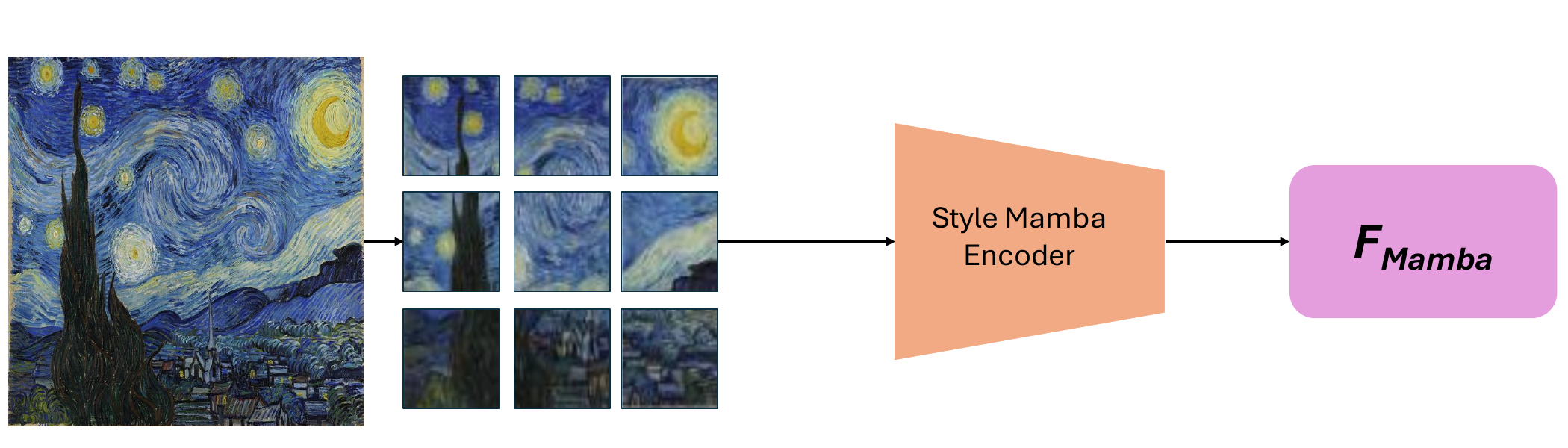}
    \caption{Mamba-ST}
  \end{subfigure}

  \begin{subfigure}[t]{\columnwidth}
    \centering
    \includegraphics[width=\textwidth]{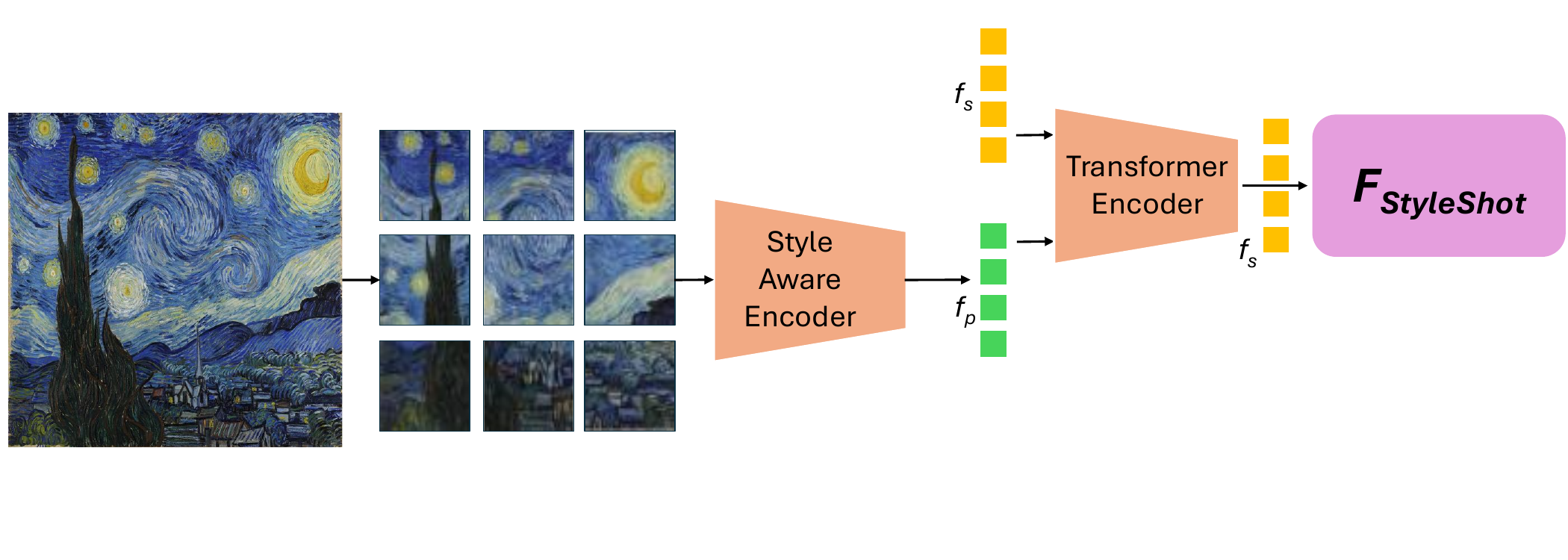}
    \caption{Styleshot}
  \end{subfigure}
  \caption{Architecture for extracting different $F_{st}$ features.}
  \label{fig:st_arc}
\end{figure}

Style-transfer is a heavily explored problem where a style-transfer model transfers the style of a style reference image to a content image. Most style-transfer approaches encode the style reference image to obtain style features which are then used by the model to transfer the style to a content image. In this subsection, we explore different state-of-the-art style-transfer approaches to identify the process used to extract the style features $F_{ST}$ from a style reference image. The architecture for extracting different $F_{ST}$ features can seen in Figure \ref{fig:st_arc}.    

\subsubsection{StyleGAN based Style Features ($F_{StyleGAN}$)}
\label{sec:stylegan_feat}
Features from different levels of the generator in StyleGAN\cite{stylegan} help in learning  different types of style information present in the images.
The GAN \cite{GAN} architecture is modified to disentangle style from content by starting from a constant vector ($z$) rather than the encoded latent vector and passing $z$ through a non-linear mapping network to obtain a intermediate latent $w$. We consider this intermediate latent as $F_{StyleGAN}$.

\subsubsection{Stytr$^2$ based style features ($F_{Stytr2}$)}
\label{sec: stytr2}
In the Stytr$^2$ \cite{stytr2} approach, the style reference image is split into patches. These patches are then passed through a linear projection layer to obtain a sequential feature embeddings. The sequential feature embeddings are then passed through a transformer encoder which consists of a multi-head self-attention block and a feed forward network. After passing it through the transformer encoder, the output features extracted from the encoder represent the style information present in an artwork. We term these style representations as  $F_{Stytr2}$ features.

\subsubsection{Mamba based style features ($F_{Mamba}$)}
\label{sec: mamba}
In the Mamba-ST \cite{mamba-st} approach, similar to Stytr2, both the style reference image and the content image are split into patches
and each patch is projected into a 1D embedding  using a patch embedding layer. These patch embeddings are then normalized and passed through the domain-specific(content and style) Mamba encoders, followed by an ST-Mamba decoder. When the model is trained using the perceptual and identity losses, the mamba encoders containing base visual state-space machines (VSSMs) learn the domain specific representations while the ST-Mamba decoder with the help of ST-VSSM learns to fuse the style and content information performing the style-transfer. We leverage the representations from the style mamba encoder since they encompass the style information present in artworks and term them as $F_{Mamba}$.

\subsubsection{Styleshot based style features ($F_{Styleshot}$)}
\label{sec:styleshot}
In the Styleshot \cite{styleshot} approach, similar to the previous approaches, the style reference image is split into patches. Unlike the previous approaches, the image is split into multi-scale patches (1/4, 1/8, and 1/16 of an image). For each scale, a distinct ResBlock is utilized to obtain the patch embeddings $f_p$ at each scale. To integrate these multi-level style embeddings, a learnable style embedding $f_s$ is concatenated with the multi-scale embedding ($f_p$) and the combined embedding is fed into a standard transformer. The learnable style embedding ($f_s$) is then extracted from the output of the transformer to obtain the rich style embedding which we term $F_{Styleshot}$. 

\begin{figure}[t]
    \centering
\begin{subfigure}[t]{\columnwidth}
    \centering
    \includegraphics[width=\textwidth]{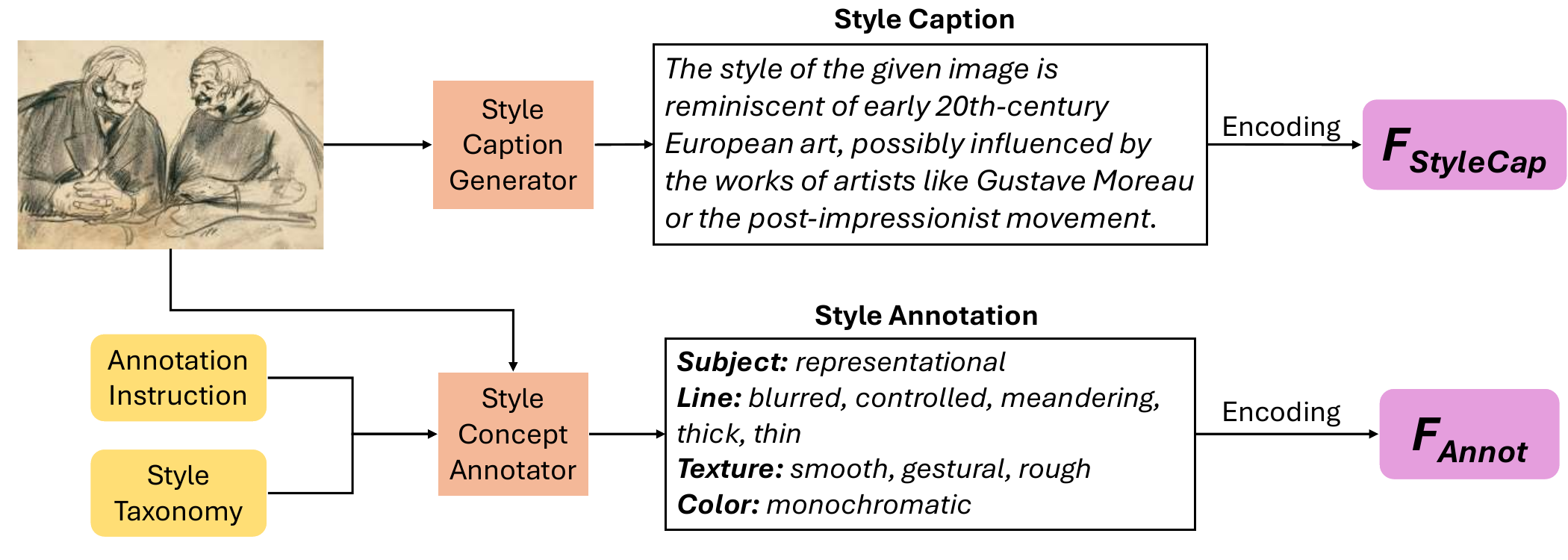}
    \caption{}
\end{subfigure}
\begin{subfigure}[b]{\columnwidth}
    \centering
    \resizebox{\textwidth}{!}{
    \begin{tabular}{l|l}
\textbf{Visual Elements} & \textbf{Concepts} \\ \hline

Subject & Representational, Non-representational \\ \hline
Line & \begin{tabular}[c]{@{}l@{}}Blurred, Broken, Controlled, Curved, Diagonal,\\ Horizontal, Vertical, Meandering, Thick, Thin, \\Active, Energetic, Straight\end{tabular} \\ \hline
Texture & Bumpy, Flat, Smooth, Gestural, Rough \\ \hline
 Color & \begin{tabular}[c]{@{}l@{}}
 Calm, Cool, Chromatic, Monochromatic, Muted,\\ Warm, Transparent \end{tabular} \\ \hline
Shape & \begin{tabular}[c]{@{}l@{}}Ambiguous, Geometric, Amorphous, Biomorphic,\\ Closed, Open, Distorted, Heavy, Linear, Organic, \\Abstract, Decorative, Kinetic, Light\end{tabular} \\ \hline
Light and Space & \begin{tabular}[c]{@{}l@{}}Bright, Dark, Medium, Atmospheric, Planar,\\ Perspective\end{tabular} \\ \hline
\begin{tabular}[c]{@{}l@{}}General Principles \\of Art \end{tabular} & \begin{tabular}[c]{@{}l@{}}Overlapping, Balance, Contrast, Harmony, Pattern,\\ Repetition, Rhythm, Unity, Variety, Symmetry,\\ Proportion, Parallel\end{tabular}
\end{tabular}}
    \caption{}
\end{subfigure}
    \caption{Architecture for the process of extracting (a) $F_{Text}$ features and the (b) style taxonomy used in extracting $F_{Annot}$ representations.}
    \label{fig:feat_ext_txt}
\end{figure}
\subsection{Textual Style Features ($F_{Text}$)}
\label{sec:text_feat}

The style representations presented thus far capture the style of an artwork in a interpretation which is difficult for a human to interpret in a meaningful way. Providing the style some sort of interpretability which is easily understood by a human could provide a meaningful way in studying the style aspects of an artwork. Expressing the style of an artwork through a textual medium allows a user to easily interpret the style of an artwork and find the correlation between the clustered artworks. To this end, we propose two types of textual style features ($F_{Text}$): $F_{StyleCap}$ and $F_{Annot}$.
The architecture for extracting $F_{Text}$ features can be seen in Figure \ref{fig:feat_ext_txt}. 

\subsubsection{Artwork Style Caption Representation ($F_{StyleCap}$)}
\label{sec: caption}
In this approach, we generate the artwork representation using the style caption of the artwork. The style information in the style caption describes the style aspects present in an artwork. Formally, we obtain $F_{StyleCap}$ for an artwork $a_i$ as:

\begin{equation}
    F_{StyleCap} = T(C(a_i, instruction))
\end{equation}

where $C$ is the style caption generator and $T$ is the text encoder to encode the caption for the artwork $a_i$.
In this work, we leverage an open-source multi-modal large language model(MLLM) called InternVL 2 \cite{internvl}. It achieves the state-of-the-art performances in most of the validation benchmarks competing with both closed source proprietary models and other open-source models. It comes in 5 variants and we use the smallest 2 billion parameter model in our work. For the vision part, the 2B variant uses InternViT model, while for the language part it uses Internlm2-chat-1 model \cite{internvl}. These models support multiple different modalities like image, text, video, etc., and can handle different outputs such as images, bounding boxes, masks, etc. thereby providing multitask functionality. The InternVL 2 model takes the instruction and the artwork as input and gives us an output text that describes the style of the image.

Next, we use Long-CLIP \cite{longclip} as the text encoder ($T$) for artwork captions. We then use these artwork representation $F_{StyleCap}$ as input artwork features in the DEC model. It is to be noted that we use InternVL and Long-CLIP as a proxy for $C$ and $T$ respectively and it could be replaced with other image captioning and text encoder models.

\subsubsection{Artwork Style Concept Annotation Representation ($F_{Annot}$)}
\label{sec: annotation}
In this approach, we annotate artworks with style concepts based on the fundamental principles of art \cite{ocvirk1968art}. A set of 59 different concepts across seven visual elements has been utilized by \cite{8334453} in their work (see Figure \ref{fig:feat_ext_txt} (b)). Formally, we obtain the artwork representation for an artwork $a_i$ with style concept annotation as:
\begin{equation}
    F_{Annot} = T(S(a_i, taxonomy, instruction)
\end{equation}
where, $S$ is the style concept annotator and $T$ is the one-hot encoder to encode the style concepts of an artwork into a one-hot vector. The style concept annotator considers the \textit{taxonomy} given in Figure \ref{fig:feat_ext_txt} (c) and the \textit{instruction} is to associate the style concepts (for each visual elements) from the taxonomy to a given artwork $a_i$. The instruction is constructed in a manner where the instruction includes a query for each style attribute. 
Similar to $F_{StyleCap}$, we leverage the InternVL 2 as the style concept annotator. After obtaining the style concepts, we turn the 59 style concepts into a one-hot vector based on if a style concept is present in the artwork or not.
The style information available with this method is fine-grained across various artistic style dimensions. It is to be noted that we use InternVL 2 as a proxy and it could be replaced with other style concept annotators.

\begin{figure}[t]
    \centering
    \includegraphics[width=\columnwidth]{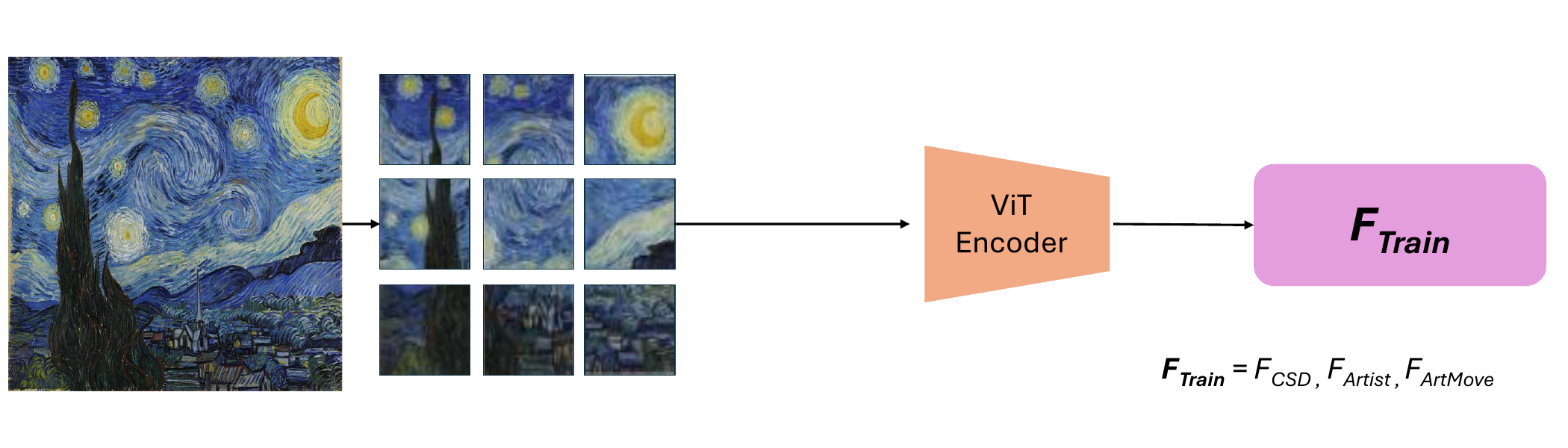}
  
  \caption{Architecture for extracting different $F_{Train}$ features.}
  \label{fig:train_arc}
\end{figure}
\subsection{Special Style-trained Image Model based Features ($F_{Train}$)}
\label{sec:artist_feat}
Here we explore the features extracted from the models that are trained on the datasets with specific style definitions like artist attribution, wikiart's artist based and art movement based definitions, etc. We also look at a model trained on generic Imagenet based definition.

\subsubsection{Vision Transformer Model trained on ImageNet ($F_{ViT-ImageNet}$)}: In this work, the Vision Transformer (ViT) models \cite{vit-imagenet} are trained for object-based image classification in a completely supervised manner. The dataset used for training is the ImageNet dataset \cite{imagenet-dataset}. We term the features obtained through this model as $F_{ViT-ImageNet}$.

\subsubsection{Contrastive Style Descriptors ($F_{CSD}$)} : In this work \cite{csd}, the Vision Transformer(ViT) models(base and large variants) are trained on a multi-label contrastive objective to learn style information from artworks that adheres to artist attributes. The dataset used for training is curated from LIAON Aesthetics by selecting and filtering images according to a predefined set of style tags. The style tags are obtained by combining the bank of artists, mediums, and movement references used on typical user prompts for Stable Diffusion. Each image in the curated dataset can have more than one tag with each tag representing a style attribute. Once the model is trained on this dataset, the features from the last layer of ViT backbone are extracted and used as $F_{CSD}$.

\subsubsection{Artwork-trained Image Model based features}:
In \cite{stclass16,stclass18}, we observe that training the models on a classification task also make their features robust for clustering. Similarly in the case of CSD \cite{csd}, we notice that the training dataset used, LAION Aesthetics, contains wikiart data as a subset and the respective features from the pre-trained model yield higher results in the clustering task than the rest of the features, as evident from the metric scores. To this end, we venture into this direction by fine-tuning a ViT model on the wikiart data considering two different ground truth labelings: Art movement based and Artist based.
\begin{enumerate}

\item {$F_{Artist}$}:
 For artist based ground truth dataset, we sort the artists based on the number of artworks they produced in the descending order and select the artworks from the top 40 artists. We do this to maintain class-balance and ensure sufficient number of samples per each artist class. The total artworks obtained are 25550 which accounts for 32\% of the whole wikiart dataset. Out of this, we use 85\% of the data for training and the remaining for testing. 
 
\item {$F_{ArtMove}$}:
Similarly for the art movement based groundtruth dataset, we sample the same number of artworks(20887 artworks) for training set as we did for the artist based data. We use the existing wikiart subset for the test set. 

Using these two different groundtruths, we fine-tune two separate models that are pre-trained on Imagenet-21k each for 45 epochs with the cross-entropy loss. We extract the features from the last layers of the fine-tuned ViT models and use them for clustering.
\end{enumerate}

\section{Datasets}
\label{sec: experiments}
\begin{figure}[!th]
 \begin{subfigure}[t]{0.42\textwidth}
    \centering
    \includegraphics[width=\textwidth]{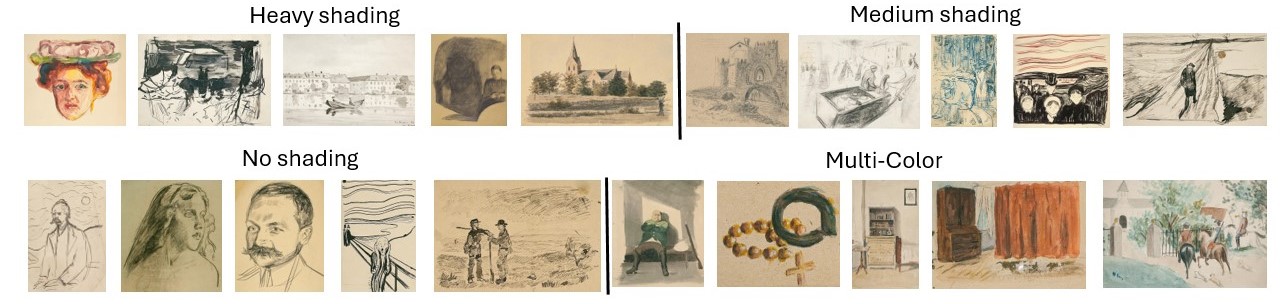}
    \caption {Edvard Munch archive}

  \end{subfigure}
   \begin{subfigure}[t]{0.42\textwidth}
    \centering
    \includegraphics[width=\textwidth]{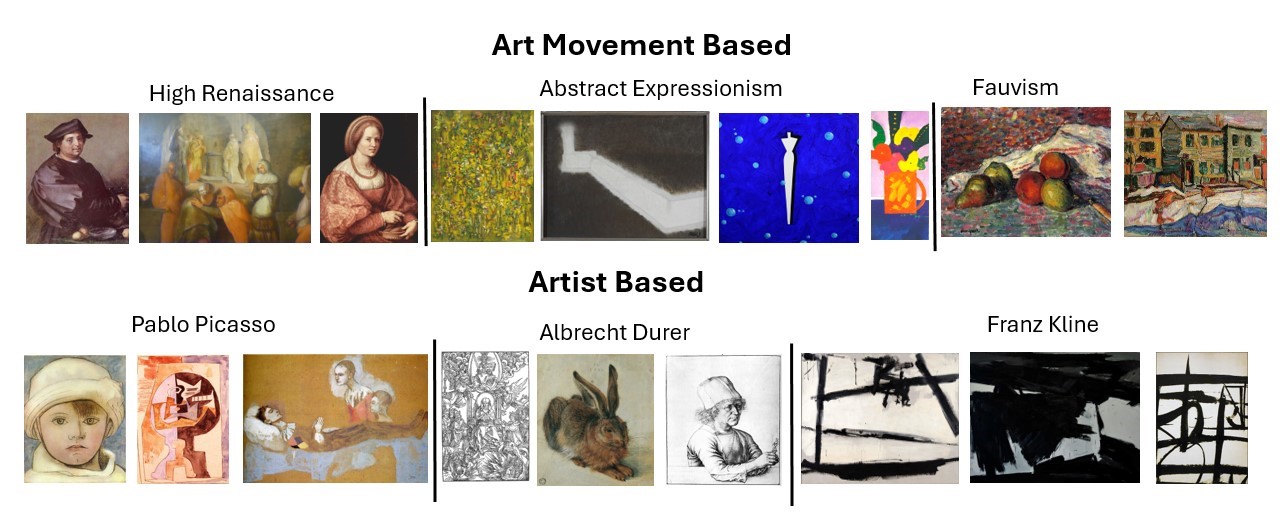}
    \caption {WikiArt dataset}

  \end{subfigure}
   \begin{subfigure}[t]{0.35\textwidth}
    \centering
    \includegraphics[width=\textwidth]{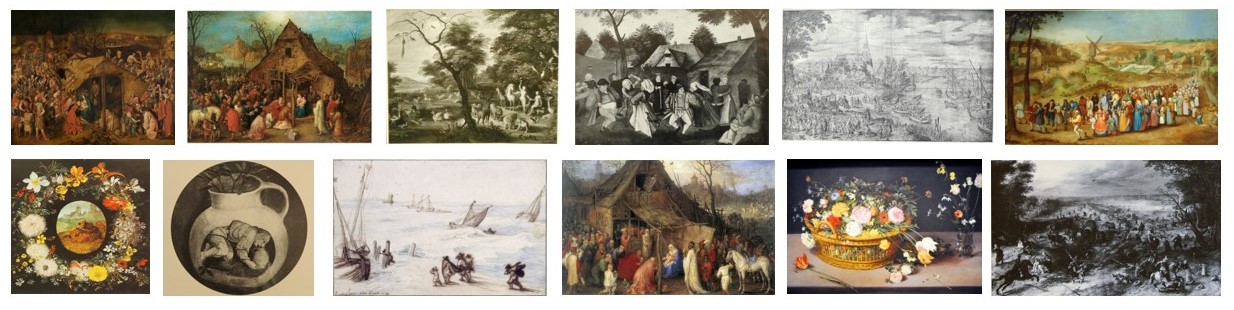}
    \caption {Brueghel dataset}

  \end{subfigure}
   \begin{subfigure}[t]{0.42\textwidth}
    \centering
    \includegraphics[width=\textwidth]{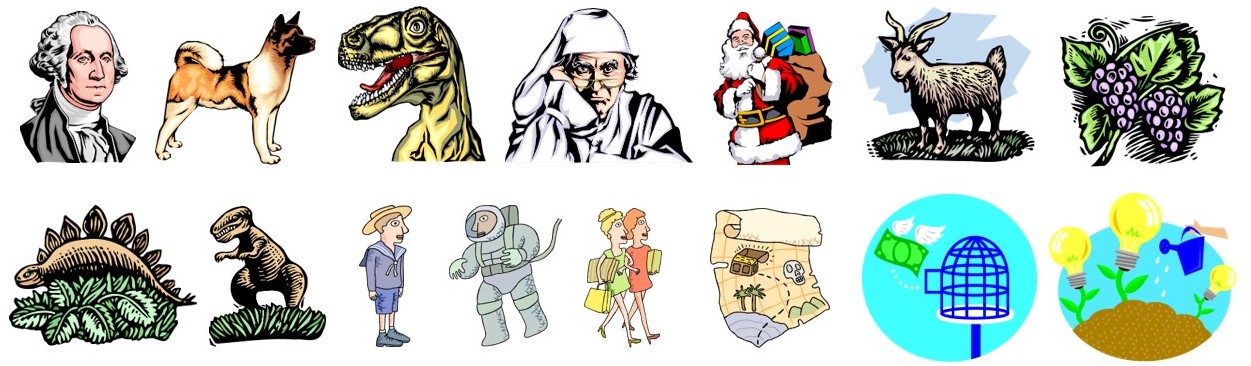}
    \caption {Clip Art Illustrations dataset}

  \end{subfigure}
   \begin{subfigure}[t]{0.42\textwidth}
    \centering
    \includegraphics[width=\textwidth]{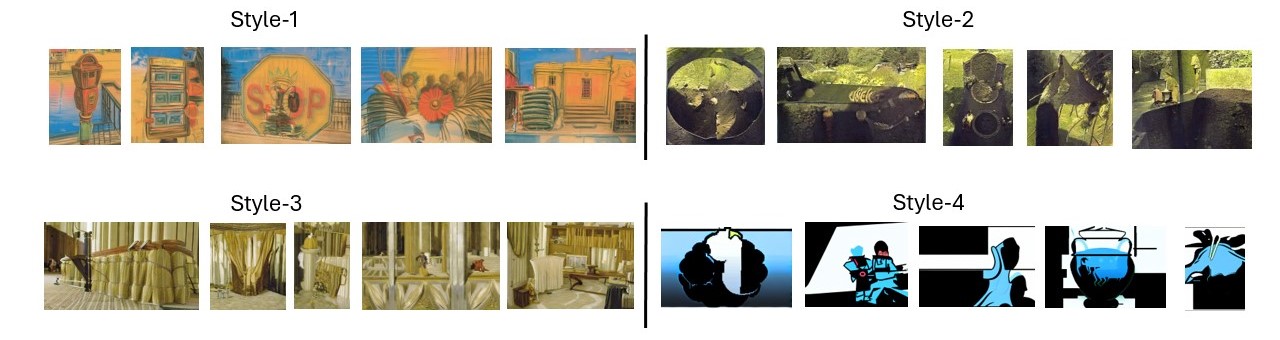}
    \caption {Mixed curated styleshot Dataset}

  \end{subfigure}
   \begin{subfigure}[t]{0.42\textwidth}
    \centering
    \includegraphics[width=\textwidth]{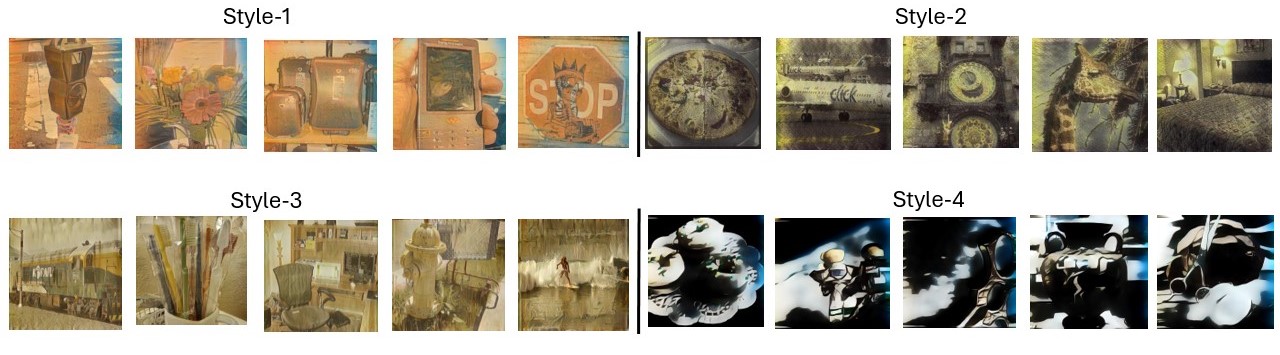}
    \caption {Mixed curated Mamba dataset}

  \end{subfigure}
  
   \begin{subfigure}[t]{0.42\textwidth}
    \centering
    \includegraphics[width=\textwidth]{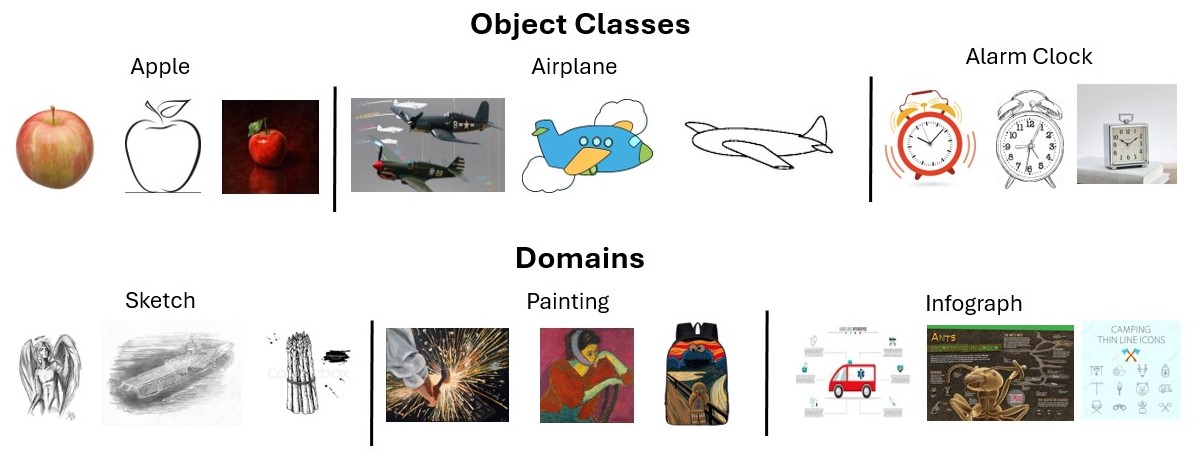}
    \caption {DomainNet dataset}

  \end{subfigure}
   \caption{Representative samples from the different datasets. (Please zoom in for finer details).}
  \label{fig:dataset_example}
\end{figure}
\begin{figure*}[htp!]
    \centering
    \includegraphics[width=\textwidth]{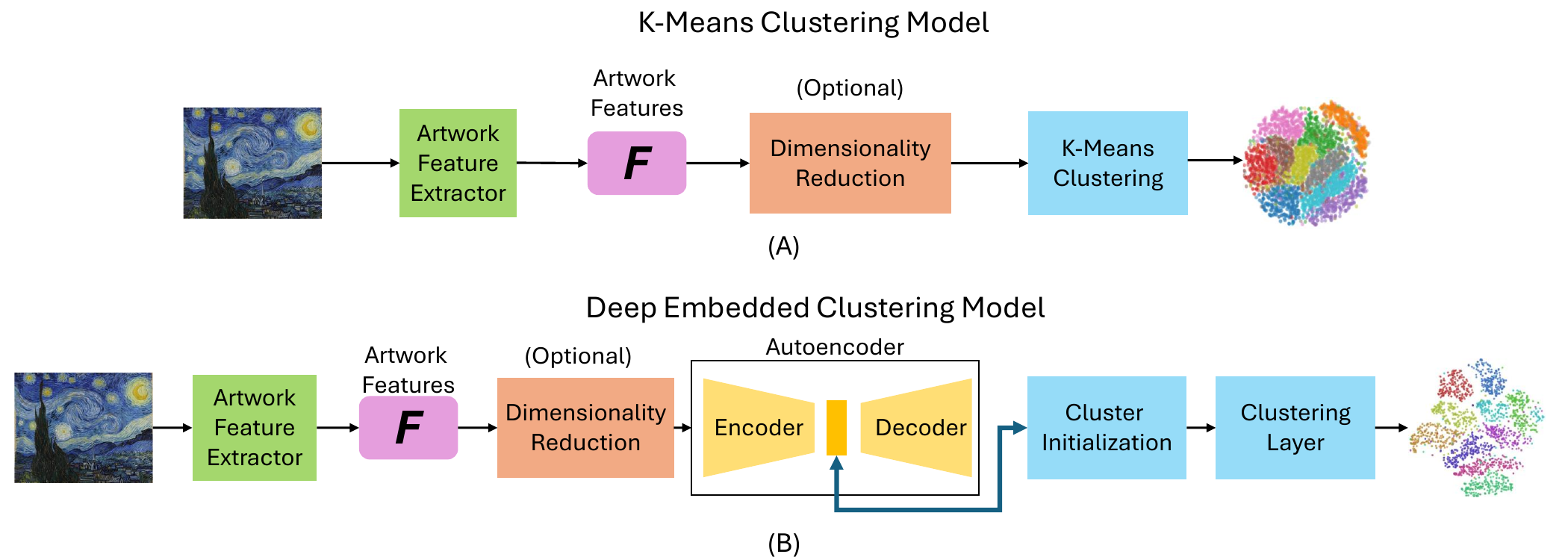}
    \caption{The architecture for our approach with the (A) K-Means clustering model \cite{kmeans:macqueen1967some} and the (B) Deep embedded clustering model (DEC) \cite{xie2016unsupervised}. For the K-Means clustering model, the dimensionality reduced artwork features are directly used for clustering, whereas for the DEC model, the features are first projected into the latent space before being used for clustering. Please refer to Section \ref{sec:dim_red} for details of our dimensionality reduction methods.
    }
    \label{fig:DEC}
\end{figure*}

In our experiments, we leverage four stylistically diverse artwork datasets as our \textbf{core} datasets. In Figure \ref{fig:dataset_example}, we present the representative samples from each dataset. Furthermore, in Sections \ref{sec: wikiart}, \ref{sec: synthetic_dataset}, and \ref{sec: domainnet}, we discuss the different configurations of the datasets used in the experiments. The four \textit{core} datasets are:

\begin{enumerate} 

    \item \textbf{Edvard Munch Archive (EMA)}: We experiment with the artwork collection dedicated to the artist Edvard Munch. We specifically consider sketch and watercolor paintings, comprising 7410 artworks created by Edvard Munch \cite{newsnow}. The artworks are categorized based on shading and color.

    \item \textbf{WikiArt Dataset (WAD)}: WikiArt is the largest collection of digitized artworks encompassing artists from several art movements. Similar to \cite{Castellano_2022}, in this work, we use the WikiArt dataset created by \cite{tan2016ceci}, which contains 78,978 artworks. The artworks are categorized based on 27 art movements.
    
    \item \textbf{Brueghel Dataset (BD)}: The Brueghel dataset \cite{brueghel-dataset} consists of 1587 artworks created by Jan Brueghel The Elder. This dataset consists of artworks in different media like oil, ink, and watercolor, along with various painting surface materials such as paper, panel, and copper. 
    
    \item \textbf{Clip Art Illustrations Dataset (CAID)}: Clip art images consist of various styles such as sketches, woodcuts, cartoons, and gradient-shading. We adopt the clip art illustrations dataset used in \cite{clip-art-dataset} consisting of 4591 clip art illustrations. 1000 of the illustrations have been collected from the Art Explosions dataset \cite{art-explo}, and 3591 of those illustrations are from the clip art included in Microsoft Office.

\end{enumerate}

\subsection{WikiArt Dataset Variants} 
\label{sec: wikiart}
As mentioned earlier, the WikiArt dataset contains 78,978 artworks with artworks from 27 different art movements (AM). We perform our experiments on three variations of the WikiArt dataset.
 
 \begin{enumerate}
     \item \textbf{WikiArt-79k}: We use the entire WikiArt dataset with 78,978 artworks with artworks from 27 different art movements. This variation of the dataset is mainly used for testing different features on the art movement based clustering and artist based clustering.
     \item \textbf{WikiArt-25k-Artist}: We pick artworks from the top 40 present in the full WikiArt-79k dataset. We obtain around 25,550 artworks. We utilize this variation to test artist-based clustering.
     \item \textbf{WikiArt-21k-Artist}: We pick the WikiArt-25k-Artist subset and remove 4000 artworks to be used as a test set (WikiArt-4k-Artist). We obtain around 21,550 artworks. We utilize this variation for training a classification model on the artist style definition to obtain $F_{Artist}$ features.
     \item \textbf{WikiArt-21k-AM}: We pick 20887 artworks from the 27 art movements present in the full WikiArt-79k dataset. We obtain around 20,887 artworks and utilize this variation for training a classification model on the art movement based style definition to obtain $F_{ArtMove}$ features.
     \item \textbf{WikiArt-4k-AM}: We pick 4050 artworks from the WikiArt dataset by picking 150 artworks from each of the 27 art movements. We use this variation to investigate the structural relationship between the artworks present in the dataset as well as to test the features including $F_{ArtMove}$ on art movement based clustering.
     \item \textbf{WikiArt-4k-Artist}: We pick 4000 artworks from the WikiArt dataset by picking 100 artworks from each of the top 40 artists. We use this variation to test the features including $F_{Artist}$ features for artist based clustering. 
 \end{enumerate}

\subsection{Synthetically Curated Dataset} 
\label{sec: synthetic_dataset}
There is a lack of dedicated evaluation metrics and ground truth datasets for style-based clustering. Even though the publicly available datasets like WikiArt might seem useful for validating cluster style quality, a quick inspection of the data reveals that there is a huge overlap in the styles and that the labelling of styles in these datasets doesn't align well with human perception of style rendering them ineffective for cluster style validation. We touch upon this further in the results section. To this extent, we develop a synthetically curated dataset for style-based clustering of artworks. We select 10 stylistically distinct style reference images from each of our four core datasets and 100 distinct content images from the MS COCO dataset \cite{mscoco}. For each dataset, we use the 10 style reference images and style-transform the content images through an image-to-image style-transfer model Styleshot \cite{styleshot} and Mamba-ST \cite{mamba-st} to obtain 1000 stylized images (please refer to the Section \ref{sec: additional_results} for a sample of style transformations). For ground truth, we treat the images stylized from the same style reference image as part of the same style cluster. Thus, we obtain 10 different style clusters, each having 100 images of the same style. 

We consider a combination of the 4 curated datasets which contains 40 different style clusters with 4000 images. We obtain two Mixed curated datasets, one dataset obtained through Styleshot and another dataset obtained through Mamba-ST. We term these two dataset, \textit{Mixed styleshot curated dataset (MSC-4k)} and \textit{Mixed mamba curated dataset (MMC-4k)}.

\subsection{DomainNet Dataset}
\label{sec: domainnet}
The DomainNet \cite{domainnet} dataset contains 0.6 million images from six domains and 345 different classes. The six domains include domains such as clip-art, infograph, real-life images, etc. The classes include single-object classes such as airplane, clock, bicycle, etc. We consider the domains as the style classes and the single-object classes as the content classes. We randomly pick 50 content classes and 10 images from each of the style class. Finally, we obtain 60 images per content classes where we obtain a dataset with 4000 images. We term this dataset \textit{DomainNet-3k}. The experiments on this dataset highlight the importance of style-based features for style-based clustering and showcase the differences between style-based and content-based clustering.

\begin{figure*}[!t]
    \centering
   \includegraphics[width=\textwidth]{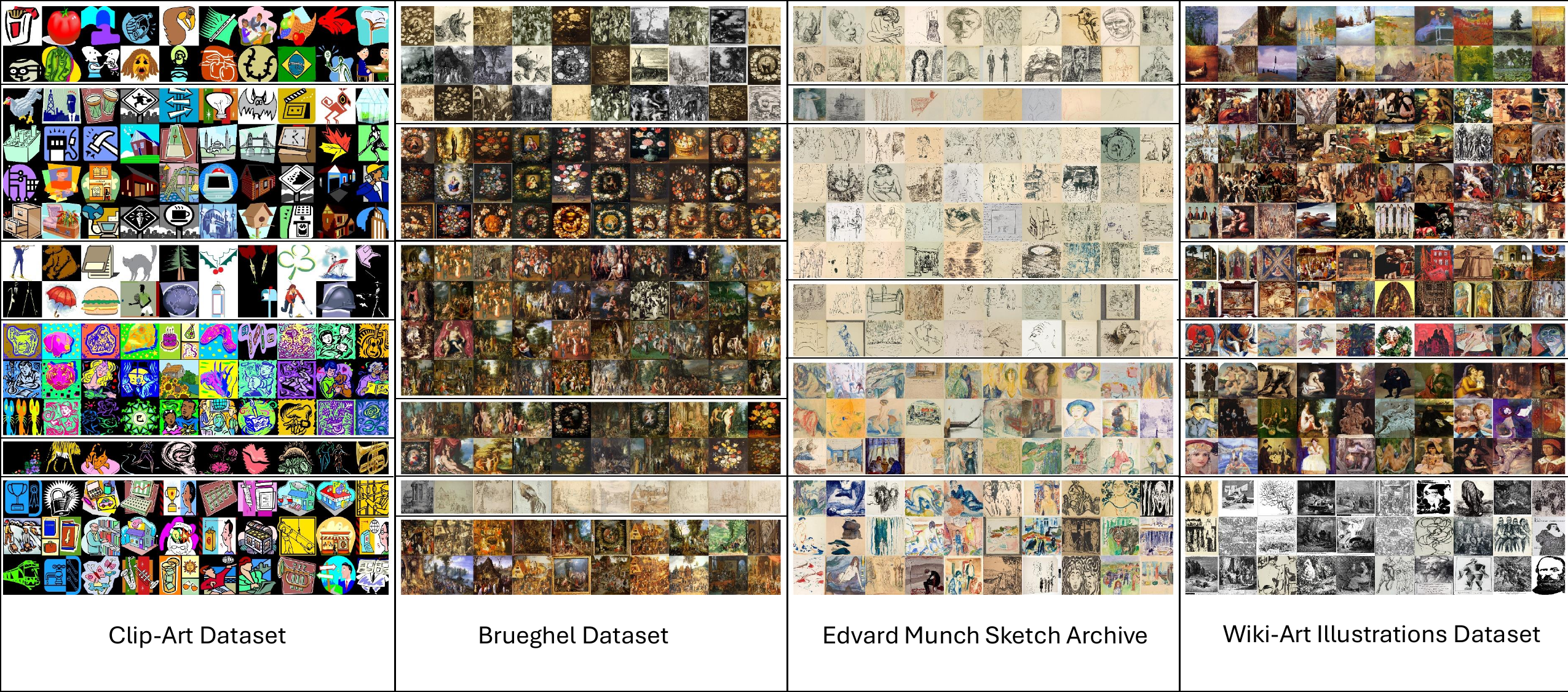}

   \caption{Cluster samples from the clustered artworks for four different core datasets using the style neural feature $F_{Gram}$ (Please zoom in for finer details.). }
    \label{fig:main_example}
\end{figure*}

\section{Experimental Setup and Evaluation Criteria}
\subsection{Clustering Models}
For our clustering approach, we utilize two different clustering models:
\begin{enumerate}
    \item K-Means Clustering Model
    \item Deep Embedding Clustering Model
\end{enumerate}
Please refer to Figure \ref{fig:DEC} for an architectural overview of both the clustering models. 

\subsubsection{K-Means Clustering Model}
The K-Means clustering model \cite{kmeans:macqueen1967some} first initializes a random set of points as cluster centroids which is equal to the number of clusters (K) provided as input. Each data point is assigned to a certain centroid based on which centroid has the smallest Euclidean distance to the specific data point. After all the data points are assigned to a certain centroid, we obtain K clusters where each cluster comprises a number of data points. After obtaining the clusters, a new set of centroids is calculated by averaging all the data points present in each cluster. The process of assigning data points to clusters is repeated again. This process is repeated until there is no change in the assignment of data points to a centroid, which gives us the final set of K clusters.  
\subsubsection{Deep Embedded Clustering Model(DEC)}
Since the K-Means Clustering model does not modify the input features, the clusters that are formed by K-Means do not pull the similar samples closer and push the dissimilar samples further apart across iterations.
Hence, we utilize the Deep Embedded Clustering model (DEC) \cite{xie2016unsupervised} wherein we project the input features into the latent space of a deep autoencoder, and then simultaneously learn the latent representations of each artwork's features and the cluster assignments. We chose the DEC model as our base clustering model since the traditional clustering methods often struggle to cluster complex high-dimensional data such as artworks and DEC can assist in finding the
non-linear relationships between the input features of artworks by learning better latent representations and cluster assignments of these features (Please refer to the Section \ref{sec:dec} for a brief background on the working of the DEC model). 

\subsection{Evaluation Metrics}
Similar to \cite{Castellano_2022}, we quantitatively evaluate the performance of our method using the two metrics: \textbf{Silhouette Coefficient (SC)} \cite{silhouette-coefficient} and \textbf{Calinski Harabasz Index (CHI)} \cite{chi}. SC is the measure of how similar a data point is to other data points in its own cluster and how similar the same data point is to the data points in a separate cluster. This metric is calculated on the data point level. SC ranges from -1 to +1, where a high value indicates that the data points are well-matched to their own clusters and poorly matched to other clusters. A lower value would indicate that the data points are wrongly assigned to clusters. CHI is the ratio of the sum of intra-cluster dispersion and inter-cluster dispersion for all clusters. This metric is calculated at the cluster level. A higher value of \textit{CHI} indicates that the clusters are more spread out and dense.

For experiments with curated datasets, as mentioned earlier we consider the images generated from the same style reference image as part of a cluster. We also consider two more evaluation metrics, viz., \textbf{Adjusted Rand Index (ARI)} \cite{ari-metric} and \textbf{Normalized Mutual Information (NMI)} \cite{metric-nmi} that take into consideration the ground truth clusters. Both these metrics are used in notable works for clustering such as \cite{chang2017deep,adaloglou2023exploringlimitsdeepimage,qian2023stableclusterdiscriminationdeep,Xu_2024}. ARI is a measure of similarity between two data clusterings. It takes all pairs of samples from both ground truth and predicted clusterings and considers all pairs of agreements and disagreements in their assignments to clusters. It then adjusts the index to account for chance by taking into account the expected similarity between the two clusterings. The ARI score ranges from -1 to 1. Values ranging between -1 to 0 indicate disagreement between the two data clusterings whereas values ranging from 0 to 1 indicate agreement between the two data clusterings. NMI is used to calculate the correlation between the ground truth clustering and the predicted clustering. The NMI ranges from 0 to 1, where a value closer to 0 would indicate no correlation between ground truth and predicted clusters whereas a value closer to 1 would indicate a near-perfect correlation between ground truth and predicted clusters.

\subsection{Human Qualitative Survey}
Doing an exhaustive Human Evaluation of the Clustering Quality comparing against different clustering across multiple neural representations is a challenging task. So we primarily rely on our quantitative results. However, we do conduct a human survey to understand the general perception of style and its cognition in the clustering through a limited study of 25 human participants with some inclination to visual arts.  Each participant was shown 5 high-resolution clustering images representing 5 different clustering approaches anonymously.  They were required to visually assess the clustering samples (refer to Fig. for as an example) and give ratings (1-10) based on the following three questions for each image:
\begin{itemize}
\item \textbf{Q1. Cluster Cohesiveness:} How close are the styles of artworks within each cluster? 
\item \textbf{Q2. Cluster Separation:} How well separated are the clusters stylistically from each other? 
\item \textbf{Q3. Overall Clustering Quality:} What is the overall quality of style-based clustering in this example?  
\end{itemize}

The survey takers were told that the clustering is based on style and other aspects of the image, such as content, are immaterial unless they influence the style.  Also, we did not impose any definition of style; the participants were made aware that they were free to use any definition or notion of style to perceive the similarity or dissimilarity of style in the given examples.
    
\section{Results and Discussion}
\label{sec: results}
We discuss the key results and insights in this section. We study the impact of twelve different neural features with two clustering models (K-Means and DEC) and gauge their effectiveness across two dimensions: \textit{stylistic clustering ability} and general \textit{cluster formation ability}. We primarily utilize \textbf{ARI} and \textbf{NMI} to gauge stylistic clustering ability against ground truths and \textbf{SC} and \textbf{CHI} metrics to the general cluster formation ability. We also present the UMAPs \cite{umap} (Uniform Manifold Approximation and Projection) for a few features in Section \ref{sec: additional_results}.
Additionally, we qualitatively evaluate the visual results from various clustering approaches. Our four key observations are structured around five questions:

\begin{table}[]
\begin{subtable}{0.45\textwidth}
\centering
\begin{tabular}{c|c|c|c}
\textbf{Features}           & \textbf{ARI} & \textbf{NMI} & \textbf{\begin{tabular}[c]{@{}c@{}}Qualitative\\ Rating\end{tabular}} \\ \hline
\textbf{$F_{Dense}$}        & 0.106        & 0.364        & Poor                                                                  \\
\textbf{$F_{ViT-ImageNet}$} & 0.1          & 0.422        & Poor                                                                  \\
\textbf{$F_{StyleCap}$}        & 0.15         & 0.435        & Poor                                                                  \\
\textbf{$F_{Mamba}$}          & 0.012        & 0.179        & Very Poor                                                             \\
\textbf{$F_{CSD}$}     & 0.116        & 0.401        & Poor                                                                 
\end{tabular}%
\caption{Content clustering}

\begin{tabular}{c|c|c|c}
\textbf{Features}           & \textbf{ARI} & \textbf{NMI} & \textbf{\begin{tabular}[c]{@{}c@{}}Qualitative\\ Rating\end{tabular}} \\ \hline
\textbf{$F_{Dense}$}        & 0.291        & 0.352        & Poor                                                                  \\
\textbf{$F_{ViT-ImageNet}$} & 0.282        & 0.49         & Poor                                                                  \\
\textbf{$F_{StyleCap}$}        & 0.547        & 0.591        & Good                                                                  \\
\textbf{$F_{Mamba}$}          & 0.514        & 0.561        & Good                                                                  \\
\textbf{$F_{CSD}$}     & 0.654        & 0.681        & Very Good                                                            
\end{tabular}%
\caption{Style clustering}
\end{subtable}
\caption{Quantitative results and indicative qualitative rating for (a) content-based clustering and (b) style-based clustering on the DomainNet-3k subset. The subset includes 3000 images from 6 style classes and 50 content classes.}

\label{tab:domainnet}
\end{table}

\subsection{Do we need specialized neural feature representations for style-based clustering of visual artworks?}
\textbf{TL;DR:} Yes, indeed.

Visual artworks present a challenge to the image classification models since the objects are not as readily identified as in the photographic images. We find that they struggle equally harder for style identification and clustering.  We consider the DomainNet Dataset (\cite{domainnet}) and five feature representations - two from generic, and three as representatives of style transfer, large visio-language, and custom-trained style models -  to investigate this question.  DomainNet divides its dataset from two perspectives - the content and the domain. We can approximate the domain to be indicative of style. We apply clustering through these features and evaluate the results from both - the content and the style-based perspectives.  

Refer to Table \ref{tab:domainnet}, and Figures \ref{fig:domain_qual_fig}, \ref{fig:content-style} for quantitative and visual comparison of the results. 
Specifically, we observe that: (i) $F_{Dense}$ and $F_{ViT-ImageNet}$ are generic representations and trained primarily on real-life images. So they perform poorly on both content and style-based clustering on the artworks. (iii) Style representations like $F_{StyleCap}$ and $F_{CSD}$  give better scores for this dataset for both types of clustering. This might be because the nature of style specification includes some element of content.  (ii)  specialized style representations, such as $F_{Mamba}$, as expected, perform poorly on the content-based clustering, but perform well on style clustering. 

This underlines a clear need and superiority of style-specific neural feature representations for style-based clustering. Next, we rigorously investigate each style-specific representation on our specially curated datasets - which create a very distinct separation of content and style images.

\subsection{Which style feature representation is most effective for style-based clustering?}

\textbf{TL;DR:} \textit{ $F_{ST}$, followed by $F_{CSD}$, features perform the best stylistic clustering.}

When we take a look at the ARI and NMI metrics for both the mixed synthetically curated datasets created using Styleshot (Table \ref{tab:mixed_styleshot}) and Mamba-ST (Table \ref{tab:mixed_mamba}), we observe that the $F_{ST}$ features consistently place in top 3 scores followed by $F_{CSD}$ features. This indicates that the style-transfer architectures, particularly the more modern ones,  effectively capture the style present in an image. They are also able to outperform $F_{Gram}$ which are widely used in different style-based applications. 
We further observe that $F_{Text}$ features give an average score on the MSC-4k dataset. It indicates that an LVLM can detect some style information present in an artwork and present it in a textual form. We observe that the generic image features like $F_{Dense}$ show a mediocre performance in style-capturing while $F_{g \cdot c}$, which performs well in style classification \cite{stclass16,stclass18}, is unable to form proper style clusters with the curated datasets.

We next observe the SC and CHI metrics in Table \ref{tab:mixed_styleshot} and Table \ref{tab:mixed_mamba} to check how well each feature can form dense and sparse clusters. We observe that the $F_{Text}$ features consistently place in the top 3 features in terms of metric scores. This indicates that $F_{text}$ features can form dense and spare clusters. This can be attributed to the textual descriptions of the artworks that the $F_{text}$ features encapsulate. $F_{StyleCap}$ features describe the style of an artwork while $F_{Annot}$ captures the specific style concepts that are present in an artwork. As both the $F_{text}$ features adhere to specific style definitions (style descriptions for $F_{StyleCap}$ and style taxonomy for $F_{Annot}$), the clustering model is better able to differentiate the style feature of each artwork. We observe that the best style features, i.e. the $F_{ST}$ features, stand next to $F_{Text}$ in the cluster formation ability, with $F_{StyleGAN}$ performing the best. Features like $F_{Dense}$ and $F_{Gram}$ form less dense and sparse clusters as seen in the KMeans metrics scores, however, they can improve significantly upon passing through the DEC.

\afterpage{
\clearpage
\begin{figure*}[!t]
 


\includegraphics[width=.85\textwidth]{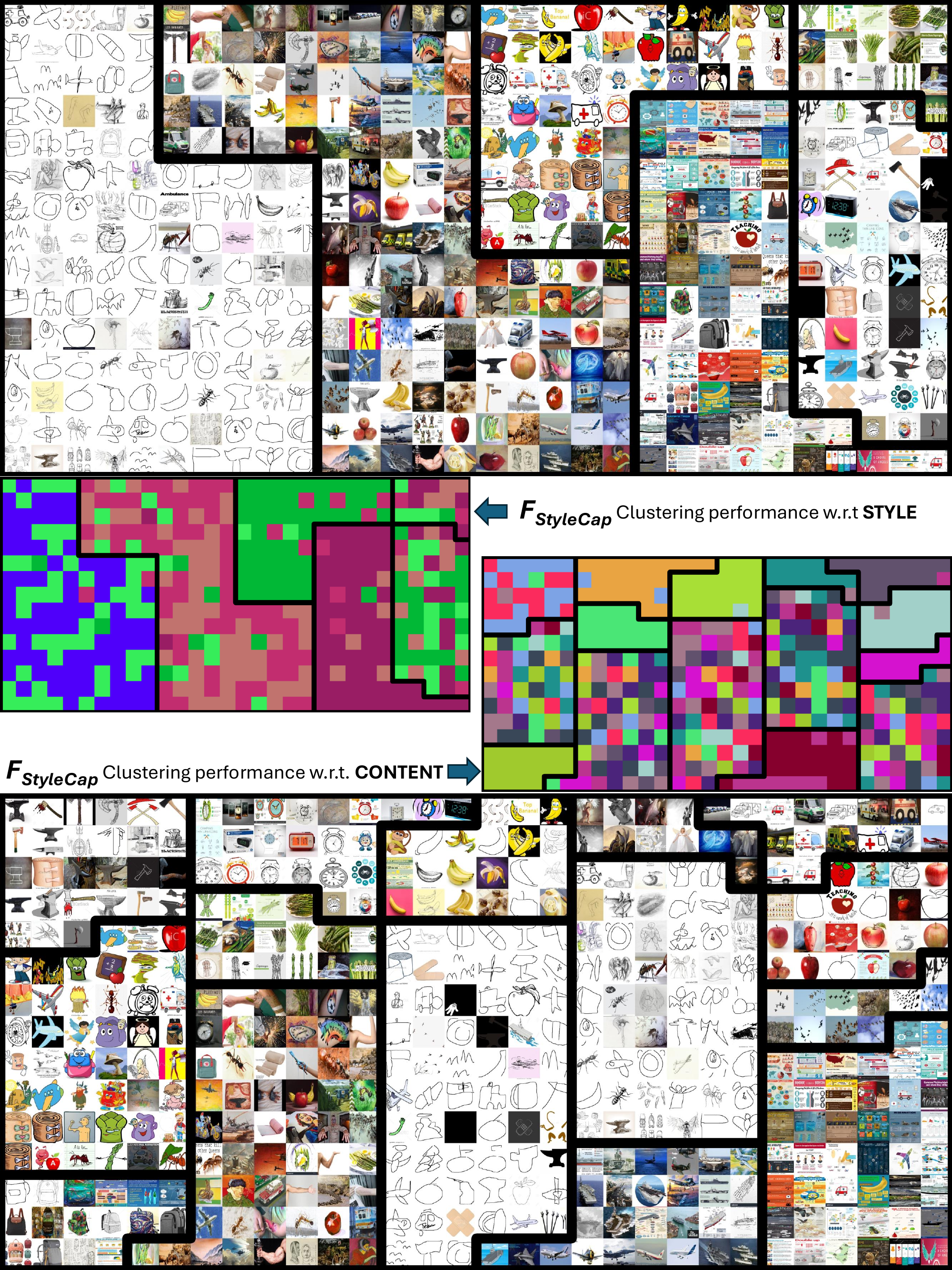}
   \caption{Qualitative comparison of content based clustering and style-based clustering on the DomainNet-3k with the $F_{StyleCap}$ features. We show 450 images in both types of clustering with 15 clusters in the content-based clustering and 6 clusters in the style-based clustering.}
  \label{fig:domain_qual_fig}
\end{figure*}

\begin{figure*}[p]
    \centering
    \includegraphics[width=.85\textwidth]{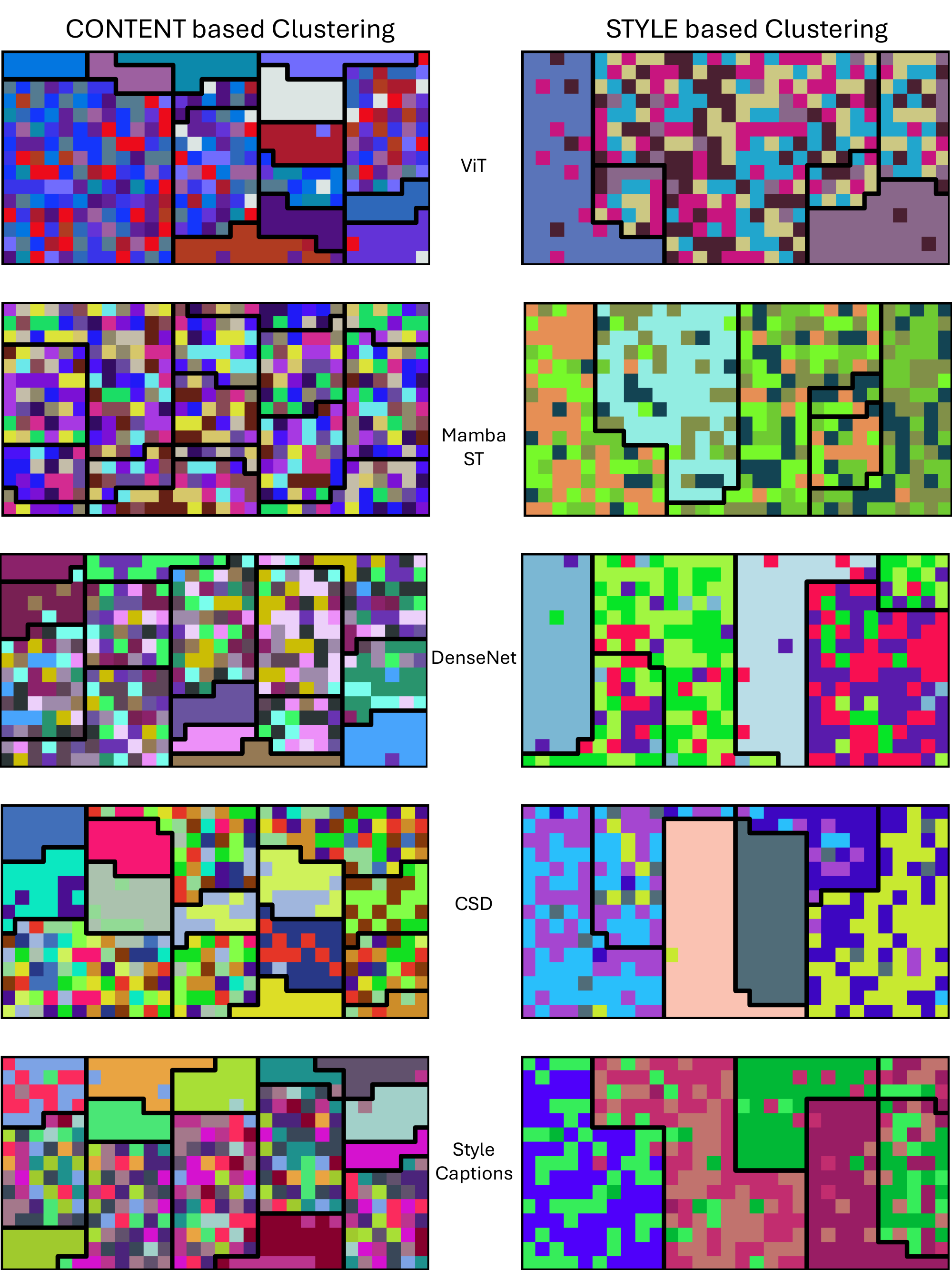}
    \caption{Visual comparison of the relative effectiveness of style-based and content-based clustering through the select five neural feature representations. For perfect clustering, each cluster would have a distinct and homogeneous color patches.}
    \label{fig:content-style}
\end{figure*}

\begin{figure*}[!t]
 \begin{subfigure}[t]{.48\textwidth}
    \centering
    \includegraphics[width=\textwidth]{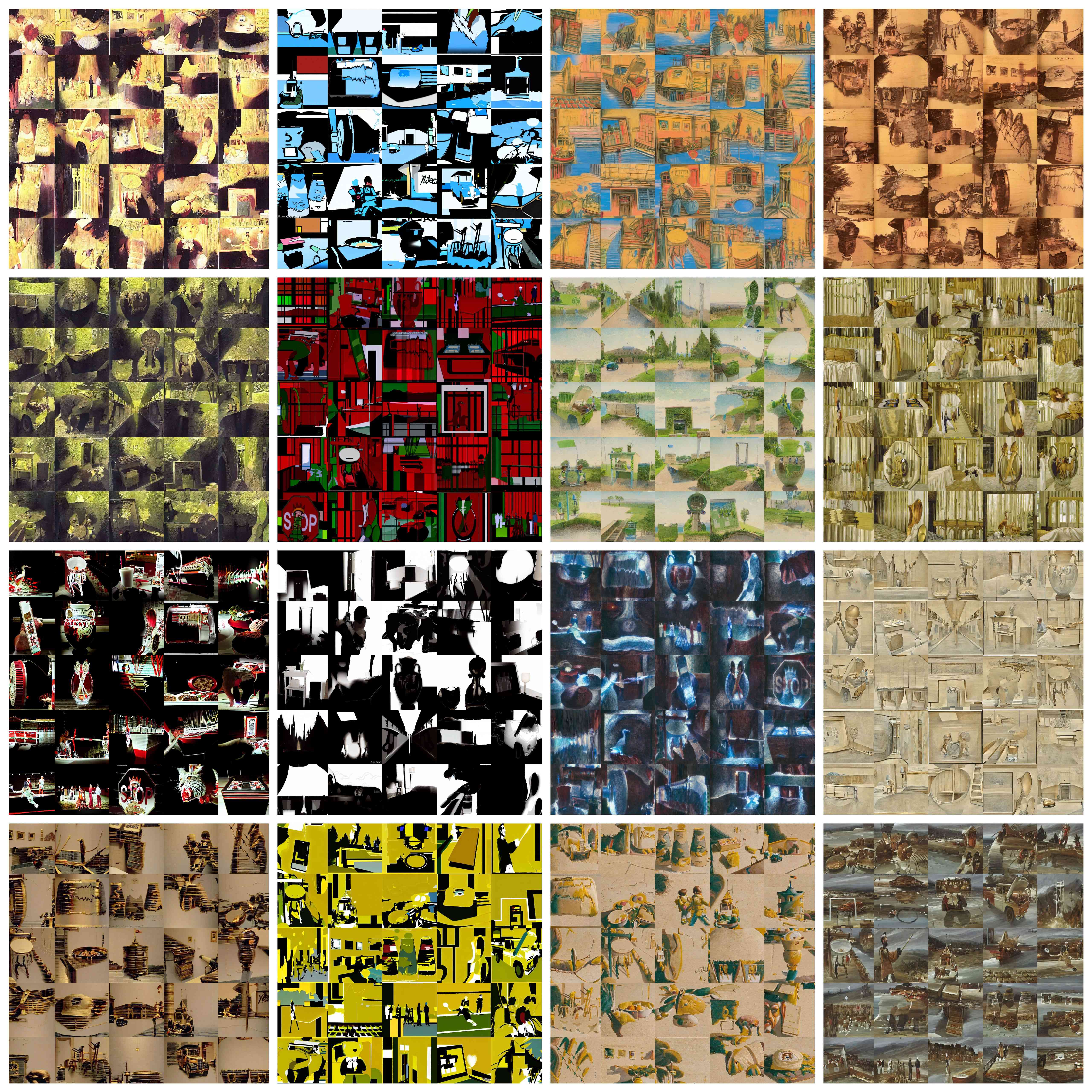}
    \caption{Ground Truth}
  \end{subfigure}
   \hfill
 \begin{subfigure}[t]{0.48\textwidth}
    \centering
    \includegraphics[width=\textwidth]{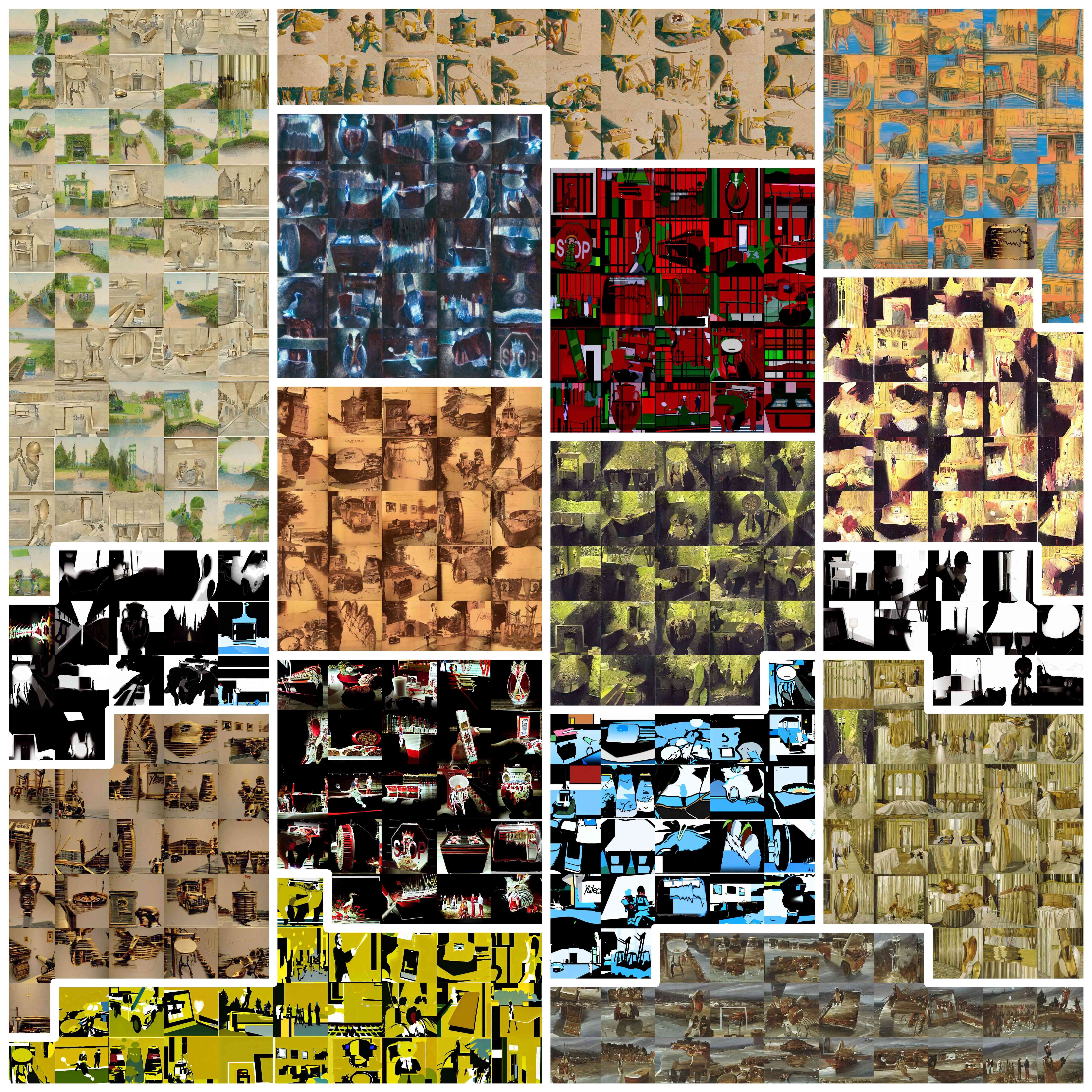}
    \caption {$F_{Stytr2}$ clusters}

  \end{subfigure}
  \begin{subfigure}[t]{.48\textwidth}
    \centering
    \includegraphics[width=\textwidth]{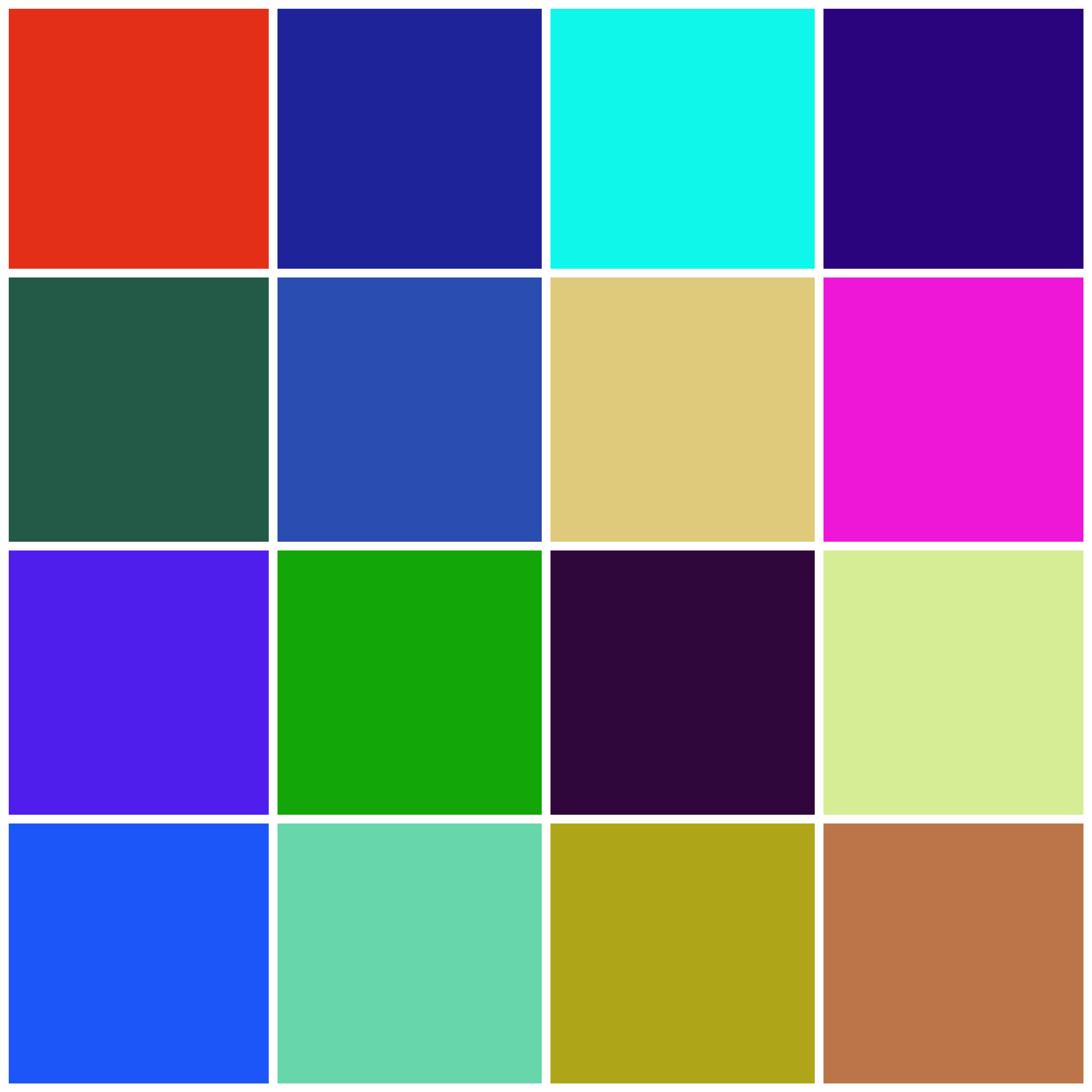}
    \caption{Ground truth color representations per cluster}
  \end{subfigure}
   \hfill
 \begin{subfigure}[t]{.48\textwidth}
    \centering
    \includegraphics[width= .96\textwidth]{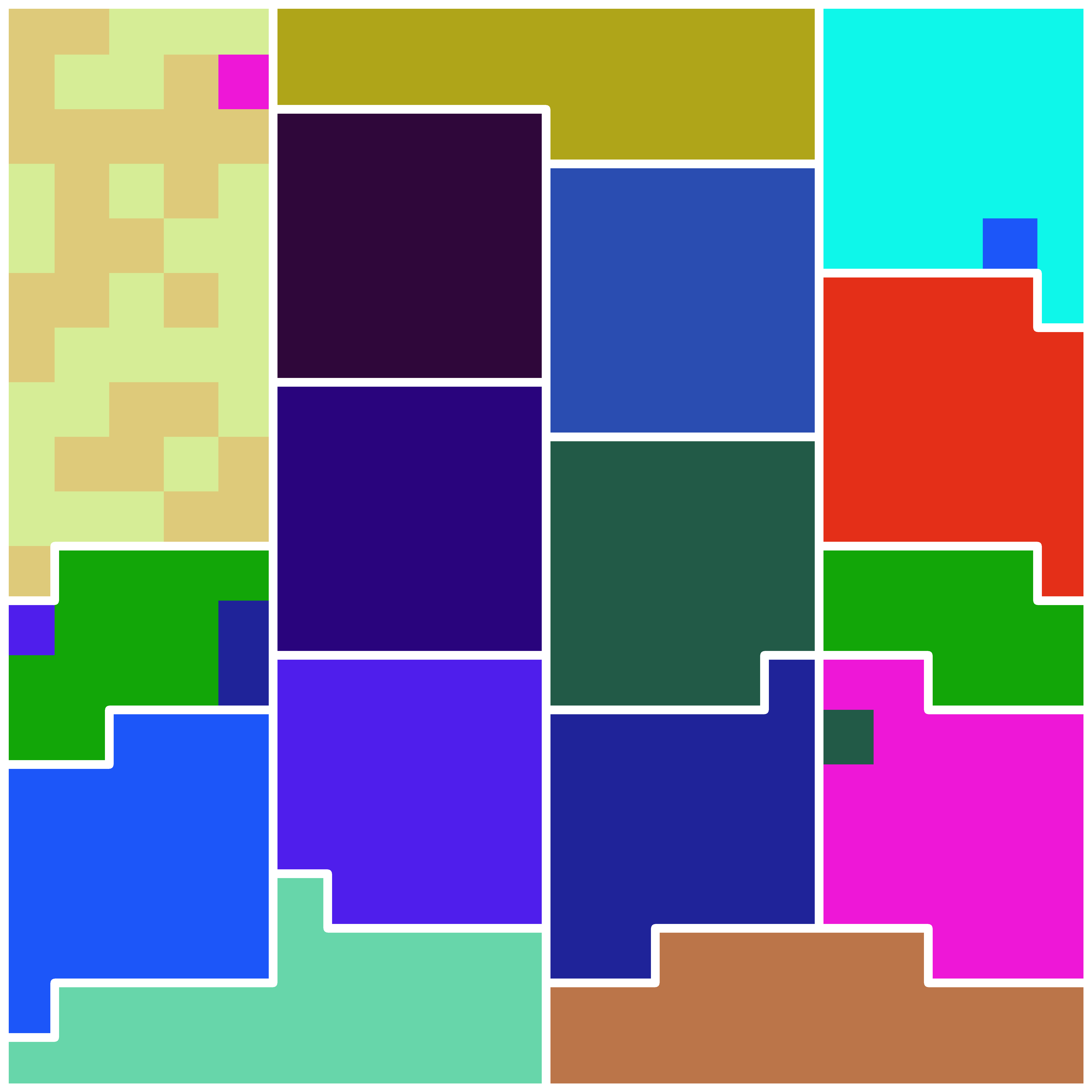}
    \caption {$F_{Stytr2}$ color representations per cluster}
  \end{subfigure}

   \caption{Qualitative comparison of 16 clusters for the Mixed Styleshot Curated (MSC-4k) ground truth and clusters obtained from $F_{Stytr2}$. Here, we show the artworks present in different clusters in (a) and (b). Furthermore, in  (c) and (d), we show the color representation associated with the ground truth for both clusterings.}
  \label{fig:qual_fig_mix}
\end{figure*}

\clearpage
}

\afterpage{
\clearpage
\begin{table*}[p]
\centering
\resizebox{\textwidth}{!}{%
\begin{tabular}{cc|cc|cc|ccc|ccc}
\multicolumn{2}{c|}{{\color[HTML]{000000} }}                                                                      & \multicolumn{2}{c|}{{\color[HTML]{000000} \textbf{ARI}}}                                                                        & \multicolumn{2}{c|}{{\color[HTML]{000000} \textbf{NMI}}}                                                                      & \multicolumn{3}{c|}{{\color[HTML]{000000} \textbf{SC}}}                                                                                                                                     & \multicolumn{3}{c}{{\color[HTML]{000000} \textbf{CHI}}}                                                                                                                                           \\ \cline{3-12} 
\multicolumn{2}{c|}{\multirow{-2}{*}{{\color[HTML]{000000} \textbf{Features}}}}                                   & \multicolumn{1}{c|}{{\color[HTML]{000000} \textbf{K-Means}}}             & {\color[HTML]{000000} \textbf{DEC}}                  & \multicolumn{1}{c|}{{\color[HTML]{000000} \textbf{K-Means}}}             & {\color[HTML]{000000} \textbf{DEC}}                  & \multicolumn{1}{c|}{{\color[HTML]{000000} \textbf{Base}}} & \multicolumn{1}{c|}{{\color[HTML]{000000} \textbf{K-Means}}}             & {\color[HTML]{000000} \textbf{DEC}}                  & \multicolumn{1}{c|}{{\color[HTML]{000000} \textbf{Base}}} & \multicolumn{1}{c|}{{\color[HTML]{000000} \textbf{K-Means}}}                & {\color[HTML]{000000} \textbf{DEC}}                     \\ \hline
\multicolumn{1}{c|}{{\color[HTML]{000000} }}                             & {\color[HTML]{000000} $F_{Dense}$}     & \multicolumn{1}{c|}{{\color[HTML]{000000} 0.594}}                        & {\color[HTML]{000000} 0.611}                         & \multicolumn{1}{c|}{{\color[HTML]{000000} 0.893}}                        & {\color[HTML]{000000} 0.861}                         & \multicolumn{1}{c|}{{\color[HTML]{000000} 0.078}}         & \multicolumn{1}{c|}{{\color[HTML]{000000} 0.071}}                        & {\color[HTML]{000000} 0.709}                         & \multicolumn{1}{c|}{{\color[HTML]{000000} 41.89}}         & \multicolumn{1}{c|}{{\color[HTML]{000000} 41.06}}                           & {\color[HTML]{000000} 3379.81}                          \\
\multicolumn{1}{c|}{{\color[HTML]{000000} }}                             & {\color[HTML]{000000} $F_{Gram}$}      & \multicolumn{1}{c|}{{\color[HTML]{000000} 0.837}}                        & {\color[HTML]{000000} 0.684}                         & \multicolumn{1}{c|}{{\color[HTML]{000000} 0.932}}                        & {\color[HTML]{000000} 0.889}                         & \multicolumn{1}{c|}{{\color[HTML]{000000} 0.221}}         & \multicolumn{1}{c|}{{\color[HTML]{000000} 0.205}}                        & {\color[HTML]{000000} 0.724}                          & \multicolumn{1}{c|}{{\color[HTML]{000000} 270.27}}        & \multicolumn{1}{c|}{{\color[HTML]{000000} 293.25}}                          & {\color[HTML]{000000} 5459.39}                          \\
\multicolumn{1}{c|}{\multirow{-3}{*}{{\color[HTML]{000000} $F_{CNN}$}}}  & {\color[HTML]{000000} $F_{g \cdot c}$} & \multicolumn{1}{c|}{{\color[HTML]{000000} 0.078}}                        & {\color[HTML]{000000} 0.079}                         & \multicolumn{1}{c|}{{\color[HTML]{000000} 0.343}}                        & {\color[HTML]{000000} 0.344}                         & \multicolumn{1}{c|}{{\color[HTML]{000000} -0.229}}        & \multicolumn{1}{c|}{{\color[HTML]{000000} 0.28}}                         & {\color[HTML]{000000} 0.51}                          & \multicolumn{1}{c|}{{\color[HTML]{000000} 805.3}}         & \multicolumn{1}{c|}{\cellcolor[HTML]{F2CEEF}{\color[HTML]{000000} 25425.8}} & \cellcolor[HTML]{F2CEEF}{\color[HTML]{000000} 84313.98} \\ \hline
\multicolumn{1}{c|}{{\color[HTML]{000000} }}                             & {\color[HTML]{000000} $F_{StyleCap}$}   & \multicolumn{1}{c|}{{\color[HTML]{000000} 0.347}}                        & {\color[HTML]{000000} 0.334}                         & \multicolumn{1}{c|}{{\color[HTML]{000000} 0.565}}                        & {\color[HTML]{000000} 0.567}                         & \multicolumn{1}{c|}{{\color[HTML]{000000} 0.018}}         & \multicolumn{1}{c|}{{\color[HTML]{000000} 0.04}}                         & \cellcolor[HTML]{CAEDFB}{\color[HTML]{000000} 0.827} & \multicolumn{1}{c|}{{\color[HTML]{000000} 30.99}}         & \multicolumn{1}{c|}{{\color[HTML]{000000} 39.09}}                           & \cellcolor[HTML]{C0F1C8}{\color[HTML]{000000} 59267.93} \\
\multicolumn{1}{c|}{\multirow{-2}{*}{{\color[HTML]{000000} $F_{Text}$}}} & {\color[HTML]{000000} $F_{Annot}$}     & \multicolumn{1}{c|}{{\color[HTML]{000000} 0.213}}                        & {\color[HTML]{000000} 0.214}                         & \multicolumn{1}{c|}{{\color[HTML]{000000} 0.467}}                        & {\color[HTML]{000000} 0.457}                         & \multicolumn{1}{c|}{{\color[HTML]{000000} -0.0003}}       & \multicolumn{1}{c|}{{\color[HTML]{000000} 0.027}}                        & \cellcolor[HTML]{F2CEEF}{\color[HTML]{000000} 0.936} & \multicolumn{1}{c|}{{\color[HTML]{000000} 31.535}}        & \multicolumn{1}{c|}{{\color[HTML]{000000} 44.25}}                           & \cellcolor[HTML]{CAEDFB}{\color[HTML]{000000} 57340.58} \\ \hline
\multicolumn{1}{c|}{{\color[HTML]{000000} }}                             & {\color[HTML]{000000} $F_{StyleGAN}$}  & \multicolumn{1}{c|}{{\color[HTML]{000000} 0.5}}                          & {\color[HTML]{000000} 0.478}                         & \multicolumn{1}{c|}{{\color[HTML]{000000} 0.758}}                        & {\color[HTML]{000000} 0.719}                         & \multicolumn{1}{c|}{{\color[HTML]{000000} -0.03}}         & \multicolumn{1}{c|}{{\color[HTML]{000000} -0.003}}                       & \cellcolor[HTML]{C0F1C8}{\color[HTML]{000000} 0.869} & \multicolumn{1}{c|}{{\color[HTML]{000000} 17.65}}         & \multicolumn{1}{c|}{{\color[HTML]{000000} 19.24}}                           & {\color[HTML]{000000} 12064.25}                         \\
\multicolumn{1}{c|}{{\color[HTML]{000000} }}                             & {\color[HTML]{000000} $F_{Stytr2}$}    & \multicolumn{1}{c|}{\cellcolor[HTML]{C0F1C8}{\color[HTML]{000000} 0.91}} & {\color[HTML]{000000} 0.676}                         & \multicolumn{1}{c|}{{\color[HTML]{000000} 0.95}}                         & {\color[HTML]{000000} 0.867}                         & \multicolumn{1}{c|}{{\color[HTML]{000000} 0.482}}         & \multicolumn{1}{c|}{\cellcolor[HTML]{F2CEEF}{\color[HTML]{000000} 0.45}} & {\color[HTML]{000000} 0.377}                         & \multicolumn{1}{c|}{{\color[HTML]{000000} 1065.08}}       & \multicolumn{1}{c|}{{\color[HTML]{000000} 1132}}                            & {\color[HTML]{000000} 3019.4}                           \\
\multicolumn{1}{c|}{{\color[HTML]{000000} }}                             & {\color[HTML]{000000} $F_{Mamba}$}     & \multicolumn{1}{c|}{\cellcolor[HTML]{C0F1C8}{\color[HTML]{000000} 0.91}} & \cellcolor[HTML]{CAEDFB}{\color[HTML]{000000} 0.771} & \multicolumn{1}{c|}{\cellcolor[HTML]{CAEDFB}{\color[HTML]{000000} 0.96}} & \cellcolor[HTML]{CAEDFB}{\color[HTML]{000000} 0.919} & \multicolumn{1}{c|}{{\color[HTML]{000000} 0.443}}         & \multicolumn{1}{c|}{\cellcolor[HTML]{C0F1C8}{\color[HTML]{000000} 0.42}} & {\color[HTML]{000000} 0.526}                         & \multicolumn{1}{c|}{{\color[HTML]{000000} 652.4}}         & \multicolumn{1}{c|}{\cellcolor[HTML]{C0F1C8}{\color[HTML]{000000} 646.91}}  & {\color[HTML]{000000} 7759.72}                          \\
\multicolumn{1}{c|}{\multirow{-4}{*}{{\color[HTML]{000000} $F_{ST}$}}}   & {\color[HTML]{000000} $F_{StyleShot}$} & \multicolumn{1}{c|}{\cellcolor[HTML]{CAEDFB}{\color[HTML]{000000} 0.9}}  & \cellcolor[HTML]{C0F1C8}{\color[HTML]{000000} 0.87}  & \multicolumn{1}{c|}{\cellcolor[HTML]{C0F1C8}{\color[HTML]{000000} 0.97}} & \cellcolor[HTML]{F2CEEF}{\color[HTML]{000000} 0.951} & \multicolumn{1}{c|}{{\color[HTML]{000000} 0.399}}         & \multicolumn{1}{c|}{\cellcolor[HTML]{CAEDFB}{\color[HTML]{000000} 0.39}} & {\color[HTML]{000000} 0.825}                         & \multicolumn{1}{c|}{{\color[HTML]{000000} 364.31}}        & \multicolumn{1}{c|}{\cellcolor[HTML]{CAEDFB}{\color[HTML]{000000} 364.6}}   & {\color[HTML]{000000} 54651.85}                          \\ \hline
\multicolumn{1}{c|}{{\color[HTML]{000000} $F_{Train}$}}                 & {\color[HTML]{000000} $F_{CSD}$}       & \multicolumn{1}{c|}{\cellcolor[HTML]{F2CEEF}{\color[HTML]{000000} 0.96}} & \cellcolor[HTML]{F2CEEF}{\color[HTML]{000000} 0.831} & \multicolumn{1}{c|}{\cellcolor[HTML]{F2CEEF}{\color[HTML]{000000} 0.98}} & \cellcolor[HTML]{C0F1C8}{\color[HTML]{000000} 0.922} & \multicolumn{1}{c|}{{\color[HTML]{000000} 0.281}}         & \multicolumn{1}{c|}{{\color[HTML]{000000} 0.27}}                         & {\color[HTML]{000000} 0.502}                         & \multicolumn{1}{c|}{{\color[HTML]{000000} 195.82}}        & \multicolumn{1}{c|}{{\color[HTML]{000000} 192.79}}                          & {\color[HTML]{000000} 1858.74}                         
\end{tabular}%
}
\caption{Metrics scores for the \textbf{Mixed curated} dataset created using \textbf{Styleshot} (MSC-4k) for all features for both K-Means and DEC model. The \colorbox[HTML]{F2CEEF}{best}, \colorbox[HTML]{C0F1C8}{second best} and the \colorbox[HTML]{CAEDFB}{third best} results are highlighted for each metric. The mixed curated dataset contains \textbf{4000} images and \textbf{40} different styles. The range of values for each metric are as follows: \textit{ARI}: -1 to 1, \textit{NMI}: 0 to 1 , \textit{SC}: -1 to 1 and \textit{CHI}: 0 to $\infty$. The \textbf{Base} column indicates the SC and CHI values with perfect ground truth and no modification to the input embedding.}
\label{tab:mixed_styleshot}
\end{table*}

\begin{table*}[p]
\centering
\resizebox{\textwidth}{!}{%
\begin{tabular}{cc|cc|cc|ccc|ccc}
\multicolumn{2}{c|}{{\color[HTML]{000000} }}                                                                      & \multicolumn{2}{c|}{{\color[HTML]{000000} \textbf{ARI}}}                                                                        & \multicolumn{2}{c|}{{\color[HTML]{000000} \textbf{NMI}}}                                                                      & \multicolumn{3}{c|}{{\color[HTML]{000000} \textbf{SC}}}                                                                                                                                     & \multicolumn{3}{c}{{\color[HTML]{000000} \textbf{CHI}}}                                                                                                                                            \\ \cline{3-12} 
\multicolumn{2}{c|}{\multirow{-2}{*}{{\color[HTML]{000000} \textbf{Features}}}}                                   & \multicolumn{1}{c|}{{\color[HTML]{000000} \textbf{K-Means}}}             & {\color[HTML]{000000} \textbf{DEC}}                  & \multicolumn{1}{c|}{{\color[HTML]{000000} \textbf{K-Means}}}             & {\color[HTML]{000000} \textbf{DEC}}                  & \multicolumn{1}{c|}{{\color[HTML]{000000} \textbf{Base}}} & \multicolumn{1}{c|}{{\color[HTML]{000000} \textbf{K-Means}}}             & {\color[HTML]{000000} \textbf{DEC}}                  & \multicolumn{1}{c|}{{\color[HTML]{000000} \textbf{Base}}} & \multicolumn{1}{c|}{{\color[HTML]{000000} \textbf{K-Means}}}                & {\color[HTML]{000000} \textbf{DEC}}                      \\ \hline
\multicolumn{1}{c|}{{\color[HTML]{000000} }}                             & {\color[HTML]{000000} $F_{Dense}$}     & \multicolumn{1}{c|}{{\color[HTML]{000000} 0.164}}                        & {\color[HTML]{000000} 0.049}                         & \multicolumn{1}{c|}{{\color[HTML]{000000} 0.25}}                         & {\color[HTML]{000000} 0.13}                          & \multicolumn{1}{c|}{{\color[HTML]{000000} 0.02}}          & \multicolumn{1}{c|}{{\color[HTML]{000000} 0.097}}                        & \cellcolor[HTML]{C0F1C8}{\color[HTML]{000000} 0.959} & \multicolumn{1}{c|}{{\color[HTML]{000000} 21.49}}         & \multicolumn{1}{c|}{{\color[HTML]{000000} 46.26}}                           & \cellcolor[HTML]{C0F1C8}{\color[HTML]{000000} 120196.98} \\
\multicolumn{1}{c|}{{\color[HTML]{000000} }}                             & {\color[HTML]{000000} $F_{Gram}$}      & \multicolumn{1}{c|}{{\color[HTML]{000000} 0.816}}                        & {\color[HTML]{000000} 0.698}                         & \multicolumn{1}{c|}{{\color[HTML]{000000} 0.961}}                        & {\color[HTML]{000000} 0.926}                         & \multicolumn{1}{c|}{{\color[HTML]{000000} 0.29}}          & \multicolumn{1}{c|}{{\color[HTML]{000000} 0.259}}                        & {\color[HTML]{000000} 0.514}                         & \multicolumn{1}{c|}{{\color[HTML]{000000} 613.29}}        & \multicolumn{1}{c|}{\cellcolor[HTML]{C0F1C8}{\color[HTML]{000000} 618.8}}   & {\color[HTML]{000000} 3750.41}                           \\
\multicolumn{1}{c|}{\multirow{-3}{*}{{\color[HTML]{000000} $F_{CNN}$}}}  & {\color[HTML]{000000} $F_{g \cdot c}$} & \multicolumn{1}{c|}{{\color[HTML]{000000} 0.134}}                        & {\color[HTML]{000000} 0.129}                         & \multicolumn{1}{c|}{{\color[HTML]{000000} 0.448}}                        & {\color[HTML]{000000} 0.446}                         & \multicolumn{1}{c|}{{\color[HTML]{000000} -0.096}}        & \multicolumn{1}{c|}{{\color[HTML]{000000} 0.16}}                         & {\color[HTML]{000000} 0.52}                          & \multicolumn{1}{c|}{{\color[HTML]{000000} 2205.62}}       & \multicolumn{1}{c|}{\cellcolor[HTML]{F2CEEF}{\color[HTML]{000000} 9985.58}} & \cellcolor[HTML]{CAEDFB}{\color[HTML]{000000} 89669.42}  \\ \hline
\multicolumn{1}{c|}{{\color[HTML]{000000} }}                             & {\color[HTML]{000000} $F_{StyleCap}$}   & \multicolumn{1}{c|}{{\color[HTML]{000000} 0.03}}                         & {\color[HTML]{000000} 0.01}                          & \multicolumn{1}{c|}{{\color[HTML]{000000} 0.092}}                        & {\color[HTML]{000000} 0.068}                         & \multicolumn{1}{c|}{{\color[HTML]{000000} -0.02}}         & \multicolumn{1}{c|}{{\color[HTML]{000000} 0.096}}                        & \cellcolor[HTML]{F2CEEF}{\color[HTML]{000000} 0.963} & \multicolumn{1}{c|}{{\color[HTML]{000000} 8.027}}         & \multicolumn{1}{c|}{{\color[HTML]{000000} 48.83}}                           & \cellcolor[HTML]{F2CEEF}{\color[HTML]{000000} 178514.62} \\
\multicolumn{1}{c|}{\multirow{-2}{*}{{\color[HTML]{000000} $F_{Text}$}}} & {\color[HTML]{000000} $F_{Annot}$}     & \multicolumn{1}{c|}{{\color[HTML]{000000} 0.053}}                        & {\color[HTML]{000000} 0.051}                         & \multicolumn{1}{c|}{{\color[HTML]{000000} 0.21}}                         & {\color[HTML]{000000} 0.212}                         & \multicolumn{1}{c|}{{\color[HTML]{000000} -0.037}}        & \multicolumn{1}{c|}{{\color[HTML]{000000} 0.024}}                        & \cellcolor[HTML]{CAEDFB}{\color[HTML]{000000} 0.936} & \multicolumn{1}{c|}{{\color[HTML]{000000} 15.37}}         & \multicolumn{1}{c|}{{\color[HTML]{000000} 44.64}}                           & {\color[HTML]{000000} 38490.53}                          \\ \hline
\multicolumn{1}{c|}{{\color[HTML]{000000} }}                             & {\color[HTML]{000000} $F_{StyleGAN}$}  & \multicolumn{1}{c|}{{\color[HTML]{000000} 0.417}}                        & {\color[HTML]{000000} 0.384}                         & \multicolumn{1}{c|}{{\color[HTML]{000000} 0.669}}                        & {\color[HTML]{000000} 0.634}                         & \multicolumn{1}{c|}{{\color[HTML]{000000} -0.03}}         & \multicolumn{1}{c|}{{\color[HTML]{000000} -0.005}}                       & {\color[HTML]{000000} 0.909}                         & \multicolumn{1}{c|}{{\color[HTML]{000000} 17.43}}         & \multicolumn{1}{c|}{{\color[HTML]{000000} 18.82}}                           & {\color[HTML]{000000} 19549.91}                          \\
\multicolumn{1}{c|}{{\color[HTML]{000000} }}                             & {\color[HTML]{000000} $F_{Stytr2}$}    & \multicolumn{1}{c|}{\cellcolor[HTML]{F2CEEF}{\color[HTML]{000000} 0.98}} & \cellcolor[HTML]{F2CEEF}{\color[HTML]{000000} 0.991} & \multicolumn{1}{c|}{\cellcolor[HTML]{F2CEEF}{\color[HTML]{000000} 0.99}} & \cellcolor[HTML]{F2CEEF}{\color[HTML]{000000} 0.995} & \multicolumn{1}{c|}{{\color[HTML]{000000} 0.6}}           & \multicolumn{1}{c|}{\cellcolor[HTML]{F2CEEF}{\color[HTML]{000000} 0.6}}  & {\color[HTML]{000000} 0.719}                         & \multicolumn{1}{c|}{{\color[HTML]{000000} 4368.14}}       & \multicolumn{1}{c|}{{\color[HTML]{000000} 4476}}                            & {\color[HTML]{000000} 15056.43}                          \\
\multicolumn{1}{c|}{{\color[HTML]{000000} }}                             & {\color[HTML]{000000} $F_{Mamba}$}     & \multicolumn{1}{c|}{\cellcolor[HTML]{CAEDFB}{\color[HTML]{000000} 0.9}}  & \cellcolor[HTML]{C0F1C8}{\color[HTML]{000000} 0.836} & \multicolumn{1}{c|}{\cellcolor[HTML]{CAEDFB}{\color[HTML]{000000} 0.97}} & \cellcolor[HTML]{C0F1C8}{\color[HTML]{000000} 0.94}  & \multicolumn{1}{c|}{{\color[HTML]{000000} 0.468}}         & \multicolumn{1}{c|}{\cellcolor[HTML]{C0F1C8}{\color[HTML]{000000} 0.45}} & {\color[HTML]{000000} 0.608}                         & \multicolumn{1}{c|}{{\color[HTML]{000000} 617.04}}        & \multicolumn{1}{c|}{\cellcolor[HTML]{CAEDFB}{\color[HTML]{000000} 617.93}}  & {\color[HTML]{000000} 7992.44}                           \\
\multicolumn{1}{c|}{\multirow{-4}{*}{{\color[HTML]{000000} $F_{ST}$}}}   & {\color[HTML]{000000} $F_{StyleShot}$} & \multicolumn{1}{c|}{\cellcolor[HTML]{C0F1C8}{\color[HTML]{000000} 0.96}} & \cellcolor[HTML]{CAEDFB}{\color[HTML]{000000} 0.748} & \multicolumn{1}{c|}{\cellcolor[HTML]{C0F1C8}{\color[HTML]{000000} 0.98}} & \cellcolor[HTML]{CAEDFB}{\color[HTML]{000000} 0.938} & \multicolumn{1}{c|}{{\color[HTML]{000000} 0.304}}         & \multicolumn{1}{c|}{\cellcolor[HTML]{CAEDFB}{\color[HTML]{000000} 0.28}} & {\color[HTML]{000000} 0.631}                         & \multicolumn{1}{c|}{{\color[HTML]{000000} 304.53}}        & \multicolumn{1}{c|}{{\color[HTML]{000000} 298.51}}                          & {\color[HTML]{000000} 5845.3}                            \\ \hline
\multicolumn{1}{c|}{{\color[HTML]{000000} $F_{Train}$}}                 & {\color[HTML]{000000} $F_{CSD}$}       & \multicolumn{1}{c|}{{\color[HTML]{000000} 0.86}}                         & {\color[HTML]{000000} 0.632}                         & \multicolumn{1}{c|}{{\color[HTML]{000000} 0.94}}                         & {\color[HTML]{000000} 0.839}                         & \multicolumn{1}{c|}{{\color[HTML]{000000} 0.141}}         & \multicolumn{1}{c|}{{\color[HTML]{000000} 0.13}}                         & {\color[HTML]{000000} 0.299}                         & \multicolumn{1}{c|}{{\color[HTML]{000000} 101.6}}         & \multicolumn{1}{c|}{{\color[HTML]{000000} 100.24}}                          & {\color[HTML]{000000} 1413.48}                          
\end{tabular}%
}
\caption{Metrics scores for the \textbf{Mixed curated} dataset created using \textbf{Mamba-ST} (MMC-4k) for all features for both K-Means and DEC model. The \colorbox[HTML]{F2CEEF}{best}, \colorbox[HTML]{C0F1C8}{second best} and the \colorbox[HTML]{CAEDFB}{third best} results are highlighted for each metric. The mixed curated dataset contains \textbf{4000} images and \textbf{40} different styles. The range of values for each metric are as follows: \textit{ARI}: -1 to 1, \textit{NMI}: 0 to 1 , \textit{SC}: -1 to 1 and \textit{CHI}: 0 to $\infty$.}
\label{tab:mixed_mamba}
\end{table*}
\clearpage
}

\subsection{Does the clustering method (K-Means or DEC) impact the style-based clustering?}

\textbf{TL;DR:}\textit{{ K-Means performs slightly more accurate stylistic clustering. DEC, on the other hand, forms distinct and well-separated stylistic clusters as compared to K-means baselines. However, improved clustering quality does not necessarily imply an improvement in style-based clustering quality. }} 

The results with the K-Means baseline reveal that the feature spaces are forming less sparse clusters. Hence, we test a deep clustering technique called DEC. DEC learns a new latent space with the help of an autoencoder and labels the data points using a clustering layer. We observe that DEC, with its newly learned latent space, can form distinct and well-separated clusters with the SC and CHI metric scores showing substantial improvement over the initial K-Means metric scores in most cases as shown in Table \ref{fig:k_vs_dec}. DEC brings the samples closer to or further away from the centroids based on the distance (Euclidean base) between their latent embeddings. It achieves this by using a KL divergence loss to maximize the cluster assignment probabilities for samples that are closer to centroids and minimize the assignment probabilities for samples that are away from the centroids. This is further supported if we take a look at the SC and CHI values in Table \ref{tab:mixed_styleshot} and Table \ref{tab:mixed_mamba}. The \textbf{base} column in both tables shows the SC and CHI scores if we provide the perfect ground truth as labels and calculate the metric without modifying the input features. We observe that the SC and CHI scores with DEC are comparatively higher than the base values indicating that DEC is learning the latent representations that can cluster well. This is further evident if we look at the UMAPs present in Section \ref{sec: additional_results}. Performance-wise, we find DEC to be more efficient.  We also observe that in our settings, for the WikiArt dataset with the most features, DEC  averages 5 minutes to cluster the artworks, while K-Means averages 18 minutes to cluster the artworks. 

Style-wise, DEC can match the performance of the K-Means baseline qualitatively, even though the ARI and NMI metrics are slightly low. This demonstrates that DEC is on par with K-Means in the stylistic clustering ability and outperforms K-Means in cluster formation ability. 

However, we also observe that an improvement in clustering quality does not necessarily imply an improvement in style-specific clustering quality. Referring to the Figures \ref{fig:k_vs_dec} we can observe that, while the SC and CHI scores improve on further iterations of the DEC model, the style-based clustering quality (as is evident from the ARI and NMI scores), mostly plateaus. This phenomenon was evident across different representations (for example, for the features $F_{StyleCap}$, $F_{Gram}$, $F_{StyleShot}$ as shown in Fig. \ref{fig:k_vs_dec}.).

\subsection{What does the style-based clustering reveal about the structural relationship among artworks' styles?}
\textbf{TL; DR:}\textit{ Style-based clustering provides an incisive peek into the style-based relationship among the artworks. Each neural style representation brings its style nuance. Further, the styles present in the majority of publicly available datasets tend to be hierarchical in nature.}

Style-based clustering provides an interesting perspective on the study of artworks and their historical evolution. Figure \ref{fig:clust_dist_graphic} shows a sample of clustering obtained through the neural style features $F_{StyleShot}$ and other neural style features. We can traverse the inter and intra-cluster pathways to discern interesting style relationships among artworks in different eras. Also, note how different neural style feature representations change the relative inter-painting style distances and how paintings grouped as different \textit{Art movements} in the WikiArt dataset occupy different groups depending upon the style representations.

In our experiments, we observe an interesting phenomenon: the styles present in the datasets that we use tend to be \textit{hierarchical}. We provide the following experimental evidence to support this observation:

    \textit{(i) Majority of the artworks are assigned to just a few clusters suggesting the presence of uneven-sized or hierarchical clusters}: For $F_{Gram}$ and $F_{g \cdot c}$, we observe that the majority of the artworks are assigned to just a few clusters (1-3) for the core datasets leading to un-even sized clusters (refer to Section \ref{sec: additional_results}). When we look at these specific clusters for all the datasets, we observe that the majority of artworks present in these clusters are style-wise close on the first inspection. However, when we perform sub-clustering on a single cluster for these datasets, we observe that the artworks can be further divided based on fine-granular style. For example, we experiment with sub-clustering a single cluster with the highest number of samples for the Brueghel dataset (refer to Figure \ref{fig: sub-cluster-samp} (i)). We observe that after sub-clustering, a seemingly singular style cluster can be further divided into distinct stylistic clusters. This indicates the presence of hierarchical clusters and that the high correlation within the $F_{Gram}$ and $F_{g \cdot c}$ features caused the artworks to be assigned to just a few clusters. The cluster distribution for different features can be found in the 
    \ref{sec: additional_results}.
    
    \textit{(ii) Applying hierarchical clustering to an artistic dataset shows the hierarchical styles present in an artistic dataset:}
    When we apply hierarchical clustering to the WikiArt dataset using $F_{StyleCap}$ features (refer to Figure \ref{fig:hier_dendro} (e)), we observe that each cluster is further divided into sub-clusters based on the style level. As we vary the cophenetic distance, we observe that when the cophenetic distance is high, the formed clusters capture the coarse-grained style information of the artworks. As we decrease the cophenetic distance, the clusters now contain artworks based on the fine-grained style aspects. This showcases that the styles in the publicly available artwork datasets tend to be hierarchical.
    
    This is further evident in Figure \ref{fig:hier_dendro} (c) and Figure \ref{fig:hier_dendro} (d) where we present the artwork-based dendrogram from the top 5 art movements and top 5 artists respectively. We observe that at lower levels of the dendrogram, the artworks present in each cluster are separated by the fine-grained style information. Moving up the dendrogram, we observe that artworks from similar art movements and artists merge into a specific cluster.
\afterpage{
    \clearpage
    
    \begin{figure*}[]
  \begin{subfigure}[t]{\textwidth}
    \centering
    \includegraphics[width=0.24\textwidth]{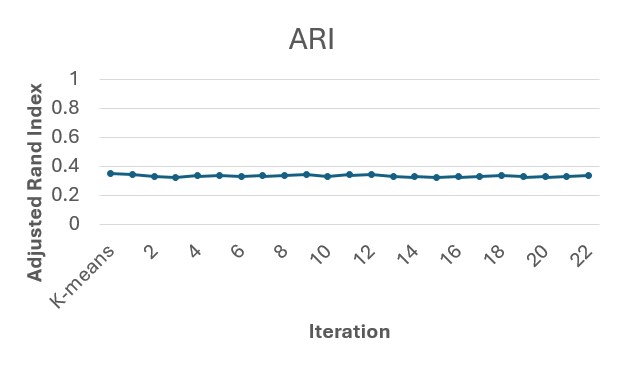}
   \hfill
    \includegraphics[width=0.24\textwidth]{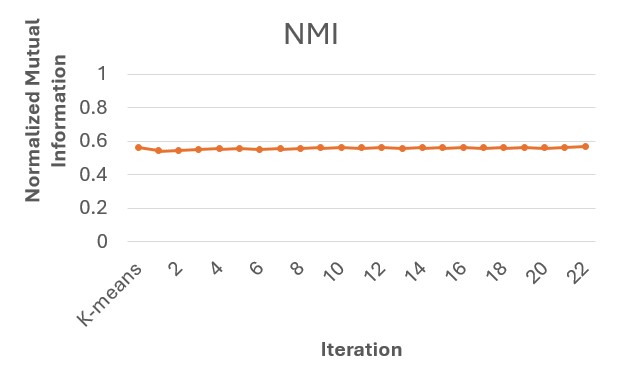}
  \hfill
    \includegraphics[width=0.24\textwidth]{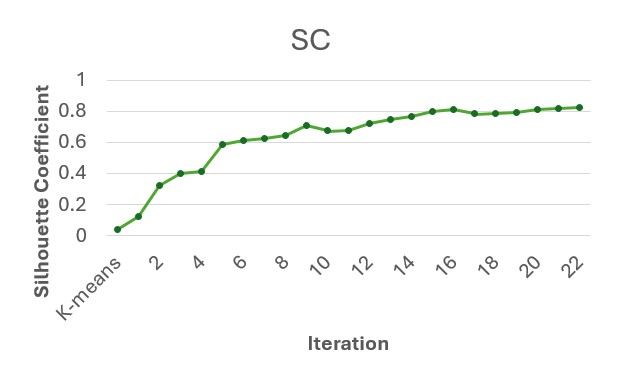}
  \hfill
    \includegraphics[width=0.24\textwidth]{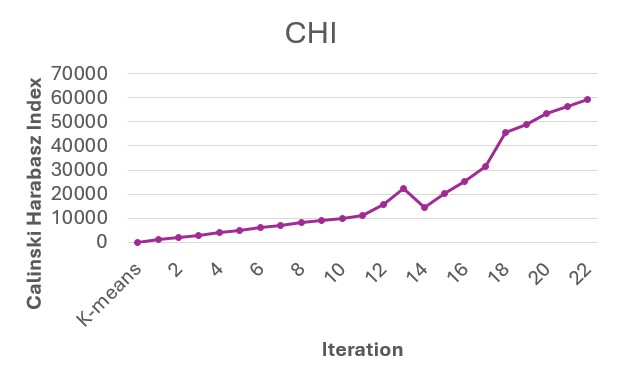}
    \caption{$F_{StyleCap}$}
  \end{subfigure}
  \begin{subfigure}[t]{\textwidth}
    \centering
    \includegraphics[width=0.24\textwidth]{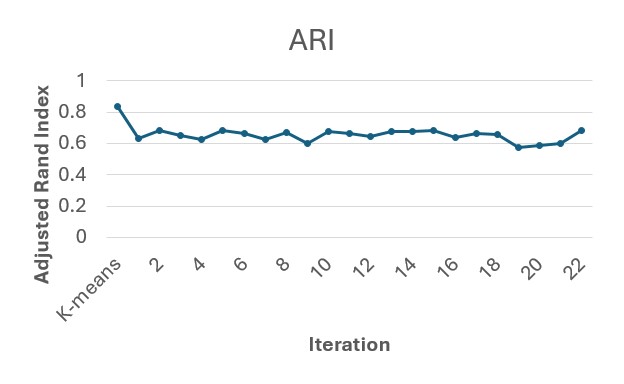}
   \hfill
    \includegraphics[width=0.24\textwidth]{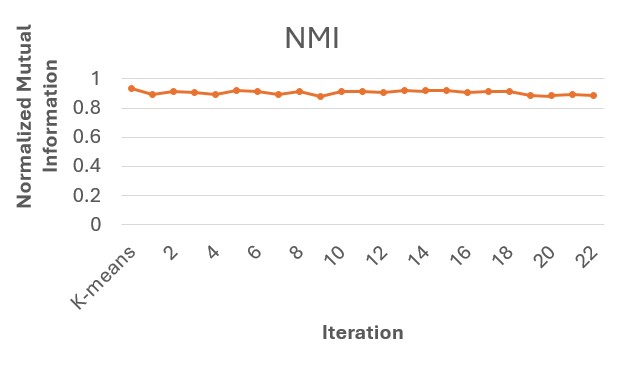}
  \hfill
    \includegraphics[width=0.24\textwidth]{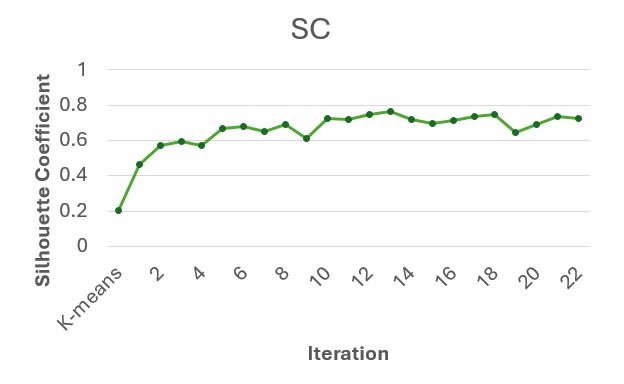}
  \hfill
    \includegraphics[width=0.24\textwidth]{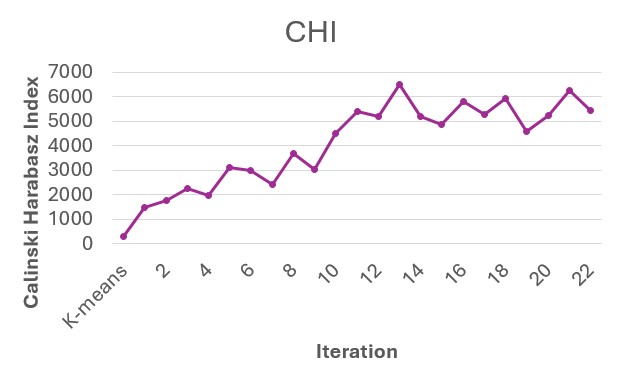}
    \caption{$F_{Gram}$}
  \end{subfigure}
  \begin{subfigure}[t]{\textwidth}
    \centering
    \includegraphics[width=0.24\textwidth]{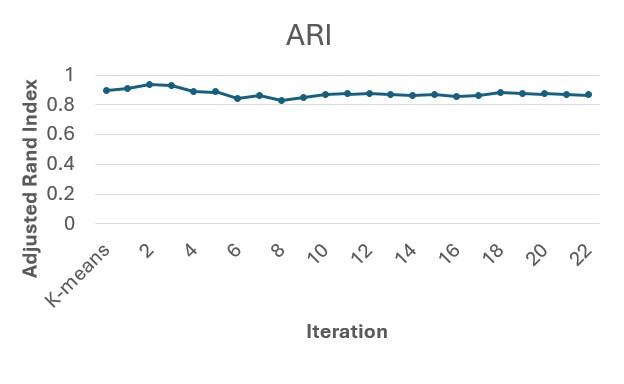}
   \hfill
    \includegraphics[width=0.24\textwidth]{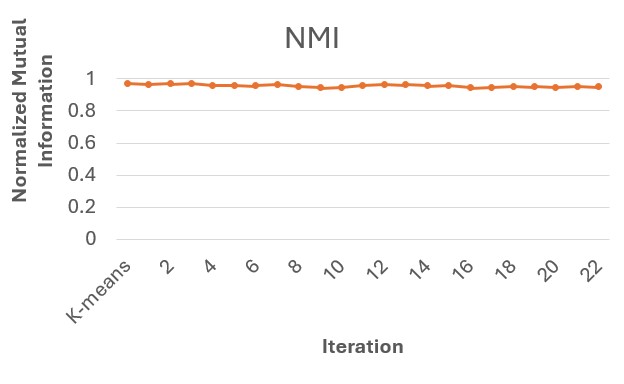}
  \hfill
    \includegraphics[width=0.24\textwidth]{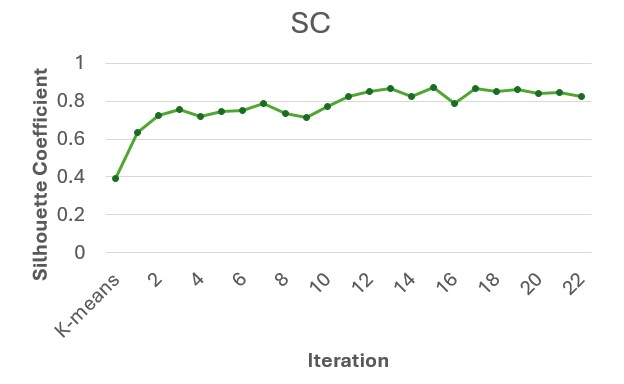}
  \hfill
    \includegraphics[width=0.24\textwidth]{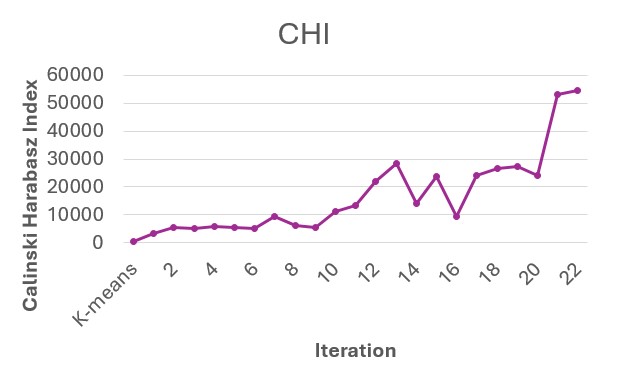}
    \caption{$F_{StyleShot}$}
  \end{subfigure}
  \caption{The change in different evaluation metric scores across DEC iterations with different features for the Mixed styleshot curated dataset (MSC-4k). The first iteration shows the results with K-Means cluster initialization in DEC. The subsequent iterations show the number of times the predicted labels are updated by the DEC model. We set the predicted labels in DEC to update after every 140 iterations.}
  \label{fig:k_vs_dec}
\end{figure*}
    
\begin{figure*}[p]

 \centering
 \fbox{\includegraphics[width=0.9\columnwidth]{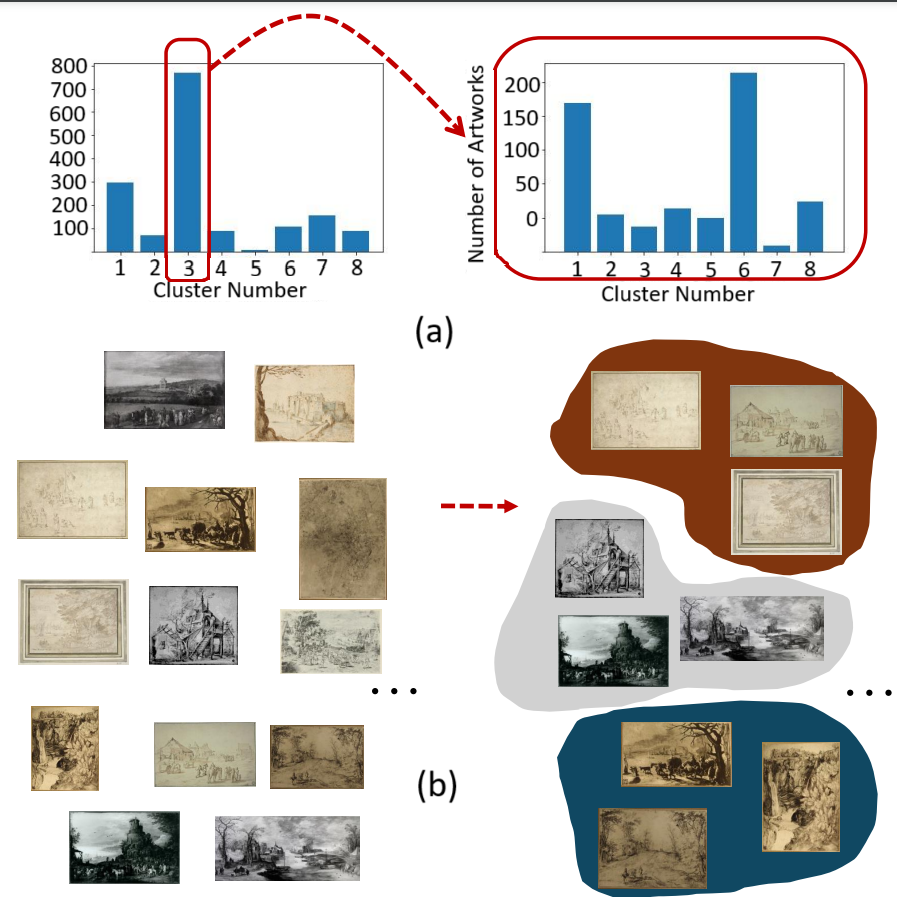}}

  \caption{Sub-clustering on a single cluster from the results of the Brueghel dataset for $F_{Gram}$ features through the DEC model. (a) shows the distribution of the number of samples in each cluster before and after sub-clustering. (b) shows the qualitative results after we obtain the sub clusters of a single cluster with most samples. Samples on the left are from the original cluster and samples on the right are from the sub-clusters.}
  \label{fig: sub-cluster-samp}
\end{figure*}
    
    \begin{figure*}[p]
    \centering
    \includegraphics[width=0.8\textwidth]{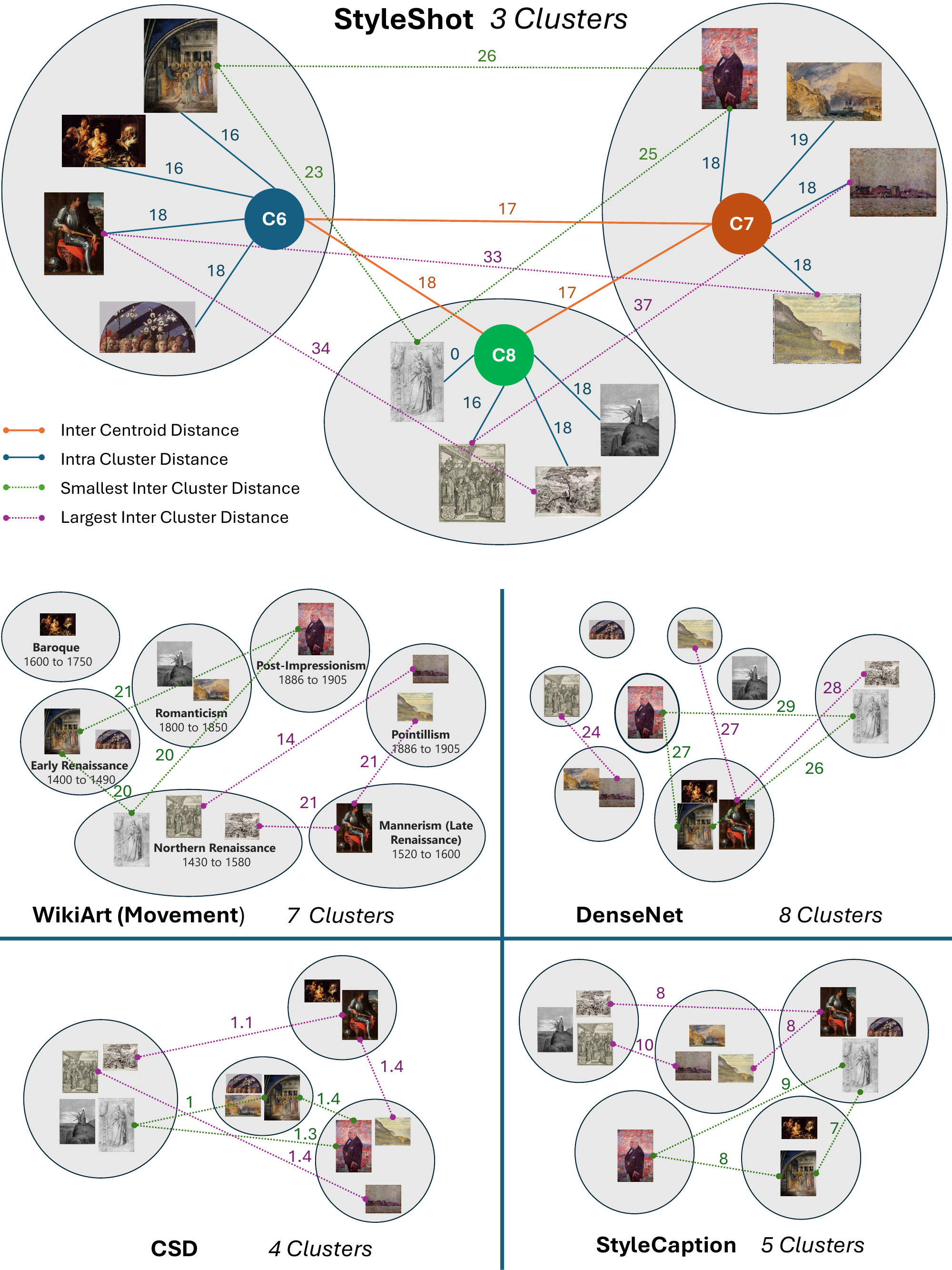}
    \caption{Inter-cluster distances and intra-cluster distances between a few samples from a few clusters obtained from the WikiArt-4k-AM subset.} 
    \label{fig:clust_dist_graphic}
\end{figure*}

    \begin{figure*}[p]

  \begin{subfigure}[t]{\textwidth}
    \centering
    \includegraphics[width=0.35\textwidth]{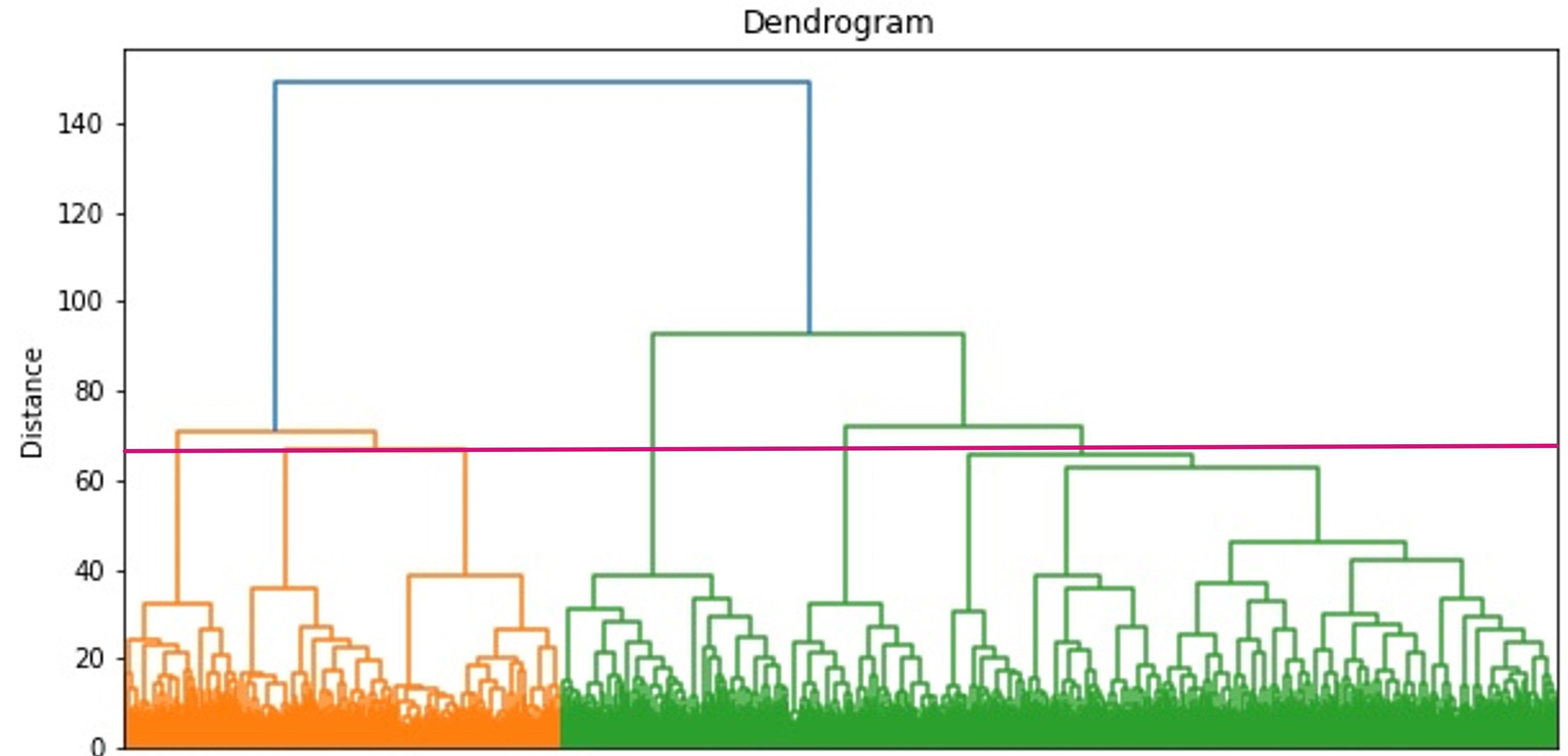}
    \caption{Complete dendrogram for the WikiArt-4k-AM subset}
  \end{subfigure}
  \newline
  \begin{subfigure}[t]{0.48\textwidth}
    \centering
    \includegraphics[width=\textwidth]{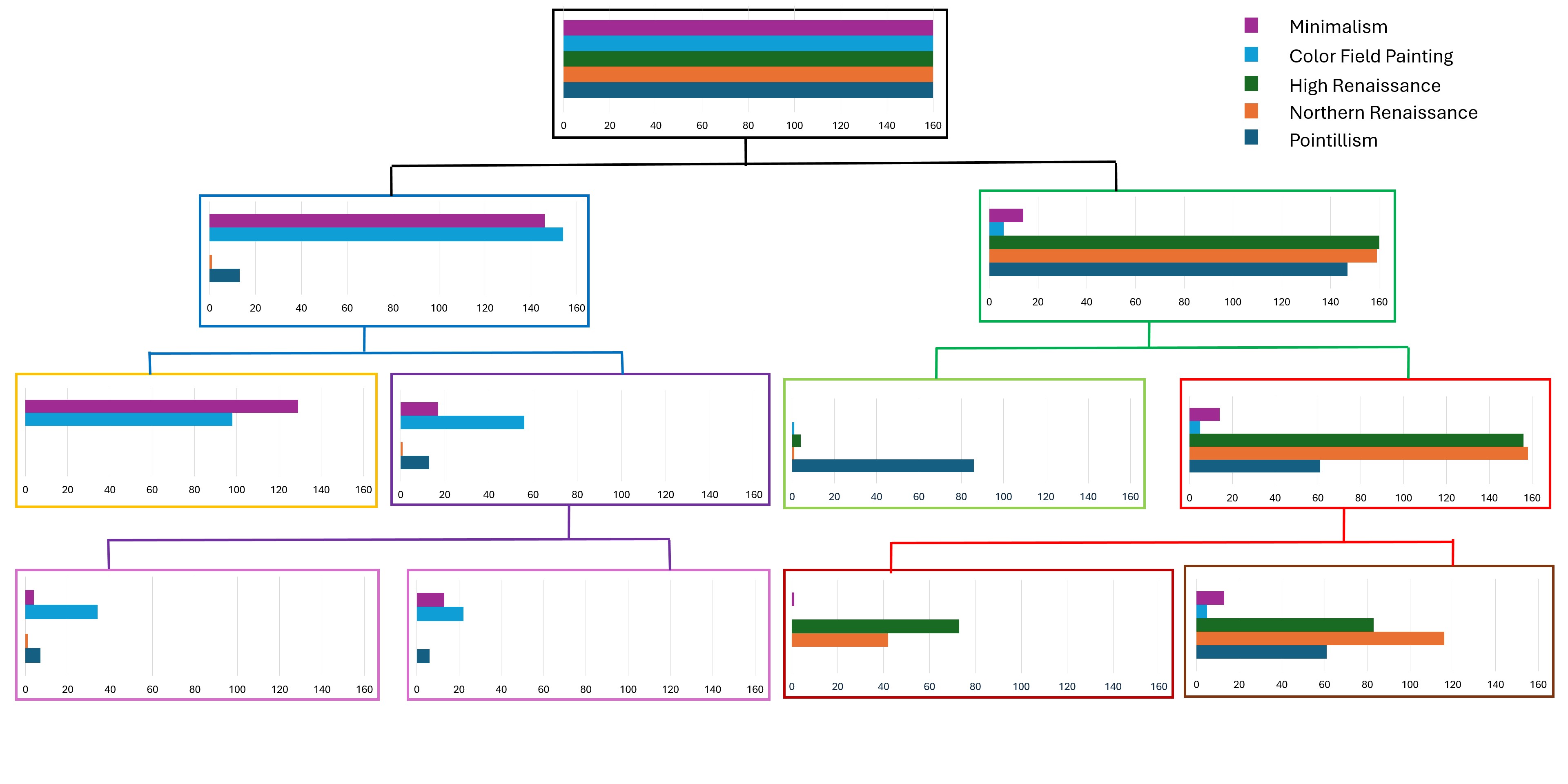}
    \caption{Art Movement distribution Dendrogram}
  \end{subfigure}
  \hfill
   \begin{subfigure}[t]{0.48\textwidth}
    \centering
    \includegraphics[width=\textwidth]{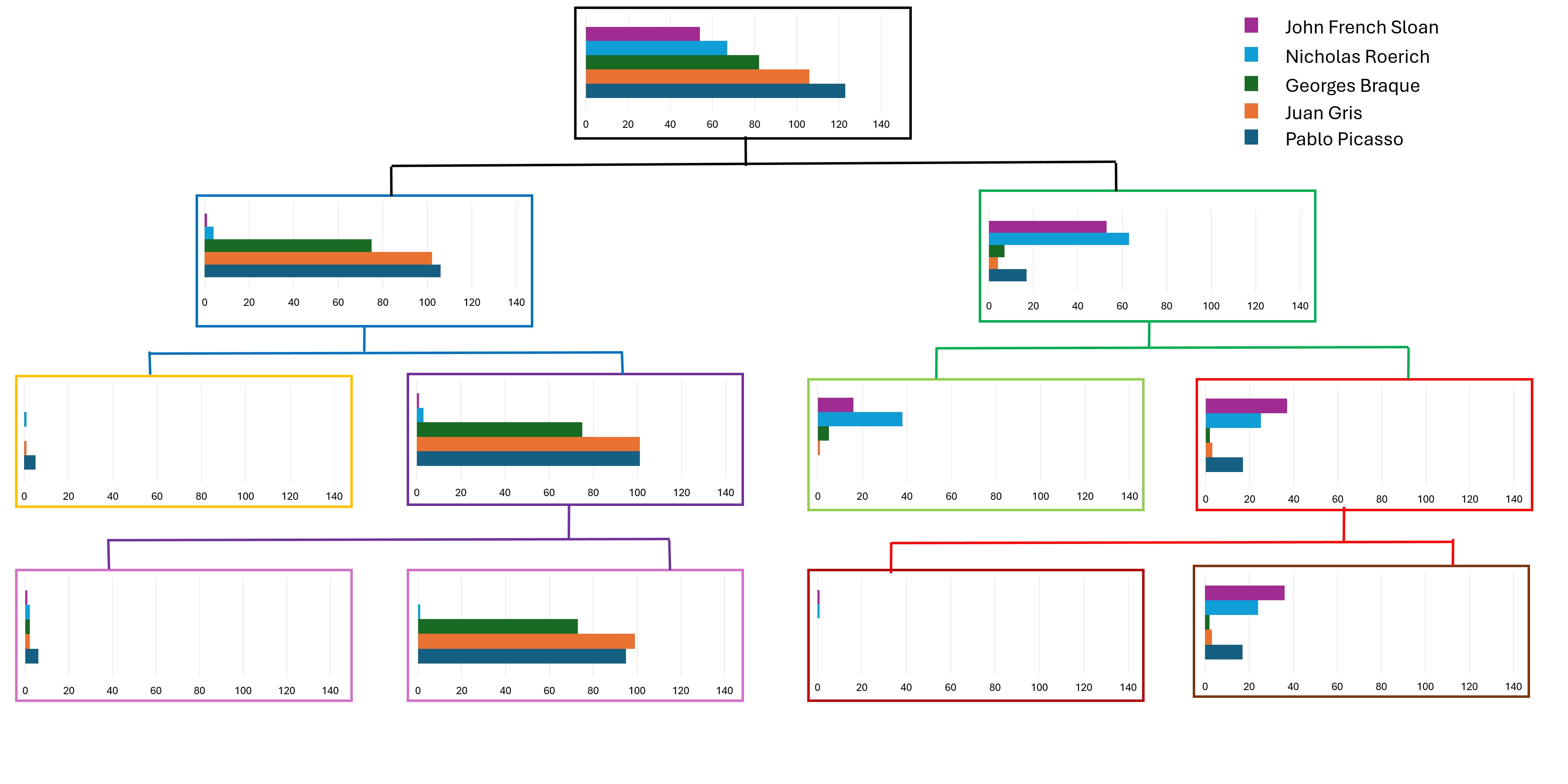}
    \caption{Artist distribution Dendrogram}
  \end{subfigure}
  \begin{subfigure}[t]{0.7\textwidth}
    \centering
    \includegraphics[width=\textwidth]{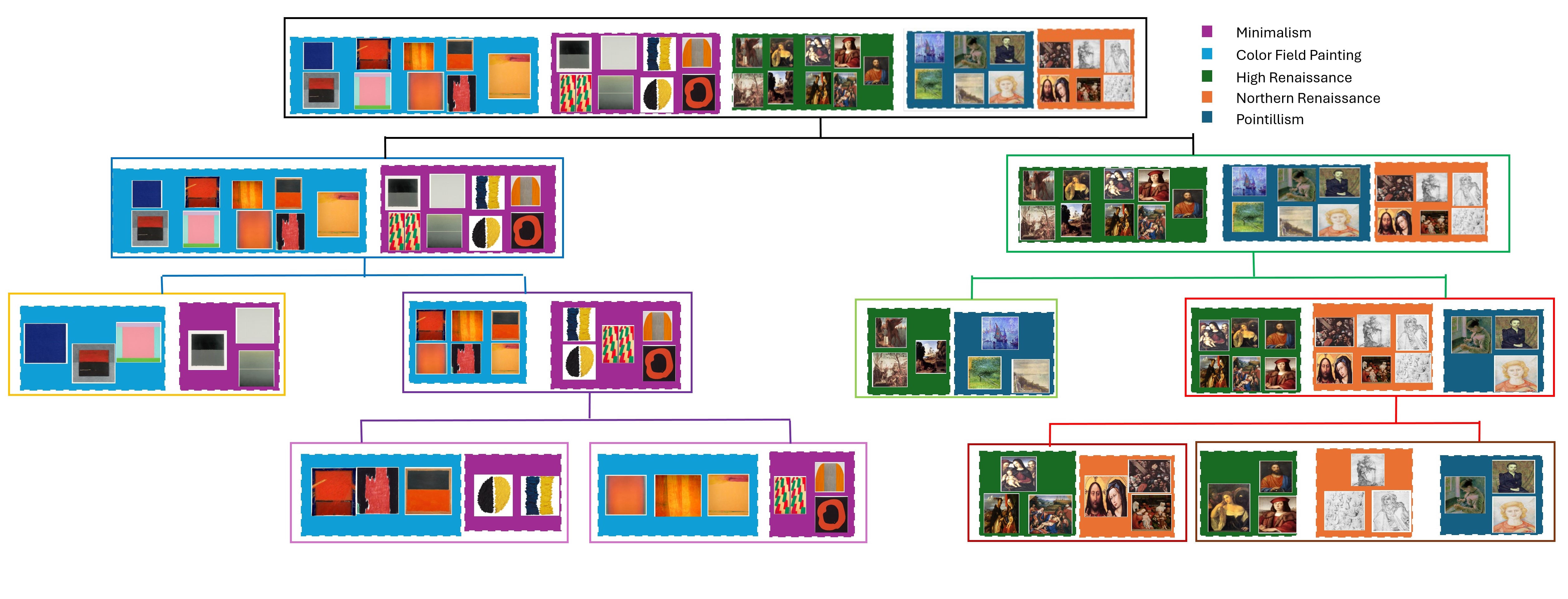}
    \caption{Artworks from each level of the hierarchy based on \textit{art movements}}
  \end{subfigure}
  \hfill
   \begin{subfigure}[t]{0.7\textwidth}
    \centering
    \includegraphics[width=\textwidth]{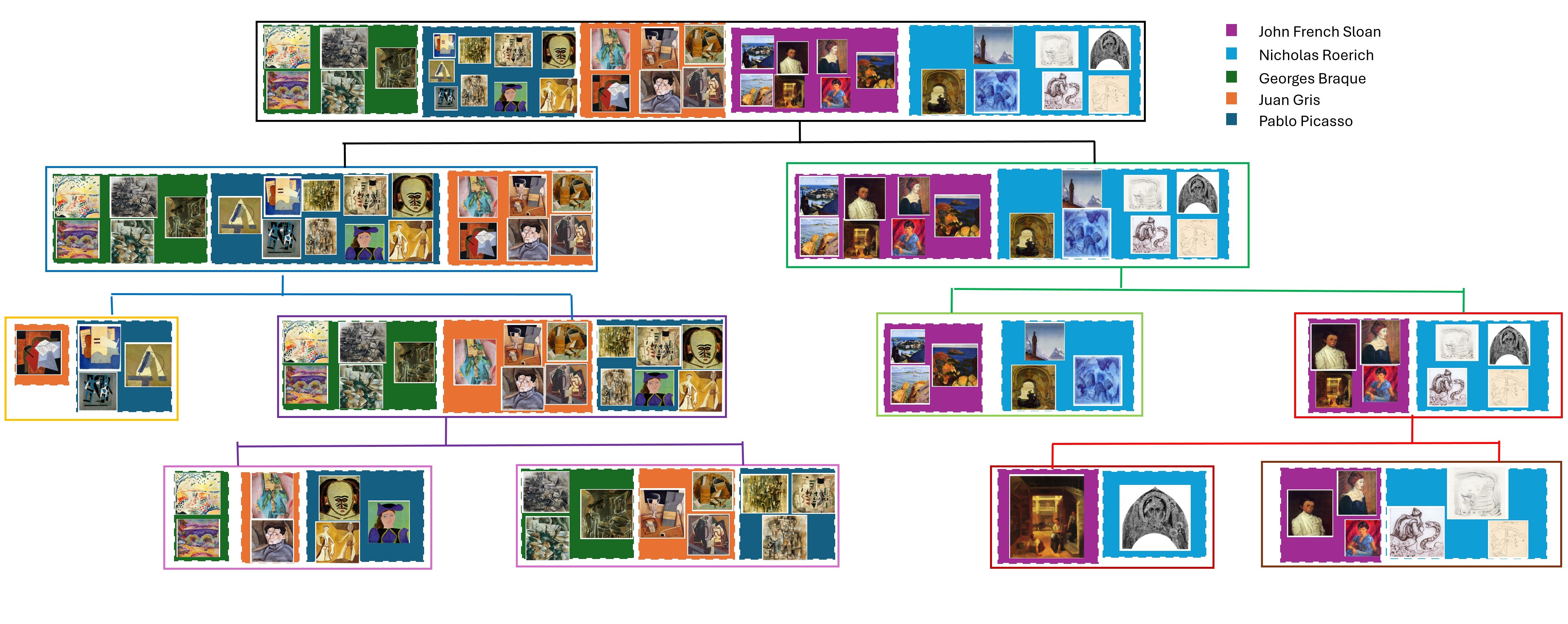}
    \caption{Artworks from each level of the hierarchy based on \textit{artists}}
  \end{subfigure}
  \caption{Hierarchical dendrogram showcasing the distribution of art movements and artists at specific points in the dendrogram. We showcase the art movement and artist dendograms in (b) and (c) respectively. The dendrogram is obtained on the WikiArt-4k-AM (4050 artworks) with 27 art movements and 765 artists with the $F_{StyleCap}$ features. We choose to display the top 5 art movements and top 5 artists in this subset based on the number of artworks. }
  \label{fig:hier_dendro}
\end{figure*}
    
    \clearpage
}


\afterpage{
\clearpage
\begin{figure*}[!t]
 \begin{subfigure}[t]{.85\textwidth}
    \centering
    \includegraphics[width=\textwidth]{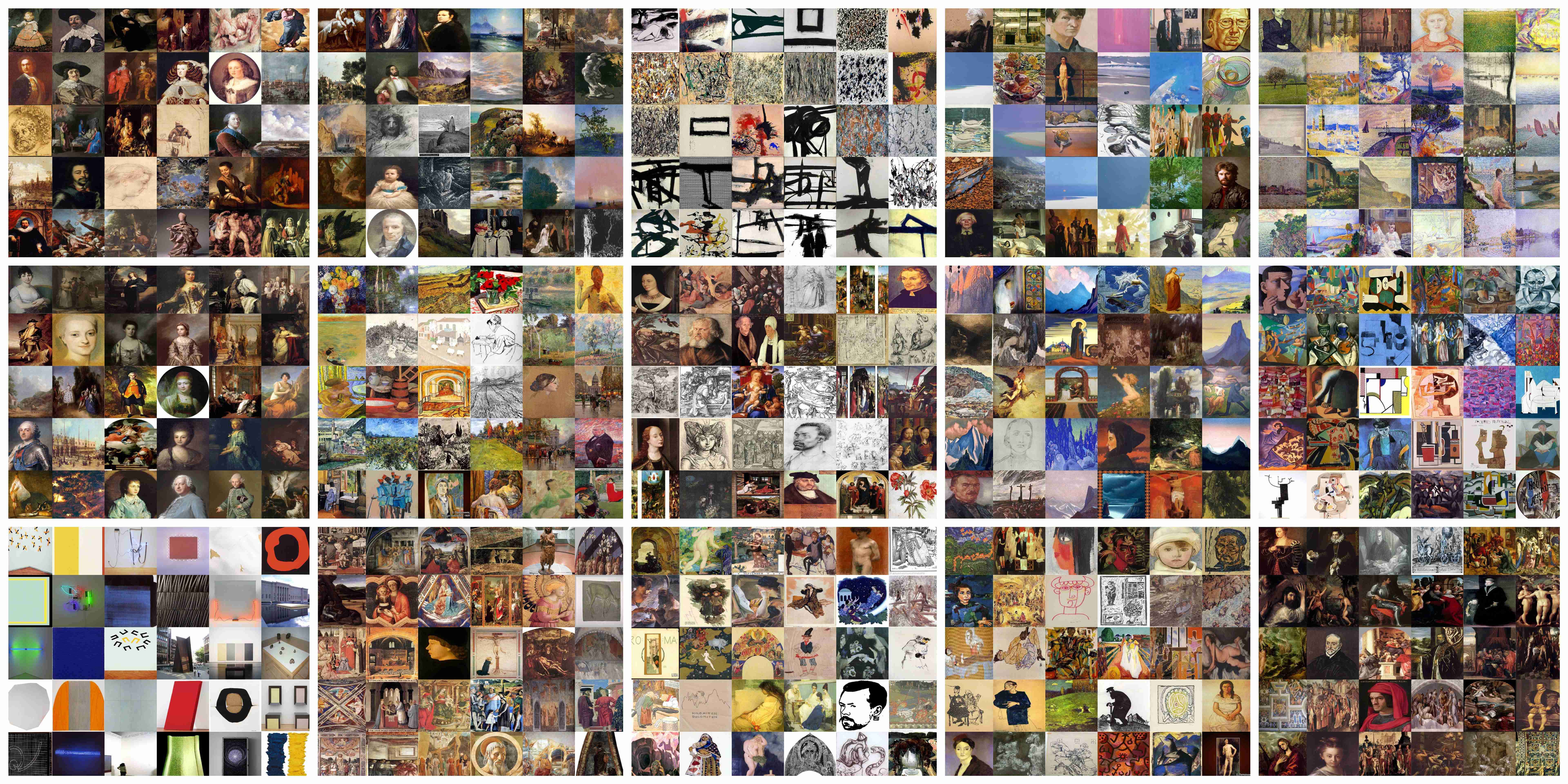}
    \caption{Ground Truth}
  \end{subfigure}
   \hfill
 
  \begin{subfigure}[t]{.4\textwidth}
    \centering
    \includegraphics[width=\textwidth]{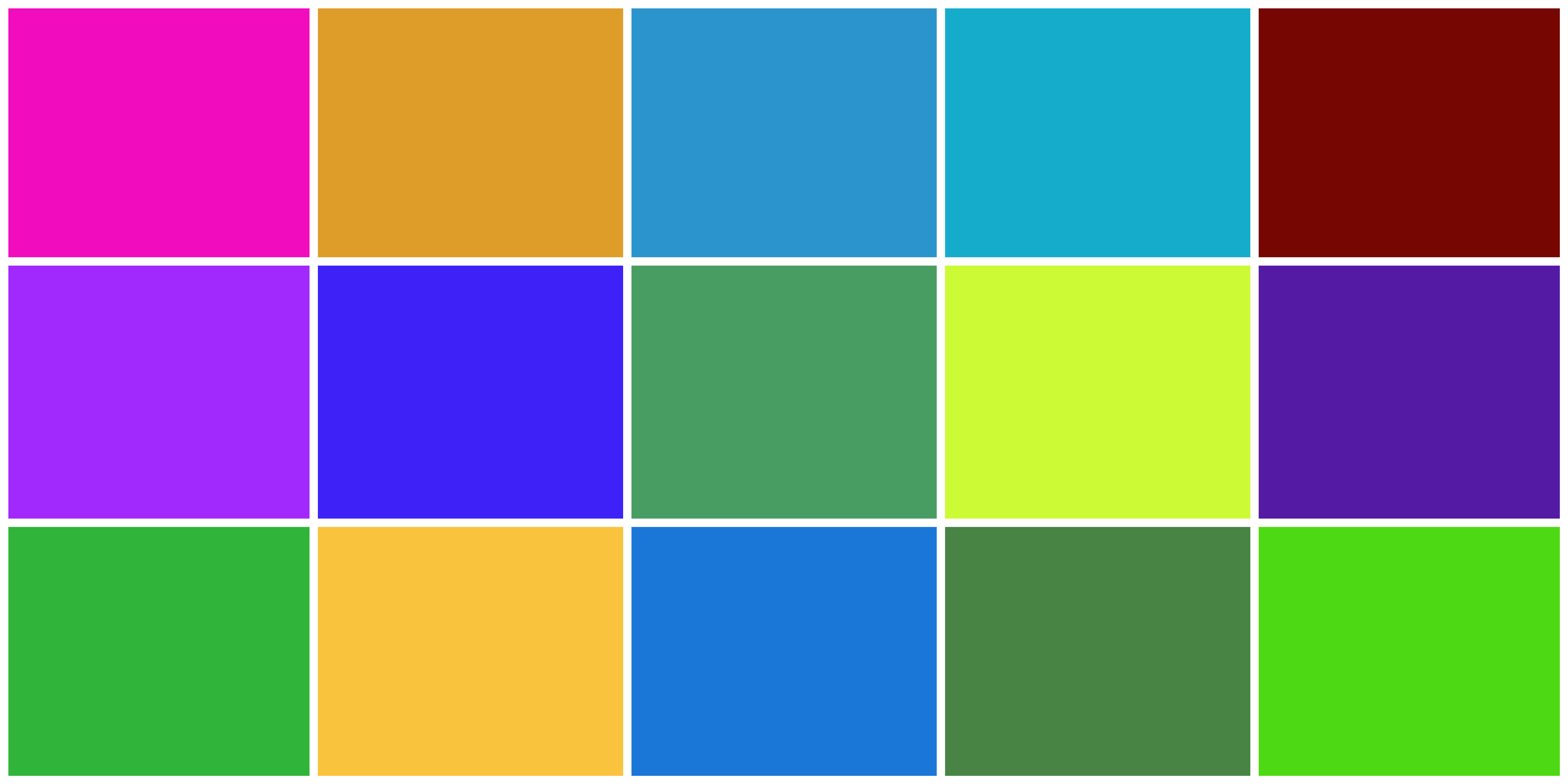}
    \caption{Ground truth color representations per cluster}
  \end{subfigure}
   \hfill
 \begin{subfigure}[t]{.4\textwidth}
    \centering
    \includegraphics[width= .96\textwidth]{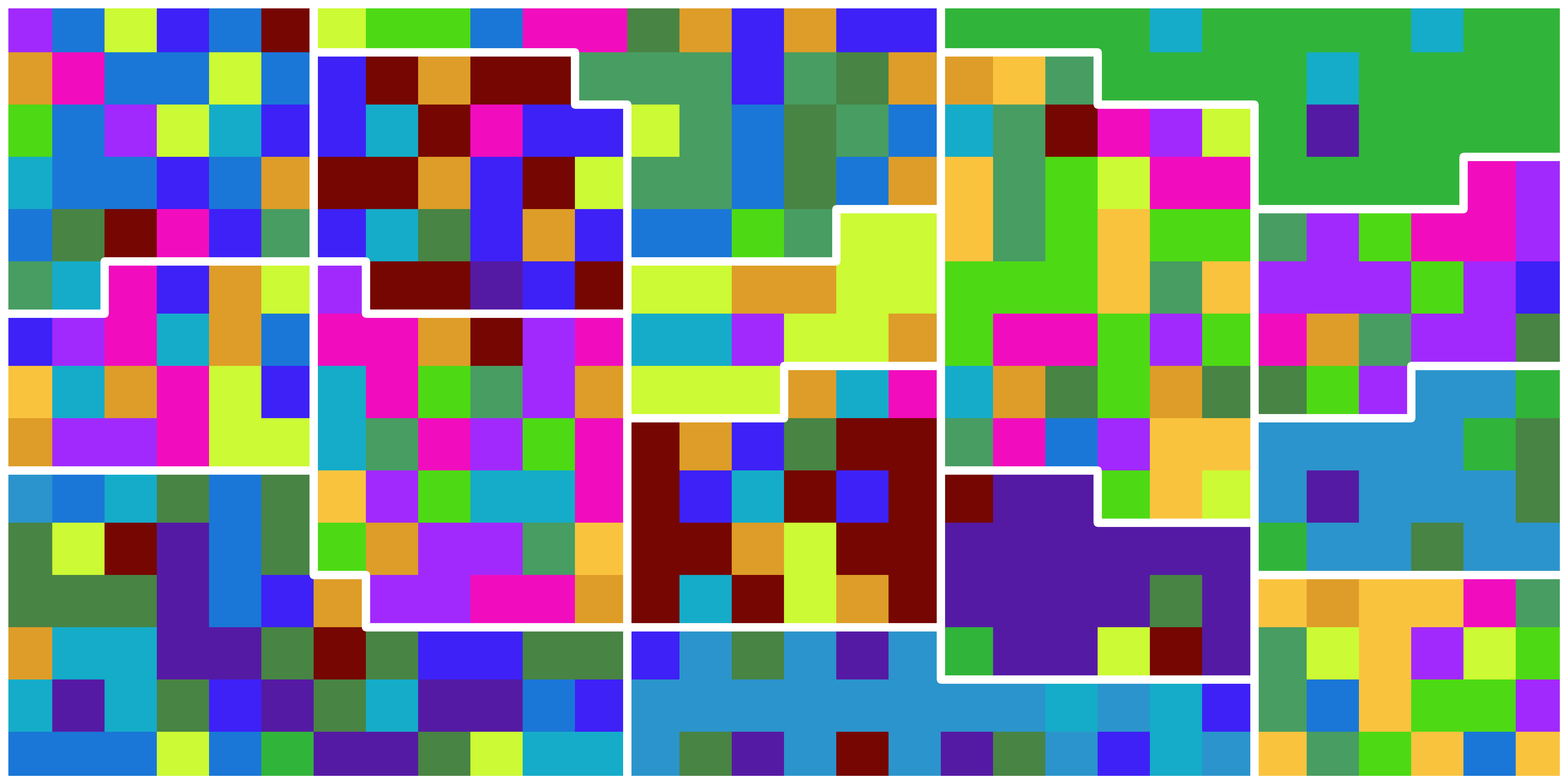}
    \caption {$F_{StyleCap}$ color representations per cluster}
  \end{subfigure}

   \begin{subfigure}[t]{.85\textwidth}
    \centering
    \includegraphics[width=\textwidth]{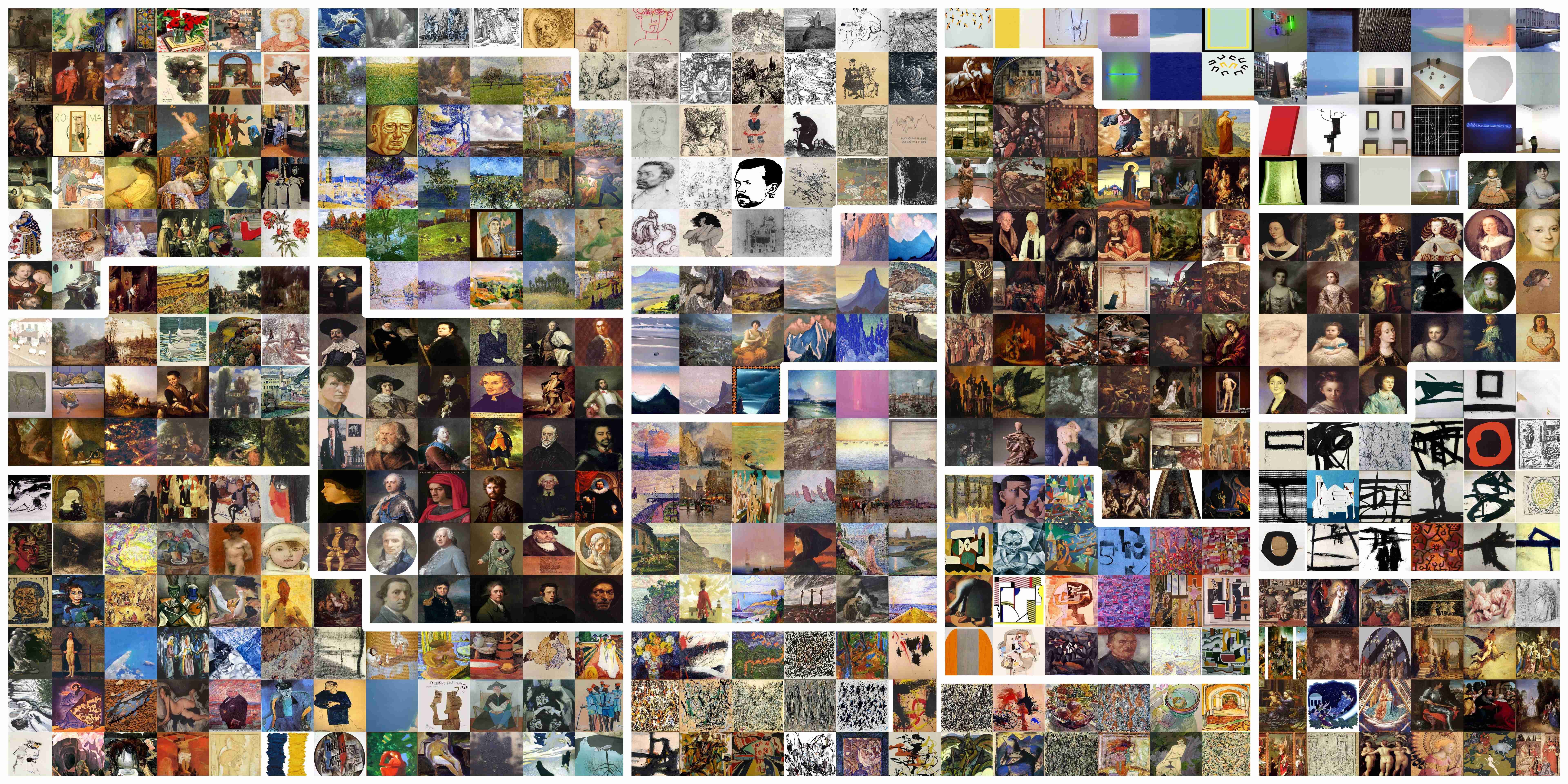}
    \caption {$F_{StyleCap}$ clusters}

  \end{subfigure}
   \caption{Qualitative comparison of 15 clusters for the art movement based ground truth and $F_{StyleCap}$ features on samples from the WikiArt-4k-AM subset. (a) and (d) showcase the artworks present in the two types of clusterings. (b) and (c) show the color representation associated with the ground truth for both clusterings.}
  \label{fig:qual_fig}
\end{figure*}
\clearpage
 }
    When we look at the art movement and artist-based distribution of artworks in Figure \ref{fig:hier_dendro} (a) and Figure \ref{fig:hier_dendro} (b), we observe that at the top level of the dendrogram, the artworks are split distinctly based on the artists and art movements that they belong to.  Moving further down the dendrogram, we observe that the clusters do not split the artworks evenly based on art movement or artist. This indicates that the artworks split based on art movement and artists on the higher levels of the dendrogram. Moving further down, we only observe fine-granular cluster formation based on visual similarity rather than the artist or art movement.

\begin{table}[]
\resizebox{\columnwidth}{!}{%
\begin{tabular}{cc|cc|cc}
\multicolumn{2}{c|}{{\color[HTML]{000000} }}                                                                      & \multicolumn{2}{c|}{{\color[HTML]{000000} \textbf{ARI}}}                                                                         & \multicolumn{2}{c}{{\color[HTML]{000000} \textbf{NMI}}}                                                                        \\ \cline{3-6} 
\multicolumn{2}{c|}{\multirow{-2}{*}{{\color[HTML]{000000} \textbf{Features}}}}                                   & \multicolumn{1}{c|}{{\color[HTML]{000000} \textbf{K-Means}}}              & {\color[HTML]{000000} \textbf{DEC}}                  & \multicolumn{1}{c|}{{\color[HTML]{000000} \textbf{K-Means}}}              & {\color[HTML]{000000} \textbf{DEC}}                  \\ \hline
\multicolumn{1}{c|}{{\color[HTML]{000000} }}                             & {\color[HTML]{000000} $F_{Dense}$}     & \multicolumn{1}{c|}{{\color[HTML]{000000} 0.052}}                         & \cellcolor[HTML]{C0F1C8}{\color[HTML]{000000} 0.059} & \multicolumn{1}{c|}{{\color[HTML]{000000} 0.172}}                         & \cellcolor[HTML]{CAEDFB}{\color[HTML]{000000} 0.177} \\
\multicolumn{1}{c|}{{\color[HTML]{000000} }}                             & {\color[HTML]{000000} $F_{Gram}$}      & \multicolumn{1}{c|}{{\color[HTML]{000000} 0.042}}                         & {\color[HTML]{000000} 0.064}                         & \multicolumn{1}{c|}{{\color[HTML]{000000} 0.13}}                          & {\color[HTML]{000000} 0.151}                         \\
\multicolumn{1}{c|}{\multirow{-3}{*}{{\color[HTML]{000000} $F_{CNN}$}}}  & {\color[HTML]{000000} $F_{g \cdot c}$} & \multicolumn{1}{c|}{{\color[HTML]{000000} 0.012}}                         & {\color[HTML]{000000} 0.012}                         & \multicolumn{1}{c|}{{\color[HTML]{000000} 0.034}}                         & {\color[HTML]{000000} 0.034}                         \\ \hline
\multicolumn{1}{c|}{{\color[HTML]{000000} }}                             & {\color[HTML]{000000} $F_{StyleCap}$}   & \multicolumn{1}{c|}{\cellcolor[HTML]{C0F1C8}{\color[HTML]{000000} 0.071}} & \cellcolor[HTML]{CAEDFB}{\color[HTML]{000000} 0.058} & \multicolumn{1}{c|}{\cellcolor[HTML]{C0F1C8}{\color[HTML]{000000} 0.225}} & \cellcolor[HTML]{C0F1C8}{\color[HTML]{000000} 0.192} \\
\multicolumn{1}{c|}{\multirow{-2}{*}{{\color[HTML]{000000} $F_{Text}$}}} & {\color[HTML]{000000} $F_{Annot}$}     & \multicolumn{1}{c|}{\cellcolor[HTML]{CAEDFB}{\color[HTML]{000000} 0.069}} & {\color[HTML]{000000} 0.043}                         & \multicolumn{1}{c|}{\cellcolor[HTML]{CAEDFB}{\color[HTML]{000000} 0.208}} & {\color[HTML]{000000} 0.162}                         \\ \hline
\multicolumn{1}{c|}{{\color[HTML]{000000} }}                             & {\color[HTML]{000000} $F_{StyleGAN}$}  & \multicolumn{1}{c|}{{\color[HTML]{000000} 0.034}}                         & {\color[HTML]{000000} 0.021}                         & \multicolumn{1}{c|}{{\color[HTML]{000000} 0.103}}                         & {\color[HTML]{000000} 0.095}                         \\
\multicolumn{1}{c|}{{\color[HTML]{000000} }}                             & {\color[HTML]{000000} $F_{Stytr2}$}    & \multicolumn{1}{c|}{{\color[HTML]{000000} 0.021}}                         & {\color[HTML]{000000} 0.03}                          & \multicolumn{1}{c|}{{\color[HTML]{000000} 0.078}}                         & {\color[HTML]{000000} 0.092}                         \\
\multicolumn{1}{c|}{{\color[HTML]{000000} }}                             & {\color[HTML]{000000} $F_{Mamba}$}     & \multicolumn{1}{c|}{{\color[HTML]{000000} 0.034}}                         & {\color[HTML]{000000} 0.029}                         & \multicolumn{1}{c|}{{\color[HTML]{000000} 0.117}}                         & {\color[HTML]{000000} 0.085}                         \\
\multicolumn{1}{c|}{\multirow{-4}{*}{{\color[HTML]{000000} $F_{ST}$}}}   & {\color[HTML]{000000} $F_{StyleShot}$} & \multicolumn{1}{c|}{{\color[HTML]{000000} 0.055}}                         & {\color[HTML]{000000} 0.043}                         & \multicolumn{1}{c|}{{\color[HTML]{000000} 0.161}}                         & {\color[HTML]{000000} 0.151}                         \\ \hline
\multicolumn{1}{c|}{{\color[HTML]{000000} $F_{Train}$}}                 & {\color[HTML]{000000} $F_{CSD}$}       & \multicolumn{1}{c|}{\cellcolor[HTML]{F2CEEF}{\color[HTML]{000000} 0.12}}  & \cellcolor[HTML]{F2CEEF}{\color[HTML]{000000} 0.095} & \multicolumn{1}{c|}{\cellcolor[HTML]{F2CEEF}{\color[HTML]{000000} 0.33}}  & \cellcolor[HTML]{F2CEEF}{\color[HTML]{000000} 0.232}
\end{tabular}%
}
\caption{Quantitative evaluation on the \textbf{WikiArt-79k} for both K-Means and DEC model. The \colorbox[HTML]{F2CEEF}{best}, \colorbox[HTML]{C0F1C8}{second best}, and the \colorbox[HTML]{CAEDFB}{third best} results are highlighted for each metric. The WikiArt-79k subset dataset contains \textbf{79496} artworks with \textbf{27} art movements. The range of values for each evaluation metric are as follows: \textit{ARI}: -1 to 1, \textit{NMI}: 0 to 1.}
\label{tab:wikiart_full}
\end{table}

\begin{table}[]
\centering
\resizebox{\columnwidth}{!}{%
\begin{tabular}{cc|cc|cc}
\multicolumn{2}{c|}{{\color[HTML]{000000} }}                                                                      & \multicolumn{2}{c|}{{\color[HTML]{000000} \textbf{ARI}}}                                                                         & \multicolumn{2}{c}{{\color[HTML]{000000} \textbf{NMI}}}                                                                        \\ \cline{3-6} 
\multicolumn{2}{c|}{\multirow{-2}{*}{{\color[HTML]{000000} \textbf{Features}}}}                                   & \multicolumn{1}{c|}{{\color[HTML]{000000} \textbf{K-Means}}}              & {\color[HTML]{000000} \textbf{DEC}}                  & \multicolumn{1}{c|}{{\color[HTML]{000000} \textbf{K-Means}}}              & {\color[HTML]{000000} \textbf{DEC}}                  \\ \hline
\multicolumn{1}{c|}{{\color[HTML]{000000} }}                             & {\color[HTML]{000000} $F_{Dense}$}     & \multicolumn{1}{c|}{{\color[HTML]{000000} 0.0002}}                        & {\color[HTML]{000000} 0.0001}                        & \multicolumn{1}{c|}{{\color[HTML]{000000} 0.008}}                         & {\color[HTML]{000000} 0.008}                         \\
\multicolumn{1}{c|}{{\color[HTML]{000000} }}                             & {\color[HTML]{000000} $F_{Gram}$}      & \multicolumn{1}{c|}{{\color[HTML]{000000} 0.0003}}                        & {\color[HTML]{000000} 0.0002}                        & \multicolumn{1}{c|}{{\color[HTML]{000000} 0.009}}                         & {\color[HTML]{000000} 0.008}                         \\
\multicolumn{1}{c|}{\multirow{-3}{*}{{\color[HTML]{000000} $F_{CNN}$}}}  & {\color[HTML]{000000} $F_{g \cdot c}$} & \multicolumn{1}{c|}{{\color[HTML]{000000} 0.0002}}                        & {\color[HTML]{000000} 0.0001}                        & \multicolumn{1}{c|}{{\color[HTML]{000000} 0.008}}                         & {\color[HTML]{000000} 0.008}                         \\ \hline
\multicolumn{1}{c|}{{\color[HTML]{000000} }}                             & {\color[HTML]{000000} $F_{StyleCap}$}   & \multicolumn{1}{c|}{{\color[HTML]{000000} 0.0008}}                        & {\color[HTML]{000000} -0.0001}                       & \multicolumn{1}{c|}{{\color[HTML]{000000} 0.007}}                         & {\color[HTML]{000000} 0.008}                         \\
\multicolumn{1}{c|}{\multirow{-2}{*}{{\color[HTML]{000000} $F_{Text}$}}} & {\color[HTML]{000000} $F_{Annot}$}     & \multicolumn{1}{c|}{{\color[HTML]{000000} 0.0006}}                        & {\color[HTML]{000000} 0.052}                         & \multicolumn{1}{c|}{{\color[HTML]{000000} 0.008}}                         & {\color[HTML]{000000} 0.197}                         \\ \hline
\multicolumn{1}{c|}{{\color[HTML]{000000} }}                             & {\color[HTML]{000000} $F_{StyleGAN}$}  & \multicolumn{1}{c|}{\cellcolor[HTML]{CAEDFB}{\color[HTML]{000000} 0.069}} & \cellcolor[HTML]{CAEDFB}{\color[HTML]{000000} 0.067} & \multicolumn{1}{c|}{\cellcolor[HTML]{CAEDFB}{\color[HTML]{000000} 0.214}} & \cellcolor[HTML]{CAEDFB}{\color[HTML]{000000} 0.204} \\
\multicolumn{1}{c|}{{\color[HTML]{000000} }}                             & {\color[HTML]{000000} $F_{Stytr2}$}    & \multicolumn{1}{c|}{{\color[HTML]{000000} 0.0017}}                        & {\color[HTML]{000000} 0.0002}                        & \multicolumn{1}{c|}{{\color[HTML]{000000} 0.008}}                         & {\color[HTML]{000000} 0.007}                         \\
\multicolumn{1}{c|}{{\color[HTML]{000000} }}                             & {\color[HTML]{000000} $F_{Mamba}$}     & \multicolumn{1}{c|}{\cellcolor[HTML]{C0F1C8}{\color[HTML]{000000} 0.075}} & \cellcolor[HTML]{C0F1C8}{\color[HTML]{000000} 0.095} & \multicolumn{1}{c|}{\cellcolor[HTML]{C0F1C8}{\color[HTML]{000000} 0.239}} & \cellcolor[HTML]{C0F1C8}{\color[HTML]{000000} 0.229} \\
\multicolumn{1}{c|}{\multirow{-4}{*}{{\color[HTML]{000000} $F_{ST}$}}}   & {\color[HTML]{000000} $F_{Styleshot}$} & \multicolumn{1}{c|}{{\color[HTML]{000000} 0.0005}}                        & {\color[HTML]{000000} 0.0009}                        & \multicolumn{1}{c|}{{\color[HTML]{000000} 0.008}}                         & {\color[HTML]{000000} 0.008}                         \\ \hline
\multicolumn{1}{c|}{{\color[HTML]{000000} $F_{Train}$}}                 & {\color[HTML]{000000} $F_{CSD}$}       & \multicolumn{1}{c|}{\cellcolor[HTML]{F2CEEF}{\color[HTML]{000000} 0.31}}  & \cellcolor[HTML]{F2CEEF}{\color[HTML]{000000} 0.262} & \multicolumn{1}{c|}{\cellcolor[HTML]{F2CEEF}{\color[HTML]{000000} 0.51}}  & \cellcolor[HTML]{F2CEEF}{\color[HTML]{000000} 0.461}
\end{tabular}%
}
\caption{Quantitative evaluation on the \textbf{WikiArt-25k-Artist} with the \textbf{top 40 artists} for both K-Means and DEC model. The \colorbox[HTML]{F2CEEF}{best}, \colorbox[HTML]{C0F1C8}{second best}, and the \colorbox[HTML]{CAEDFB}{third best} results are highlighted for each metric. The WikiArt-25k subset dataset with top 40 artists contains \textbf{25,550} artworks from the artists with the highest amount of artworks. The range of values for each evaluation metric are as follows: \textit{ARI}: -1 to 1, \textit{NMI}: 0 to 1.}
\label{tab:wiki_full_artists}
\end{table}

\begin{table}[]
\centering
\resizebox{\columnwidth}{!}{%
\begin{tabular}{cc|cc|cc}
\multicolumn{2}{c|}{{\color[HTML]{000000} }}                                                                       & \multicolumn{2}{c|}{{\color[HTML]{000000} \textbf{ARI}}}                                                                         & \multicolumn{2}{c}{{\color[HTML]{000000} \textbf{NMI}}}                                                                          \\ \cline{3-6} 
\multicolumn{2}{c|}{\multirow{-2}{*}{{\color[HTML]{000000} \textbf{Features}}}}                                    & \multicolumn{1}{c|}{{\color[HTML]{000000} \textbf{K-Means}}}              & {\color[HTML]{000000} \textbf{DEC}}                  & \multicolumn{1}{c|}{{\color[HTML]{000000} \textbf{K-Means}}}              & {\color[HTML]{000000} \textbf{DEC}}                  \\ \hline
\multicolumn{1}{c|}{{\color[HTML]{000000} }}                              & {\color[HTML]{000000} $F_{Dense}$}     & \multicolumn{1}{c|}{{\color[HTML]{000000} 0.007}}                         & {\color[HTML]{000000} 0.024}                         & \multicolumn{1}{c|}{{\color[HTML]{000000} 0.152}}                         & {\color[HTML]{000000} 0.167}                         \\
\multicolumn{1}{c|}{{\color[HTML]{000000} }}                              & {\color[HTML]{000000} $F_{Gram}$}      & \multicolumn{1}{c|}{{\color[HTML]{000000} 0.051}}                         & {\color[HTML]{000000} 0.077}                         & \multicolumn{1}{c|}{{\color[HTML]{000000} 0.191}}                         & {\color[HTML]{000000} 0.251}                         \\
\multicolumn{1}{c|}{\multirow{-3}{*}{{\color[HTML]{000000} $F_{CNN}$}}}   & {\color[HTML]{000000} $F_{g \cdot c}$} & \multicolumn{1}{c|}{{\color[HTML]{000000} 0.01}}                          & {\color[HTML]{000000} 0.009}                         & \multicolumn{1}{c|}{{\color[HTML]{000000} 0.069}}                         & {\color[HTML]{000000} 0.067}                         \\ \hline
\multicolumn{1}{c|}{{\color[HTML]{000000} }}                              & {\color[HTML]{000000} $F_{StyleCap}$}   & \multicolumn{1}{c|}{\cellcolor[HTML]{CAEDFB}{\color[HTML]{000000} 0.123}} & \cellcolor[HTML]{CAEDFB}{\color[HTML]{000000} 0.136} & \multicolumn{1}{c|}{\cellcolor[HTML]{CAEDFB}{\color[HTML]{000000} 0.326}} & \cellcolor[HTML]{CAEDFB}{\color[HTML]{000000} 0.32}  \\
\multicolumn{1}{c|}{\multirow{-2}{*}{{\color[HTML]{000000} $F_{Text}$}}}  & {\color[HTML]{000000} $F_{Annot}$}     & \multicolumn{1}{c|}{{\color[HTML]{000000} 0.095}}                         & {\color[HTML]{000000} 0.095}                         & \multicolumn{1}{c|}{{\color[HTML]{000000} 0.284}}                         & {\color[HTML]{000000} 0.268}                         \\ \hline
\multicolumn{1}{c|}{{\color[HTML]{000000} }}                              & {\color[HTML]{000000} $F_{StyleGAN}$}  & \multicolumn{1}{c|}{{\color[HTML]{000000} 0.046}}                         & {\color[HTML]{000000} 0.049}                         & \multicolumn{1}{c|}{{\color[HTML]{000000} 0.152}}                         & {\color[HTML]{000000} 0.145}                         \\
\multicolumn{1}{c|}{{\color[HTML]{000000} }}                              & {\color[HTML]{000000} $F_{Stytr2}$}    & \multicolumn{1}{c|}{{\color[HTML]{000000} 0.029}}                         & {\color[HTML]{000000} 0.033}                         & \multicolumn{1}{c|}{{\color[HTML]{000000} 0.131}}                         & {\color[HTML]{000000} 0.122}                         \\
\multicolumn{1}{c|}{{\color[HTML]{000000} }}                              & {\color[HTML]{000000} $F_{Mamba}$}     & \multicolumn{1}{c|}{{\color[HTML]{000000} 0.053}}                         & {\color[HTML]{000000} 0.075}                         & \multicolumn{1}{c|}{{\color[HTML]{000000} 0.185}}                         & {\color[HTML]{000000} 0.198}                         \\
\multicolumn{1}{c|}{\multirow{-4}{*}{{\color[HTML]{000000} $F_{ST}$}}}    & {\color[HTML]{000000} $F_{StyleShot}$} & \multicolumn{1}{c|}{{\color[HTML]{000000} 0.103}}                         & {\color[HTML]{000000} 0.087}                         & \multicolumn{1}{c|}{{\color[HTML]{000000} 0.248}}                         & {\color[HTML]{000000} 0.23}                          \\ \hline
\multicolumn{1}{c|}{{\color[HTML]{000000} }}                              & {\color[HTML]{000000} $F_{CSD}$}       & \multicolumn{1}{c|}{\cellcolor[HTML]{C0F1C8}{\color[HTML]{000000} 0.25}}  & \cellcolor[HTML]{F2CEEF}{\color[HTML]{000000} 0.219} & \multicolumn{1}{c|}{\cellcolor[HTML]{F2CEEF}{\color[HTML]{000000} 0.47}}  & \cellcolor[HTML]{F2CEEF}{\color[HTML]{000000} 0.419} \\
\multicolumn{1}{c|}{\multirow{-2}{*}{{\color[HTML]{000000} $F_{Train}$}}} & {\color[HTML]{000000} $F_{ArtMove}$}   & \multicolumn{1}{c|}{\cellcolor[HTML]{F2CEEF}0.27}                         & \cellcolor[HTML]{C0F1C8}0.161                        & \multicolumn{1}{c|}{\cellcolor[HTML]{C0F1C8}0.45}                         & \cellcolor[HTML]{C0F1C8}0.325                       
\end{tabular}%
}
\caption{Quantitative evaluation on the \textbf{WikiArt-4k-AM} dataset for both K-Means and DEC model. The \colorbox[HTML]{F2CEEF}{best}, \colorbox[HTML]{C0F1C8}{second best}, and the \colorbox[HTML]{CAEDFB}{third best} results are highlighted for each metric. The WikiArt-4k-AM dataset contains \textbf{4050} artworks with \textbf{27} art movements. The range of values for each evaluation metric are as follows: \textit{ARI}: -1 to 1, \textit{NMI}: 0 to 1.}
\label{tab:wiki_quant_4k}
\end{table}

\begin{table}[]
\centering
\resizebox{\columnwidth}{!}{%
\begin{tabular}{cc|cc|cc}
\multicolumn{2}{c|}{{\color[HTML]{000000} }}                                                                       & \multicolumn{2}{c|}{{\color[HTML]{000000} \textbf{ARI}}}                                                                        & \multicolumn{2}{c}{{\color[HTML]{000000} \textbf{NMI}}}                                                                        \\ \cline{3-6} 
\multicolumn{2}{c|}{\multirow{-2}{*}{{\color[HTML]{000000} \textbf{Features}}}}                                    & \multicolumn{1}{c|}{{\color[HTML]{000000} \textbf{K-Means}}}             & {\color[HTML]{000000} \textbf{DEC}}                  & \multicolumn{1}{c|}{{\color[HTML]{000000} \textbf{K-Means}}}             & {\color[HTML]{000000} \textbf{DEC}}                 \\ \hline
\multicolumn{1}{c|}{{\color[HTML]{000000} }}                              & {\color[HTML]{000000} $F_{Dense}$}     & \multicolumn{1}{c|}{{\color[HTML]{000000} 0.16}}                         & {\color[HTML]{000000} 0.137}                         & \multicolumn{1}{c|}{{\color[HTML]{000000} 0.39}}                         & {\color[HTML]{000000} 0.378}                        \\
\multicolumn{1}{c|}{{\color[HTML]{000000} }}                              & {\color[HTML]{000000} $F_{Gram}$}      & \multicolumn{1}{c|}{{\color[HTML]{000000} 0.08}}                         & {\color[HTML]{000000} 0.109}                         & \multicolumn{1}{c|}{{\color[HTML]{000000} 0.32}}                         & {\color[HTML]{000000} 0.368}                        \\
\multicolumn{1}{c|}{\multirow{-3}{*}{{\color[HTML]{000000} $F_{CNN}$}}}   & {\color[HTML]{000000} $F_{g \cdot c}$} & \multicolumn{1}{c|}{{\color[HTML]{000000} 0.01}}                         & {\color[HTML]{000000} 0.01}                          & \multicolumn{1}{c|}{{\color[HTML]{000000} 0.1}}                          & {\color[HTML]{000000} 0.108}                        \\ \hline
\multicolumn{1}{c|}{{\color[HTML]{000000} }}                              & {\color[HTML]{000000} $F_{StyleCap}$}   & \multicolumn{1}{c|}{{\color[HTML]{000000} 0.14}}                         & {\color[HTML]{000000} 0.143}                         & \multicolumn{1}{c|}{{\color[HTML]{000000} 0.37}}                         & {\color[HTML]{000000} 0.371}                        \\
\multicolumn{1}{c|}{\multirow{-2}{*}{{\color[HTML]{000000} $F_{Text}$}}}  & {\color[HTML]{000000} $F_{Annot}$}     & \multicolumn{1}{c|}{{\color[HTML]{000000} 0.09}}                         & {\color[HTML]{000000} 0.074}                         & \multicolumn{1}{c|}{{\color[HTML]{000000} 0.31}}                         & {\color[HTML]{000000} 0.283}                        \\ \hline
\multicolumn{1}{c|}{{\color[HTML]{000000} }}                              & {\color[HTML]{000000} $F_{StyleGAN}$}  & \multicolumn{1}{c|}{{\color[HTML]{000000} 0.06}}                         & {\color[HTML]{000000} 0.063}                         & \multicolumn{1}{c|}{{\color[HTML]{000000} 0.24}}                         & {\color[HTML]{000000} 0.247}                        \\
\multicolumn{1}{c|}{{\color[HTML]{000000} }}                              & {\color[HTML]{000000} $F_{Stytr2}$}    & \multicolumn{1}{c|}{{\color[HTML]{000000} 0.05}}                         & {\color[HTML]{000000} 0.047}                         & \multicolumn{1}{c|}{{\color[HTML]{000000} 0.23}}                         & {\color[HTML]{000000} 0.196}                        \\
\multicolumn{1}{c|}{{\color[HTML]{000000} }}                              & {\color[HTML]{000000} $F_{Mamba}$}     & \multicolumn{1}{c|}{{\color[HTML]{000000} 0.08}}                         & {\color[HTML]{000000} 0.082}                         & \multicolumn{1}{c|}{{\color[HTML]{000000} 0.3}}                          & {\color[HTML]{000000} 0.271}                        \\
\multicolumn{1}{c|}{\multirow{-4}{*}{{\color[HTML]{000000} $F_{st}$}}}    & {\color[HTML]{000000} $F_{StyleShot}$} & \multicolumn{1}{c|}{\cellcolor[HTML]{CAEDFB}{\color[HTML]{000000} 0.18}} & \cellcolor[HTML]{CAEDFB}{\color[HTML]{000000} 0.185} & \multicolumn{1}{c|}{\cellcolor[HTML]{CAEDFB}{\color[HTML]{000000} 0.41}} & \cellcolor[HTML]{CAEDFB}{\color[HTML]{000000} 0.42} \\ \hline
\multicolumn{1}{c|}{{\color[HTML]{000000} }}                              & {\color[HTML]{000000} $F_{CSD}$}       & \multicolumn{1}{c|}{\cellcolor[HTML]{C0F1C8}{\color[HTML]{000000} 0.35}} & \cellcolor[HTML]{C0F1C8}{\color[HTML]{000000} 0.282} & \multicolumn{1}{c|}{\cellcolor[HTML]{C0F1C8}{\color[HTML]{000000} 0.59}} & \cellcolor[HTML]{C0F1C8}{\color[HTML]{000000} 0.53} \\
\multicolumn{1}{c|}{\multirow{-2}{*}{{\color[HTML]{000000} $F_{Train}$}}} & {\color[HTML]{000000} $F_{Artist}$}   & \multicolumn{1}{c|}{\cellcolor[HTML]{F2CEEF}0.54}                        & \cellcolor[HTML]{F2CEEF}0.471                        & \multicolumn{1}{c|}{\cellcolor[HTML]{F2CEEF}0.68}                        & \cellcolor[HTML]{F2CEEF}0.634                      
\end{tabular}%
}
\caption{Quantitative evaluation on the \textbf{Wikiart-4k-Artist} with the \textbf{top 40 artists} for both K-Means and DEC model. The \colorbox[HTML]{F2CEEF}{best}, \colorbox[HTML]{C0F1C8}{second best}, and the \colorbox[HTML]{CAEDFB}{third best} results are highlighted for each metric. The Wikiart-4k-Artist subset dataset with top 40 artists contains \textbf{4000} artworks from the artists with the highest amount of artworks. The range of values for each evaluation metric are as follows: \textit{ARI}: -1 to 1, \textit{NMI}: 0 to 1.}
\label{tab:wiki_quant_artists}
\end{table}

\subsection{Does the definition of Style matter?}
\textbf{TL; DR:}\textit{ Absolutely; the effectiveness of neural style representations for style-based clustering is highly dependent on the definition of style.}

When we experiment with the art movements included in the WikiArt dataset and consider them as ground truth labels, we observe that the the clusters formed with $F_{ST}$ as well as $F_{Gram}$ do not adhere to the art movements ground truth as shown in Table \ref{tab:wiki_quant_4k}. We also observe that the $F_{Text}$ features perform better in comparison to the other features on this ground truth. This indicates that when we consider the art movements as the style definition for clustering, we get better clustering results with $F_{Text}$ and, when we consider the visual style similarity (such as texture, color, and nature of drawn lines) as the style definition, the $F_{ST}$ features perform the best. This further indicates that the style-based clustering is dependent on the style definition.

The same can be observed when we cluster the artworks based on artists. We observe that again the $F_{Text}$ features can provide better results apart from $F_{StyleShot}$ as shown in Table \ref{tab:wiki_quant_artists}. This further indicates that $F_{Text}$ provides a more malleable style representation that performs moderately well across different style definitions.

We experiment further to verify whether fine-tuning a model on the data that is specific to a style definition helps in learning features that adhere to that particular definition. We verify this on Wikiart's artist and art movement-based style definitions. The results shown in tables \ref{tab:wiki_quant_4k} and \ref{tab:wiki_quant_artists} are in favor of both the definitions with the $F_{Artist}$ features significantly outperforming the rest on the artist-based data while $F_{ArtMove}$ features are comparable to the best-performing features, $F_{CSD}$, on art movement data. This shows that a little fine-tuning on different style definitions can easily align the features to represent those definitions well and in turn form clusters adhering to those definitions. 

\begin{figure}[t!]
     \centering
     \includegraphics[width=\linewidth]{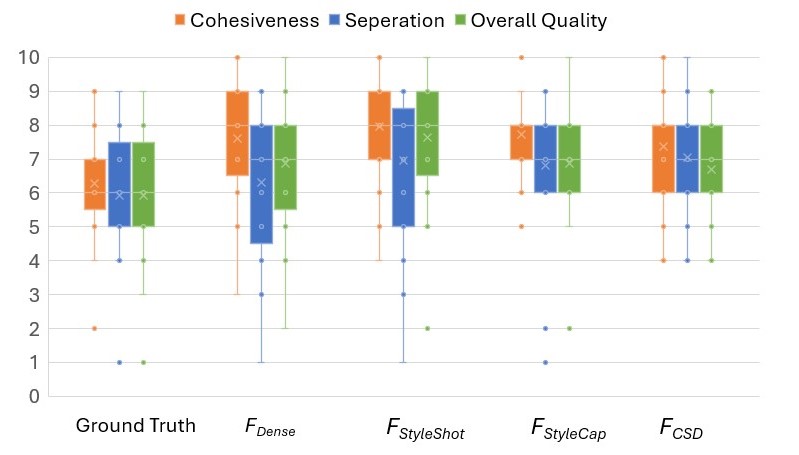}
  
  \caption{Box plots for the survey conducted on the clusters obtained from 450 samples from the WikiArt-4k-AM subset. Clusters were obtained on the subset using $F_{dense}$, $F_{caption}$, $F_{styleshot}$ and $F_{CSD}$ features and K-Means clustering model. The survey was conducted on the art movement ground truth as well. The survey included questions relating to cluster cohesion, cluster separation, and overall clustering quality. Overall, 25 participants responded to the survey.}
  \label{fig:surv_fig}
\end{figure}

Our Human Survey on the clustering quality showed the preference for particular styles (refer to Figure \ref{fig:surv_fig}) among the participants. Please note that we also put the Ground Truth (WikiArt Art Movement) as one of the survey examples. However, the survey results show that the survey participants did not consider the ground truth to be a good example of style clustering. Studies to understand the human perception-driven definitions of style promises to be an interesting area for future studies.

\subsection{Limitations}

We point to the two main limitations of our study:

\begin{enumerate}

\item For ground truth, we have relied on two sources: (1) Our curated dataset, and (2) WikiArt classification. While (1) does style classification based on styles in individual representative style images, which primarily focuses on the definition of style which is influenced by texture, color, shape, etc., the latter classifies on the basis of art movements where the definition of style is coarse and has not been carefully applied. Our technique could extend to other datasets such as Fashion and Architecture, etc.

 \item We rely on the LLMs for textual style descriptions. The ability to get accurate results depends upon the ability of the LLMs to interpret the visual information for style. However, we consider that ours is still an important experiment and study as it sets the benchmark for future performance of LLM-based style analysis or even human-curated style descriptions.
\end{enumerate}

 \section{Conclusion}
In this paper, we propose and explore style-based clustering of visual artworks. We devise visual artwork style representations from the deep neural features in classification networks, generative style transfer models, and large vision-language-based annotations. We find that these representations greatly enhance the capability of clustering methods for targeting artistic styles. 

Machine-based or aided development of the notion of style and embodying it in a representation continues to be an exciting field of research. We feel that style-based clustering is an essential tool in this discovery. We found that it is possible to train for new style definitions by example-driven methods. Large Visio-Language models (VLMs) give another opportunity for style learning and explication. However, a VLM's notion of style depends upon the notions we impart to them through training. Our present experience shows that these models perform reasonably well on direct style captioning but have to evolve more to have a better attribute (or taxonomy)-based understanding of style notions. 

On a more fundamental level our approach questions, in a clustering scenario, \textit{the notion of style}. We believe that `Style' is a social and cultural construct and every human has some implicit notion of it. Most of the neural representations, particularly in the style transfer realm, tend towards a notion that tries to separate the style from the content and hence tends to focus, though implicitly, on features such as colors, texture, medium, and line stroke properties. Though limited in scope, in our experiment with human evaluation of clustering, we get a hint that even humans, particularly non-trained artists, tend to develop a similar notion of style in visual artworks.

Finally, our work also opens up several compelling application and research questions about artwork clustering based on artistic style. For example: (i) the approach opens up discovering finer-grained style concepts within a corpus of work in an unsupervised manner and helps the artists and museums digitally catalog and monetize their styles, and (ii) it also helps explore and create the art style evolution-based narratives and curative practices for an artwork collection.

\balance
\bibliographystyle{ACM-Reference-Format}
\bibliography{sample-base}

\clearpage
\appendix

\newcolumntype{P}[1]{>{\centering\arraybackslash}p{#1}}
\newcolumntype{M}[1]{>{\centering\arraybackslash}m{#1}}

\begin{table*}[p]
  \centering
  \begin{tabular}{M{0.2\linewidth}|M{0.2\linewidth}|M{0.2\linewidth}}
    \centering 
    \textbf{Expressionism} & \textbf{Symbolism} & \textbf{Impressionism} \\
    \hline
    \vspace{3pt}
    \stackunder[2pt]{\includegraphics[width=\linewidth]{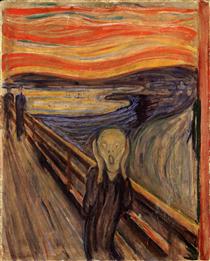}}{\textit{The Scream}} \vspace{1pt} &
    \stackunder[2pt]{\includegraphics[width=\linewidth]{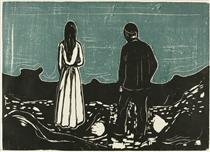}}{\textit{The Lonely Ones}} &
    \stackunder[2pt]{\includegraphics[width=\linewidth]{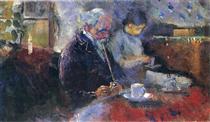}}{\textit{At The Coffee Table}}  \\
    \hline
    \vspace{3pt}
    \stackunder[2pt]{\includegraphics[width=\linewidth]{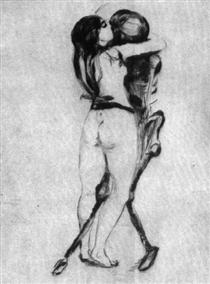}}{\textit{Girl and Death}} \vspace{1pt} &
    \stackunder[2pt]{\includegraphics[width=\linewidth]{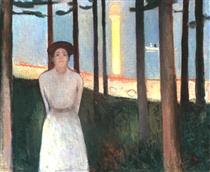}}{\textit{The Voice}} &
    \stackunder[2pt]{\includegraphics[width=\linewidth]{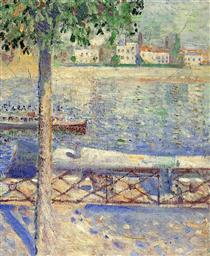}}{\textit{View over the Rover}} \\
    \hline
    \stackunder[2pt]{\includegraphics[width=\linewidth]{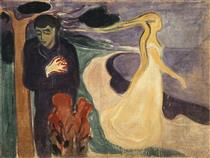}}{\textit{Separation}} &
    \vspace{3pt}
    \stackunder[2pt]{\includegraphics[width=\linewidth]{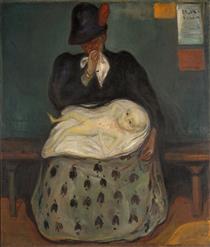}}{\textit{Inheritance}} \vspace{1pt} &
    \stackunder[2pt]{\includegraphics[width=\linewidth]{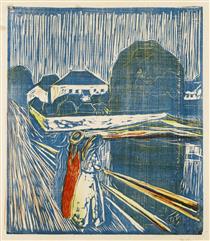}}{\textit{Girls on the Bridge}} \\
    \hline
    \vspace{3pt}
    \stackunder[2pt]{\includegraphics[width=\linewidth]{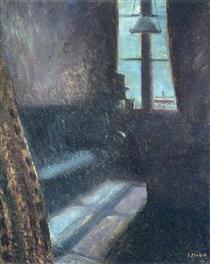}}{\textit{Night}} &
    \stackunder[2pt]{\includegraphics[width=\linewidth]{supp_figures/art_movements/separation_symb.jpg}}{\textit{Separation}} &
    \vspace{3pt}
    \stackunder[2pt]{\includegraphics[width=\linewidth]{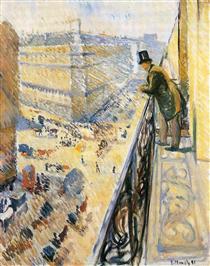}}{\textit{Street Lafayette}} \\
    
  \end{tabular}
  \vspace{10pt}
  \caption{A few Edvard Munch artworks that fall under Expressionism, Symbolism and Impressionism art movements. (Source: WikiArt \cite{wikiart-web,tan2016ceci})
  }
    \label{tab:munch_arts}
\end{table*}
\clearpage

\section{Appendix: Summary}
 
 This Appendix provides the supplementary material for the main paper.
This appendix is organized as follows:
\begin{itemize}
    \item In Section \textbf{Inadequacy of Existing Style Categorizations}, we highlight the problems in the existing style categorizations in the context of style-based clustering of artworks.
    \item In Section \textbf{Background on Deep Embedded Clustering}, we present the background on the Deep Embedded clustering model that we utilize in our paper for different experiments.
    \item In Section \textbf{Dimensionality Reduction}, we present additional details on the dimensionality reduction method we use for all the features.
    \item In Section \textbf{Additional Figures and Results}, we present the visual examples of clustering results that we obtain from different feature representations and datasets. We also present a few additional experimental details and results such as UMAPs and cluster distributions.
\end{itemize}

\section{Inadequacy of Existing Style Categorizations}

In this section, we illustrate the shortcomings of existing broad style categorizations. We observe that they do not accurately classify artworks based on style. For example, in Table \ref{tab:munch_arts}, we present a few artworks of Edvard Munch that belong to different art movements and emphasize that these categorizations do not yield desired clusterings based on style. 

Existing categorization of artworks based on style, for example, done on the basis for  \textit{art movements}, gives us a broad categorization. It builds upon the shared style characteristics within different artworks belonging to an art movement. We don't get much information about styles other than the common philosophy or the trend specific to an art movement. One can easily identify a lot of style-dissimilarity among artworks within the same art movement in terms of different style characteristics like colors, brush strokes, shadings, and visual entities. For instance, when we compare the paintings \textit{`The Scream'}(row1, col1) and \textit{`Night'}(row4, col1) from the same \textit{Expressionism} movement (Table \ref{tab:munch_arts}), we observe that the strokes in the former are quite curved and follow a different pattern in the latter artwork. Similarly, the color scheme in \textit{`The Scream'}(row1, col1) looks quite different than that's used in \textit{`The Girl and Death'}(row2, col1). One looks more vivid while the other tends to be monochromatic. If you observe the artworks (Table \ref{tab:munch_arts}), you can find that \textit{`Separation'} is classified as representing both \textit{Expressionism}(row3, col1) and \textit{Symbolism}(row4, col2). This implies overlaps among different categorizations. So, we emphasize that we cannot rely on the pre-existing style-based categorization of artworks and we should explore methods to identify and distinguish fine-granular styles in artworks.

\section{Background on Deep Embedded Clustering}
\label{sec:dec}

\begin{figure}[b!]
    \centering
    \includegraphics[width=0.5\textwidth]{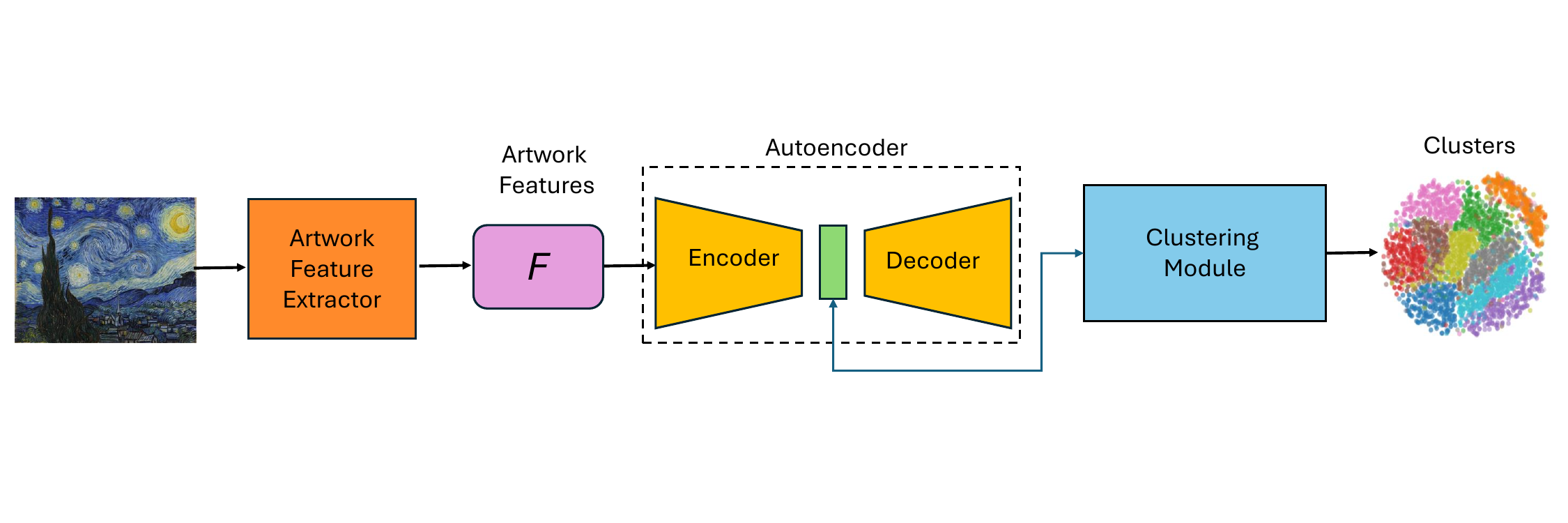}
    \caption{Deep Embedded Clustering Architecture}
    \label{fig:dec}
\end{figure}

We briefly discussed the Deep embedded clustering model in Section \textit{Experiments and Evaluation Criteria} of our paper. In this section, we discuss in detail 
the generalized architecture for Deep Embedded Clustering(DEC) with a deep neural network that produces deep-layer features $F$ for image data, an autoencoder, and a clustering module is presented in Figure \ref{fig:dec}.  The fundamental idea of the DEC method \cite{xie2016unsupervised} is to learn a mapping from the data space to a lower-dimensional feature space which is iteratively optimized with a clustering objective. The model consists of an autoencoder and a clustering layer connected to the embedding layer of the autoencoder. 

\textbf{Autoencoder:} Autoencoders are deep neural networks that can project the input data into latent space using an encoder and reconstruct the original input from latent space using a decoder. The encoder present in the autoencoder first takes the input data and transforms the data with a non-linear mapping $\phi_{\theta} : X \xrightarrow{} Z$ where X is the input space of the data and Z is the hidden latent space. 
The decoder learns to reconstruct the original input based on the latent representation, $\psi : Z \xrightarrow{} X$. 
The latent embedded features are then propagated through the decoder so it can reconstruct the latent features back to the original input space. The non-linear mapping of $\phi$ and $\psi$ is learnt by updating the autoencoder parameters by minimizing a classic mean squared reconstruction loss:
\begin{equation}
    L_r = \frac{1}{n} \sum_{i=1}^{n}||x_i^{\prime}-x_i||^{2}=\frac{1}{n}\sum_{i=1}^{n}||\psi(\phi(x_i))-x_i||^{2}
\end{equation}
    
where n is the cardinality of the input features, $x_i$ is the \textit{i}-th input sample, $x_i^{\prime}$ is the reconstruction performed by the decoder and $||\cdot||$ is the Eucledian Distance.

\textbf{Clustering Layer:} The clustering layer takes the latent embedded features from the encoder based on the non-linear mapping $\phi : X \xrightarrow{} Z$ and initially assigns each embedded point to \textit{k} cluster centroids by using k-means clustering $\bigl\{ c_j \in Z\bigr\}_{j=1}^{k}$ where $c_j$ represents the \textit{j}th cluster centroid. 

After the initialization, each embedded point, $z_i = \phi(x_i)$ is mapped to a cluster centroid $c_j$ by using a cluster assignment \textit{Q} based on Student's t-distribution:
\begin{equation}
    q_{ij}= \frac{(1+||z_i - c_j||^2)^{-1}}{\sum_{j^{\prime}{}}(1+||z_i - c_{j^{\prime}}||^2)^{-1}}
    \label{eq: q}
\end{equation}
where $j^\prime$ represent every cluster and $q_{ij}$ represents the membership probability of $z_i$ to belong to the cluster \textit{j} which basically soft assigns $z_i$ to cluster centroid $c_j$. $q_{ij}$ represents the similarity between a datapoint $z_i$ and the cluster centroid $c_j$ which gives us the confidence of a datapoint being assigned to a particular cluster.

The decoder of the autoencoder is abandoned and the DEC model jointly optimizes clustering layer and encoder based on the auxiliary target distribution $p_{ij}$ calculated from $q_{ij}$ derived from Eq.\ref{eq: q} which emphasizes the data points that have higher confidence assigned to them while also minimizing the loss contribution of each centroid:
\begin{equation}
    p_{ij}=\frac{q_{ij}^{2}/f_j}{\sum_{j^{\prime}{}}q_{i{j^\prime}}^{2}/f_{j^\prime}}
    \label{eq:p}
\end{equation}
where $f_j=\sum_{j}{}q_{ij}$ are the soft cluster frequencies. The DEC model optimizes the target function by minimizing the Kullback-Leibler(KL) divergence between P and Q where P is the auxiliary target function defined in Eq.\ref{eq:p} and Q is the cluster assignment based on Student's t-distribution. This improves the initial cluster estimate by learning from previous high-confidence predictions.
\begin{equation}
    L_c=KL(P||Q)=\sum_{i}{}\sum_{j}{}p_{ij}\log\biggl(\frac{p_{ij}}{q_{ij}}\biggr)
\end{equation}

The cluster centers $c_j$ and the encoder parameters $\theta$ (of autoencoder) are then jointly optimized using Stochastic Gradient Descent (SGD) with momentum \cite{xie2016unsupervised}.

\section{Dimensionality Reduction}
\label{sec:dim_red}
In the Section \ref{sec: feat_explo} of our paper, we mentioned that we choose global average pooling to reduce the dimensions of our $F_{CNN}$ features. In this section, we explain in detail our reasoning for choosing the global pooling average as our dimensionality reduction method.
The dimensions of the $F_{CNN}$ features are very high - ranging from 50000 to 200000 approximately. The feature space is too large for efficient clustering; hence we perform dimensionality reduction before clustering. We utilize Global Average Pooling (GAP) \cite{gpa} as our dimensionality reduction method.
We choose GAP over Principal Component Analysis (PCA) \cite{pca} as it is computationally intensive to apply PCA on a large dataset such as WikiArt (78,978 artworks) and the reduced features with both methods form similar clusters. We verify this by experimenting with PCA and GAP on the smaller Brueghel dataset and finding that the quantitative results are quite similar to each other as can be seen in Table \ref{tab:quant-dr-brueghel}. We use the remaining features ($F_{ST}$,$F_{Text}$ and $F_{Train}$) directly for clustering as their dimensionality is sufficiently low. The dimensions before and after using GAP, as well as the respective dimensions of all features can be seen in Table \ref{tab:dim-red}. 

We also test with varying the DEC encoder's final layer's size (refer to \ref{tab:enc_size}) to see if it has an effect on the features. We observe that the there is minimal change in the cluster ability of the features when the they are encoded to different sized latents.

\begin{table}[ht!]
\centering
\resizebox{.8\linewidth}{!}{%
\begin{tabular}{c|c|c}
\textbf{\begin{tabular}[c]{@{}c@{}}Dimensionality\\ Reduction\end{tabular}} & \textbf{SC} & \textbf{CHI} \\ \hline
\begin{tabular}[c]{@{}c@{}}No Dimensionality\\ Reduction\end{tabular}   & 0.16           & \textbf{595.19} \\ \hline
\begin{tabular}[c]{@{}c@{}}Principal Component \\ Analysis (PCA)\end{tabular} & 0.191          & 333.84          \\ \hline
\begin{tabular}[c]{@{}c@{}}Global Average \\ Pooling (GAP)\end{tabular}       & \textbf{0.204} & 405.41         
\end{tabular}%
}
\caption{Quantitative results on the Brueghel dataset for different dimensionality reduction methods using the $F_{gram}$ features and the DEC model.}
\label{tab:quant-dr-brueghel}
\end{table}
\begin{table}[ht!]
\centering
\resizebox{.8\columnwidth}{!}{%
\begin{tabular}{c|c|c}
\textbf{\begin{tabular}[c]{@{}c@{}}Encoder Final\\ Layer Size\end{tabular}} & \textbf{SC} & \textbf{CHI} \\ \hline
10  & 0.118 & 8059.72 \\
50  & 0.102 & 226.7   \\
100 & 0.204 & 465.31 
\end{tabular}%
}
\caption{Quantitative results for the Brueghel dataset when the size of final layer of the encoder used in DEC is varied. We observe minimal change in the results even when we change the final layer of the encoder.}
\label{tab:enc_size}
\end{table}
\begin{table}[ht!]
\resizebox{\columnwidth}{!}{%
\begin{tabular}{cc|c|c|c|c}
\multicolumn{2}{c|}{{\color[HTML]{000000} \textbf{Features}}}                                                      & {\color[HTML]{000000} \textbf{\begin{tabular}[c]{@{}c@{}}Dimensionality \\ Reduction \\ (Yes/No)\end{tabular}}} & {\color[HTML]{000000} \textbf{\begin{tabular}[c]{@{}c@{}}Dimensionality \\ Reduction \\ Method\end{tabular}}} & {\color[HTML]{000000} \textbf{\begin{tabular}[c]{@{}c@{}}Dimensions\\ Before\\ Reduction\end{tabular}}} & {\color[HTML]{000000} \textbf{\begin{tabular}[c]{@{}c@{}}Dimensions \\ After\\ Reduction\end{tabular}}} \\ \hline
\multicolumn{1}{c|}{{\color[HTML]{000000} }}                              & {\color[HTML]{000000} $F_{Dense}$}     & {\color[HTML]{000000} Yes}                                                                                      & {\color[HTML]{000000} GAP}                                                                                    & {\color[HTML]{000000} 1024$\times$7$\times$7}                                                           & {\color[HTML]{000000} 1024}                                                                             \\
\multicolumn{1}{c|}{{\color[HTML]{000000} }}                              & {\color[HTML]{000000} $F_{Gram}$}      & {\color[HTML]{000000} Yes}                                                                                      & {\color[HTML]{000000} GAP}                                                                                    & {\color[HTML]{000000} 512$\times$512}                                                    & {\color[HTML]{000000} 512}                                                                              \\
\multicolumn{1}{c|}{\multirow{-3}{*}{{\color[HTML]{000000} $F_{CNN}$}}}   & {\color[HTML]{000000} $F_{g \cdot c}$} & {\color[HTML]{000000} Yes}                                                                                      & {\color[HTML]{000000} GAP}                                                                                    & {\color[HTML]{000000} 512 $\times$}                                                    & {\color[HTML]{000000} 512}                                                                              \\ \hline
\multicolumn{1}{c|}{{\color[HTML]{000000} }}                              & {\color[HTML]{000000} $F_{StyleCap}$}  & {\color[HTML]{000000} No}                                                                                       & {\color[HTML]{000000} -}                                                                                      & {\color[HTML]{000000} 768}                                                                              & {\color[HTML]{000000} -}                                                                                \\
\multicolumn{1}{c|}{\multirow{-2}{*}{{\color[HTML]{000000} $F_{Text}$}}}  & {\color[HTML]{000000} $F_{Annot}$}     & {\color[HTML]{000000} No}                                                                                       & {\color[HTML]{000000} -}                                                                                      & {\color[HTML]{000000} 58}                                                                               & {\color[HTML]{000000} -}                                                                                \\ \hline
\multicolumn{1}{c|}{{\color[HTML]{000000} }}                              & {\color[HTML]{000000} $F_{StyleGAN}$}  & {\color[HTML]{000000} No}                                                                                       & {\color[HTML]{000000} -}                                                                                      & {\color[HTML]{000000} 512}                                                                              & {\color[HTML]{000000} -}                                                                                \\
\multicolumn{1}{c|}{{\color[HTML]{000000} }}                              & {\color[HTML]{000000} $F_{Stytr2}$}    & {\color[HTML]{000000} No}                                                                                       & {\color[HTML]{000000} -}                                                                                      & {\color[HTML]{000000} 512}                                                                              & {\color[HTML]{000000} -}                                                                                \\
\multicolumn{1}{c|}{{\color[HTML]{000000} }}                              & {\color[HTML]{000000} $F_{Mamba}$}     & {\color[HTML]{000000} No}                                                                                       & {\color[HTML]{000000} -}                                                                                      & {\color[HTML]{000000} 512}                                                                              & {\color[HTML]{000000} -}                                                                                \\
\multicolumn{1}{c|}{\multirow{-4}{*}{{\color[HTML]{000000} $F_{ST}$}}}    & {\color[HTML]{000000} $F_{StyleShot}$} & {\color[HTML]{000000} No}                                                                                       & {\color[HTML]{000000} -}                                                                                      & {\color[HTML]{000000} 9216}                                                                             & {\color[HTML]{000000} -}                                                                                \\ \hline
\multicolumn{1}{c|}{{\color[HTML]{000000} }}                              & {\color[HTML]{000000} $F_{CSD}$}       & {\color[HTML]{000000} No}                                                                                       & {\color[HTML]{000000} -}                                                                                      & {\color[HTML]{000000} 768}                                                                              & {\color[HTML]{000000} -}                                                                                \\
\multicolumn{1}{c|}{{\color[HTML]{000000} }}                              & {\color[HTML]{000000} $F_{Artist}$}    & {\color[HTML]{000000} No}                                                                                       & {\color[HTML]{000000} -}                                                                                      & {\color[HTML]{000000} 1024}                                                                             & {\color[HTML]{000000} -}                                                                                \\
\multicolumn{1}{c|}{\multirow{-3}{*}{{\color[HTML]{000000} $F_{Train}$}}} & {\color[HTML]{000000} $F_{ArtMove}$}   & {\color[HTML]{000000} No}                                                                                       & {\color[HTML]{000000} -}                                                                                      & {\color[HTML]{000000} 1024}                                                                             & {\color[HTML]{000000} -}                                                                               
\end{tabular}%
}
\caption{Summary of the dimensions of different features and the dimensionality reduction method used. GAP stands for global average pooling \cite{gpa}. We utilize GAP only for the $F_{CNN}$ features due to their high dimensions.}
\label{tab:dim-red}
\end{table}

\section{Additional Observations and Results}
\label{sec: additional_results}
In this section, we detail our additional experiments and provide a few more details and results for experiments described in Section \textit{Results and Discussion} of our paper.

\subsection{Examples from the Synthetically Curated Dataset}For the synthetically curated dataset mentioned in Section \textbf{Datasets and Evaluation Criteria}, we present a few examples for each dataset obtained by using Styleshot \cite{styleshot} in Tables \ref{tab:munch_ground_truth}, \ref{tab:wikiart_ground_truth}, \ref{tab:brueghel_ground_truth} and \ref{tab:clip-art_ground_truth}.

\subsection{Examples for $F_{StyleCap}$ and $F_{Annot}$}
\begin{figure}[!ht]
  \begin{subfigure}[t]{0.48\textwidth}
    \centering
    \includegraphics[width=\linewidth]{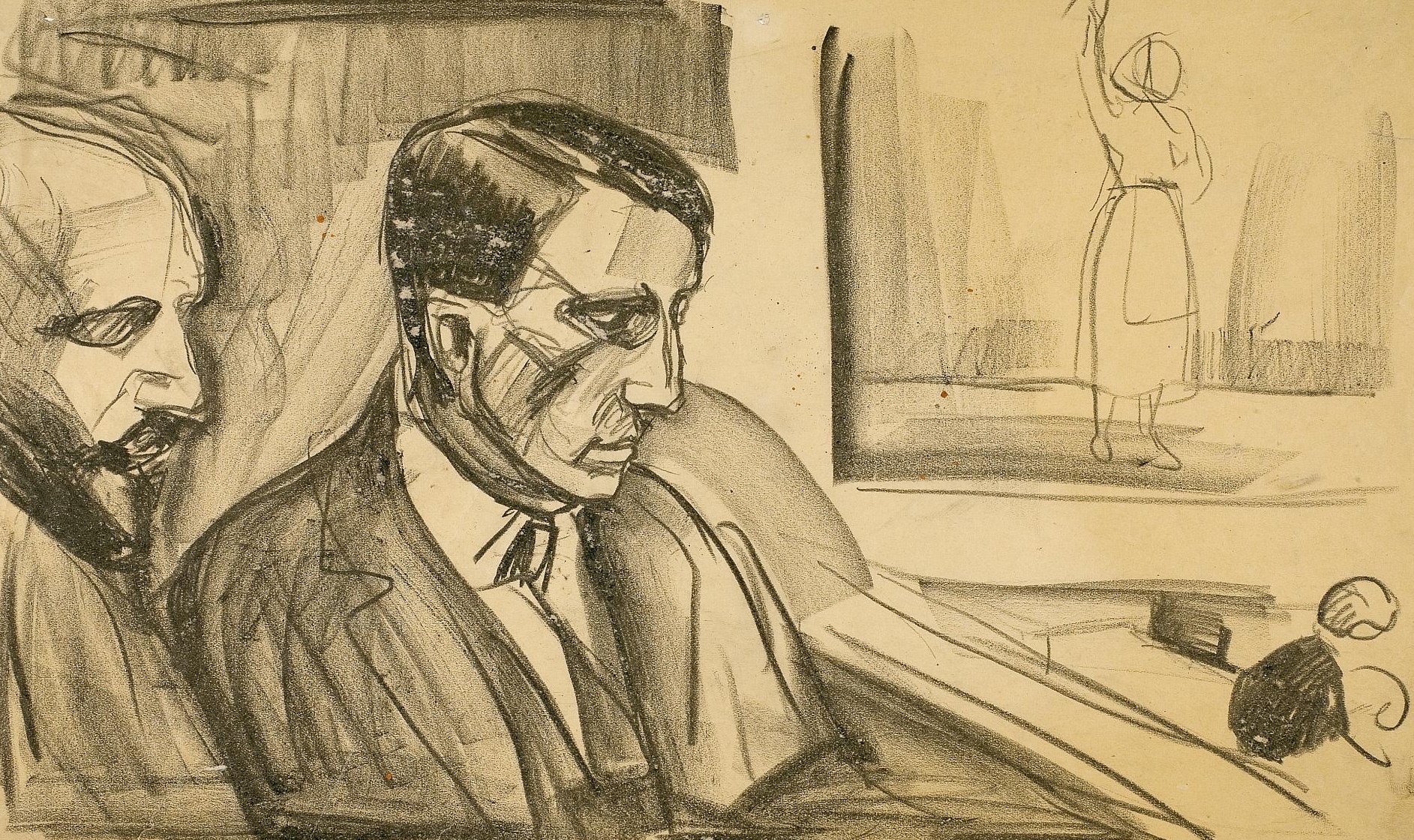}
    \caption{The style of the given image is reminiscent of early 20th-century European art, possibly influenced by the works of artists like Gustave Moreau or the post-impressionist movement. The use of bold, expressive lines and a monochromatic palette is characteristic of this era. The detailed shading and cross-hatching techniques suggest a focus on texture and depth, creating a sense of three-dimensionality and movement within the composition. The figure in the background, rendered in a simplified manner, adds a sense of context and narrative to the scene. The overall style is evocative of a time when artistic expression was deeply intertwined with the emotional and psychological states of the subject matter.}
  \end{subfigure}
   \hfill
   \begin{subfigure}[t]{0.48\textwidth}
    \centering
    \includegraphics[width=\linewidth]{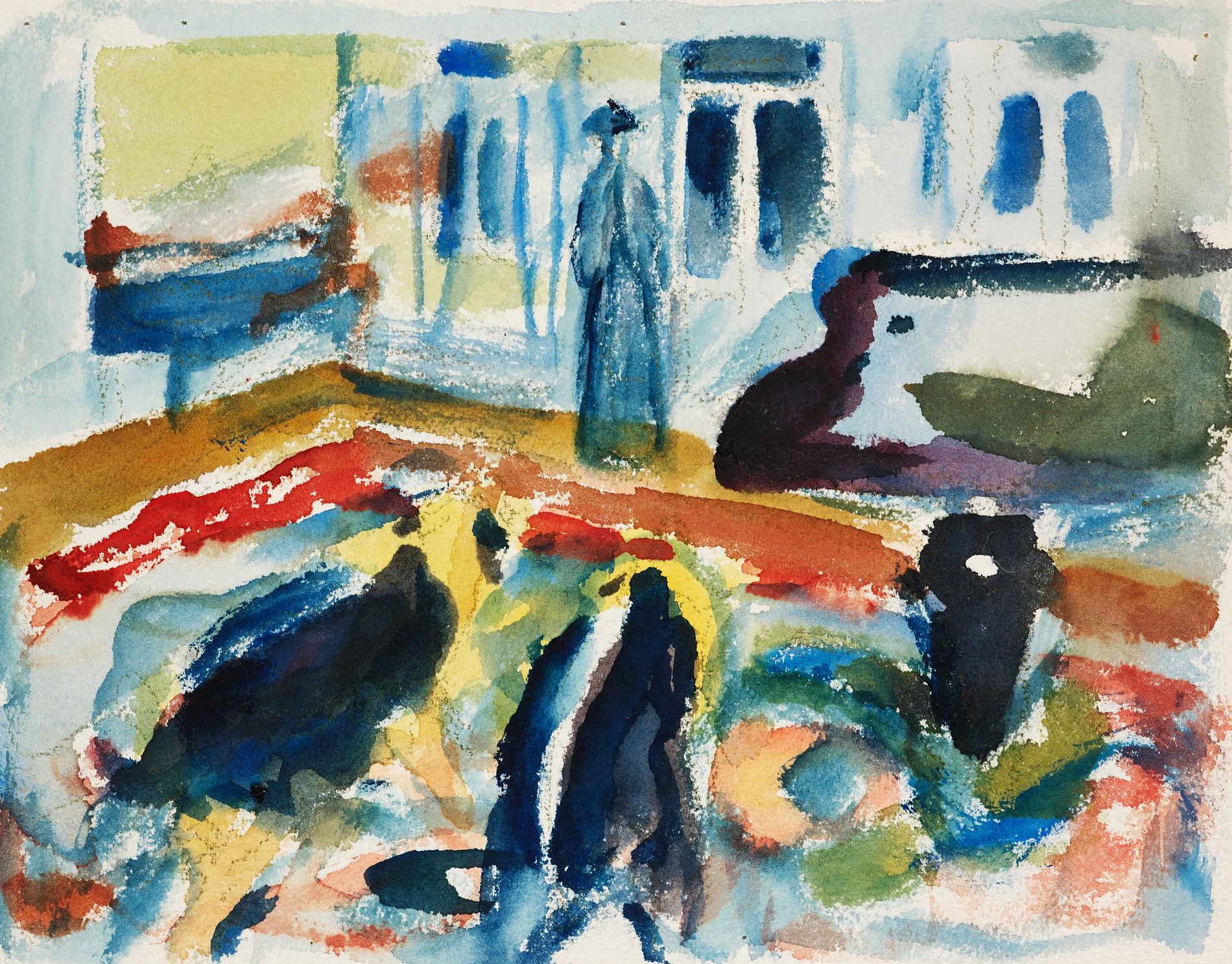}
    \caption{The style of the given image is Abstract Expressionism. This style is characterized by loose, gestural brushwork, vibrant colors, and a focus on the physical act of painting. The painting appears to convey a sense of movement and emotion, with a dynamic interplay of shapes and colors that evoke a lively and energetic atmosphere. The use of bold, contrasting colors and abstract forms suggests a departure from traditional representational art, instead embracing an expressive and personal interpretation of the subject matter.}
  \end{subfigure}
  
  \caption{Examples of artwork captions with the InternVL 2 model for the Edvard Munch archive.}
  \label{fig: caption_example}
\end{figure}

\begin{figure}[!tbh]
  \begin{subfigure}[t]{0.22\textwidth}
    \centering
    \includegraphics[width=\linewidth]{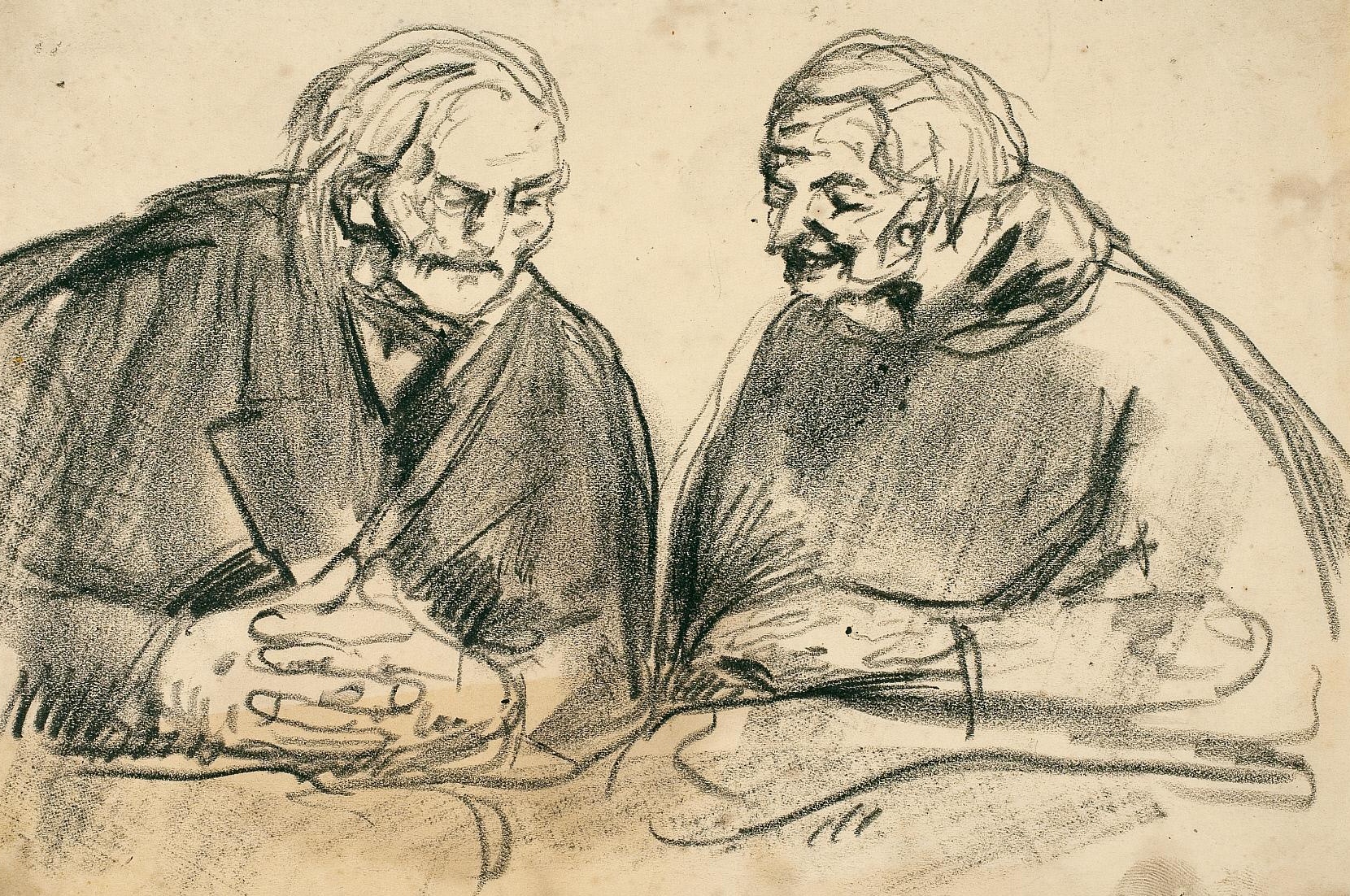}
    \caption{\textbf{Subject}: representational \\\textbf{Line}: blurred, controlled, meandering, thick, thin \\\textbf{Texture}: smooth, gestural, rough\\ \textbf{Color}: monochromatic\\ \textbf{Shape}: ambiguous, biomorphic, organic, abstract, decorative\\ \textbf{Light and Space}: dark, planar, perspective\\ \textbf{General Principles of Art}: overlapping, balance, contrast, harmony, pattern, repetition}
  \end{subfigure}
   \hfill
   \begin{subfigure}[t]{0.22\textwidth}
    \centering
    \includegraphics[width=0.6\linewidth]{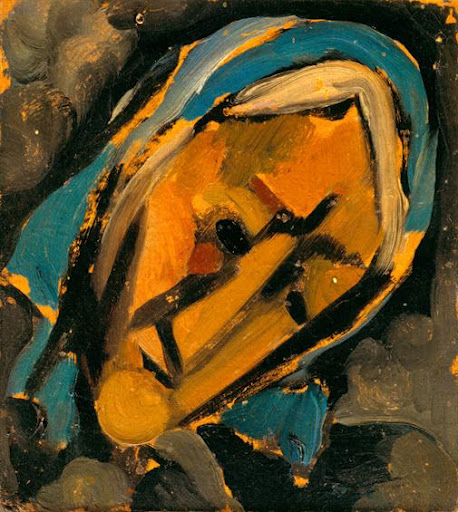}
    \caption{\textbf{Subject}: representational, non-representational\\ \textbf{Line}: blurred, controlled, energetic, straight\\ \textbf{Texture}: smooth, gestural, rough\\ \textbf{Color}: cool, warm, muted, chromatic\\ \textbf{Shape}: ambiguous, organic, abstract, decorative\\ \textbf{Light and Space}: bright, dark, atmospheric, planar\\ \textbf{General Principles of Art}: balance, contrast, harmony, pattern, repetition, rhythm, unity, variety, symmetry, proportion, parallel.
}
  \end{subfigure}
  \caption{Artwork style concept annotation with the Intern-VL2 model.}
  \label{fig: annotation_example}
\end{figure}

We present a few examples for the $F_{Text}$ features described in Section \textbf{Methodology}. We obtain the captions and annotations from a Vision language model for $F_{StyleCap}$ and $F_{Annot}$ features and present them in Figure \ref{fig: caption_example} and Figure \ref{fig: annotation_example} respectively.

\subsection{Qualitative Clustering Visual for Mixed Mamba Curated Dataset} We present the qualitative clustering visual for the Mixed Mamba Curated dataset in Figure \ref{fig:qual_fig_mix_mamba} for the $F_{Stytr2}$ feautres. We observe that we're able to obtain a perfect clustering 

\subsection{Example questions from survey conducted on different clustering}
We present the example clustering shown to a user as well as the questions associated with the clustering in Figure \ref{fig:surv_exmpl}.

\subsection{UMAPs for different features on the Mixed StyleShot curated dataset}
We present the diferent UMAPs produced by K-Means and DEC algorithm in Figure \ref{fig:umap}. We obserrve that DEC is able to improve the clusters by a huge margin for features with poor cluster ability like $F_{StyleCap}$

\subsection{Cluster distribution for different features on different dataset}
We present the cluster distributions for different features on the WikiArt and Brueghel dataset in Figure \ref{fig:clust_hist}. We observe that both $F_{Dense}$ and $F_{gram}$ are unable to distribute artworks evenly accross clusters whereas $F_{StyleCap}$ is able distribute the artworks evenly.


\newpage



\afterpage{
    \clearpage

\begin{table*}[ht!]
\small
\resizebox{\columnwidth}{!}{%
\begin{NiceTabular}{c|c|c|c}
\diagbox{Style \\ Images}{Content \\ Images}
& \includegraphics[align=c,width=2cm,height=2cm]{tables/ground-truth-table/content-images/content-1.jpg} & \includegraphics[align=c,width=2cm,height=2cm]{tables/ground-truth-table/content-images/content-2.jpg} & \includegraphics[align=c,width=2cm,height=2cm]{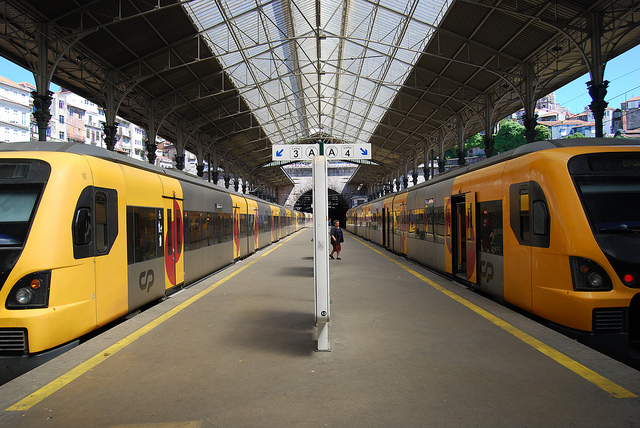}  \\ \hline
 \includegraphics[align=c,width=2cm,height=2cm]{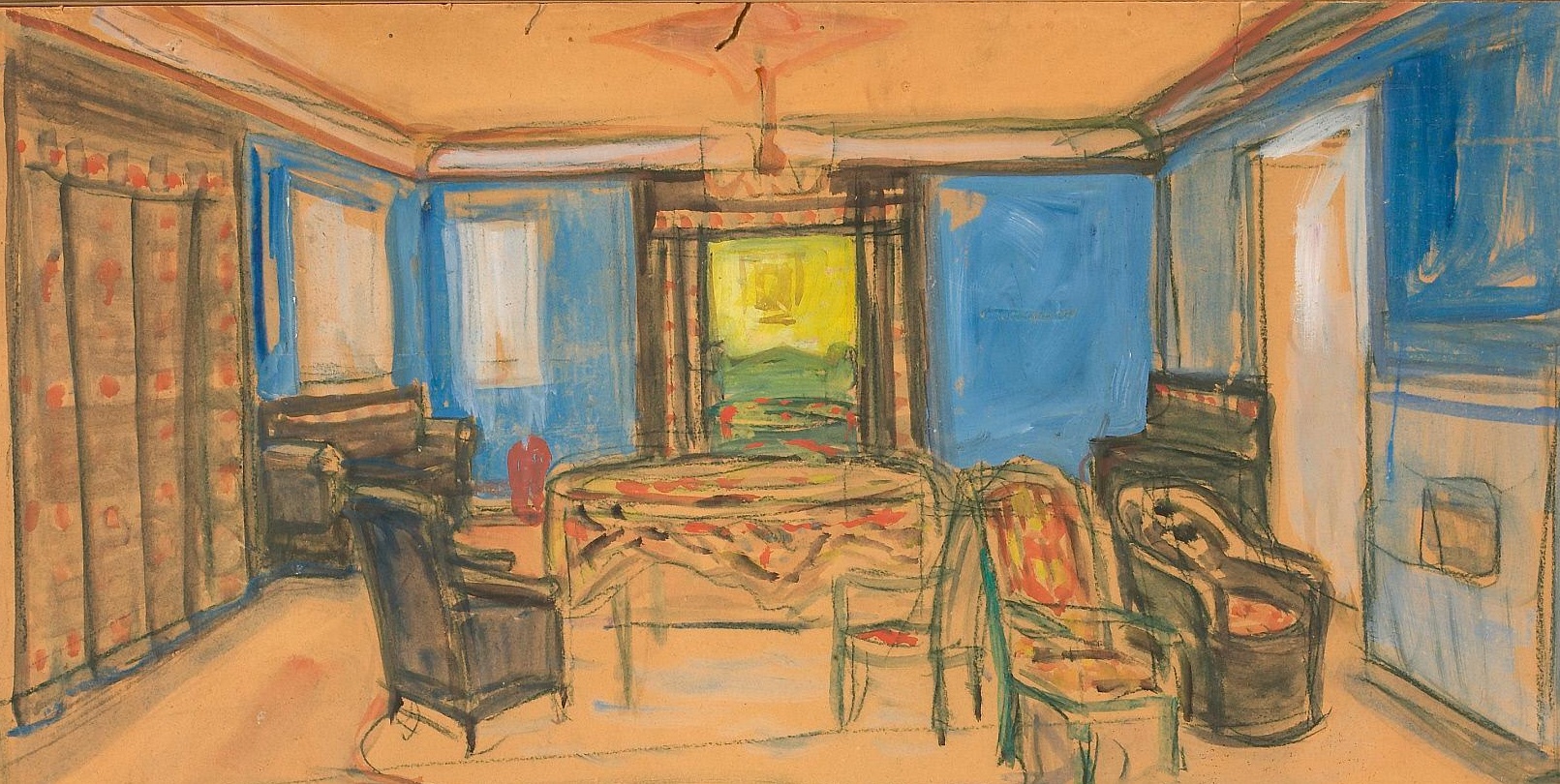}                                         & \includegraphics[align=c,width=2cm,height=2cm]{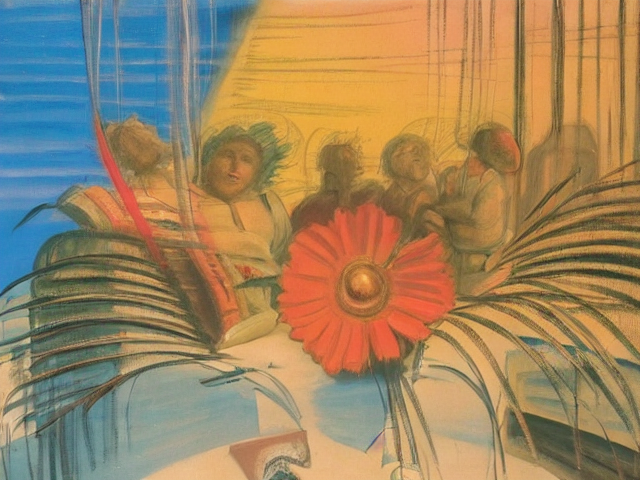} & \includegraphics[align=c,width=2cm,height=2cm]{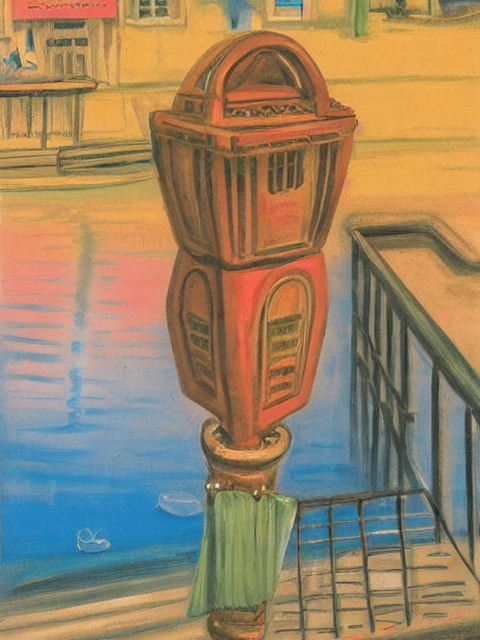} & \includegraphics[align=c,width=2cm,height=2cm]{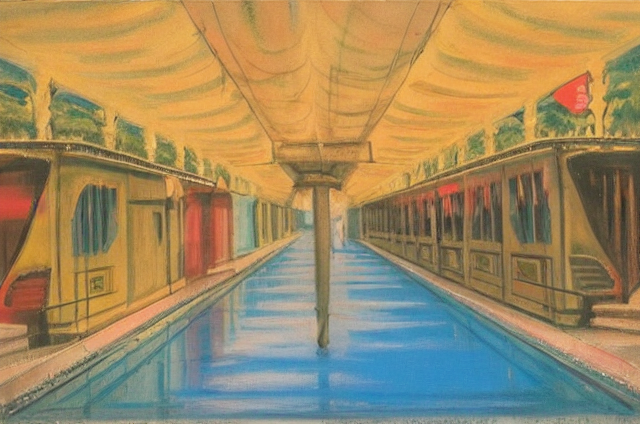} \\ \hline
\includegraphics[align=c,width=2cm,height=2cm]{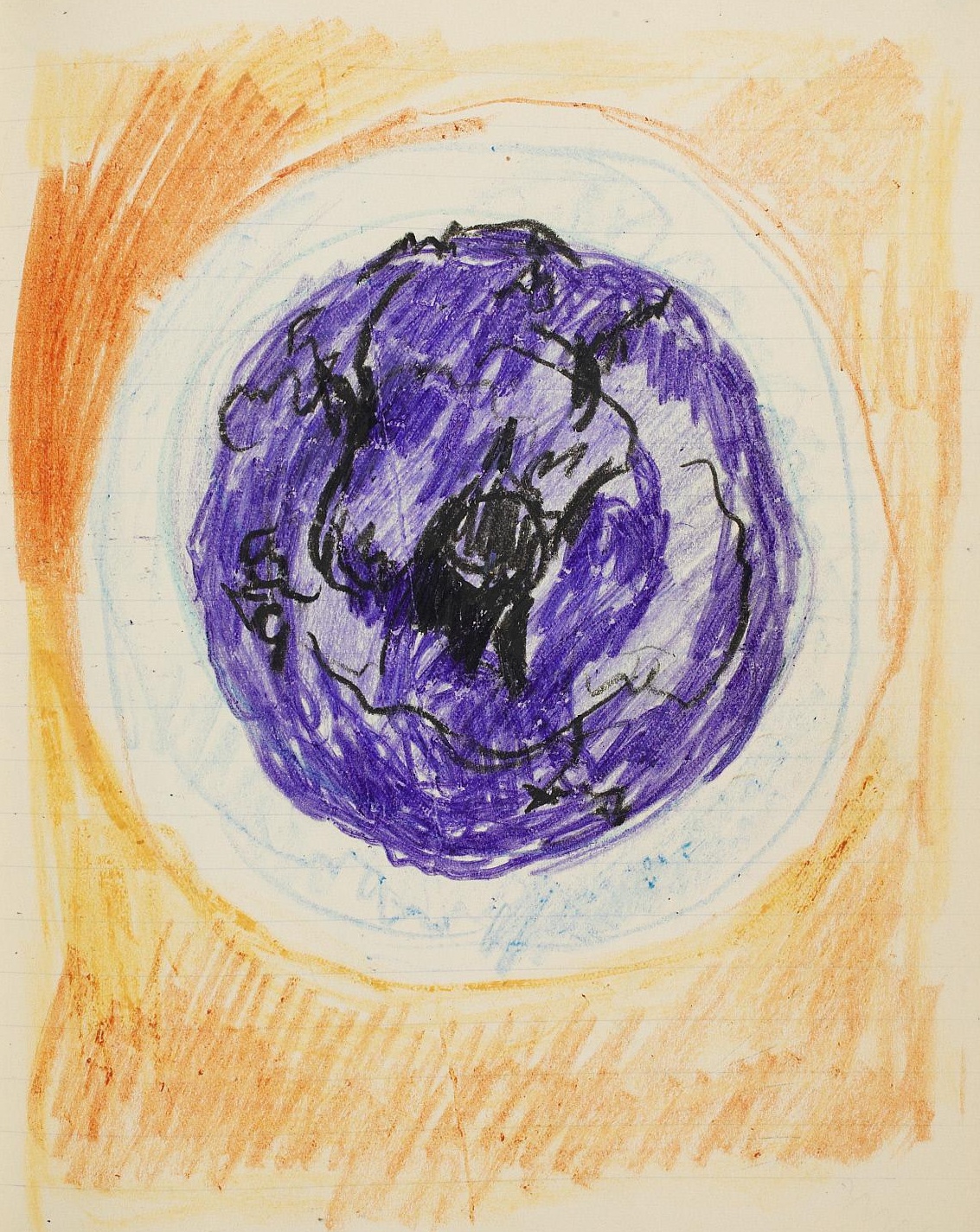}                                         & \includegraphics[align=c,width=2cm,height=2cm]{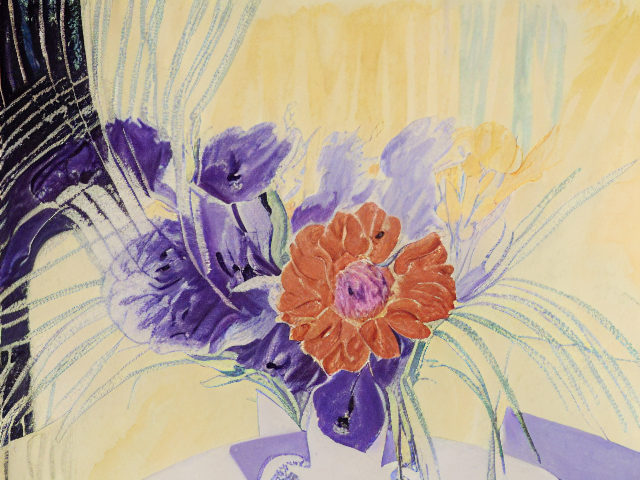} & \includegraphics[align=c,width=2cm,height=2cm]{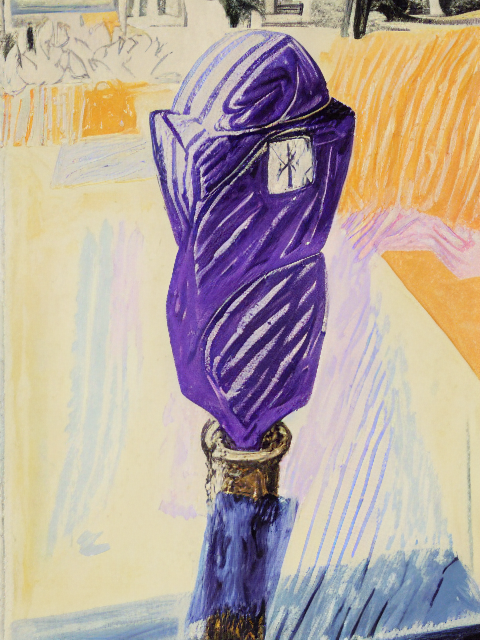} & \includegraphics[align=c,width=2cm,height=2cm]{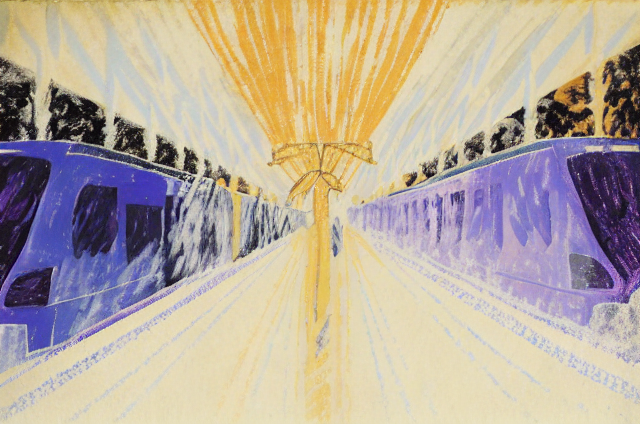}  \\ \hline
\includegraphics[align=c,width=2cm,height=2cm]{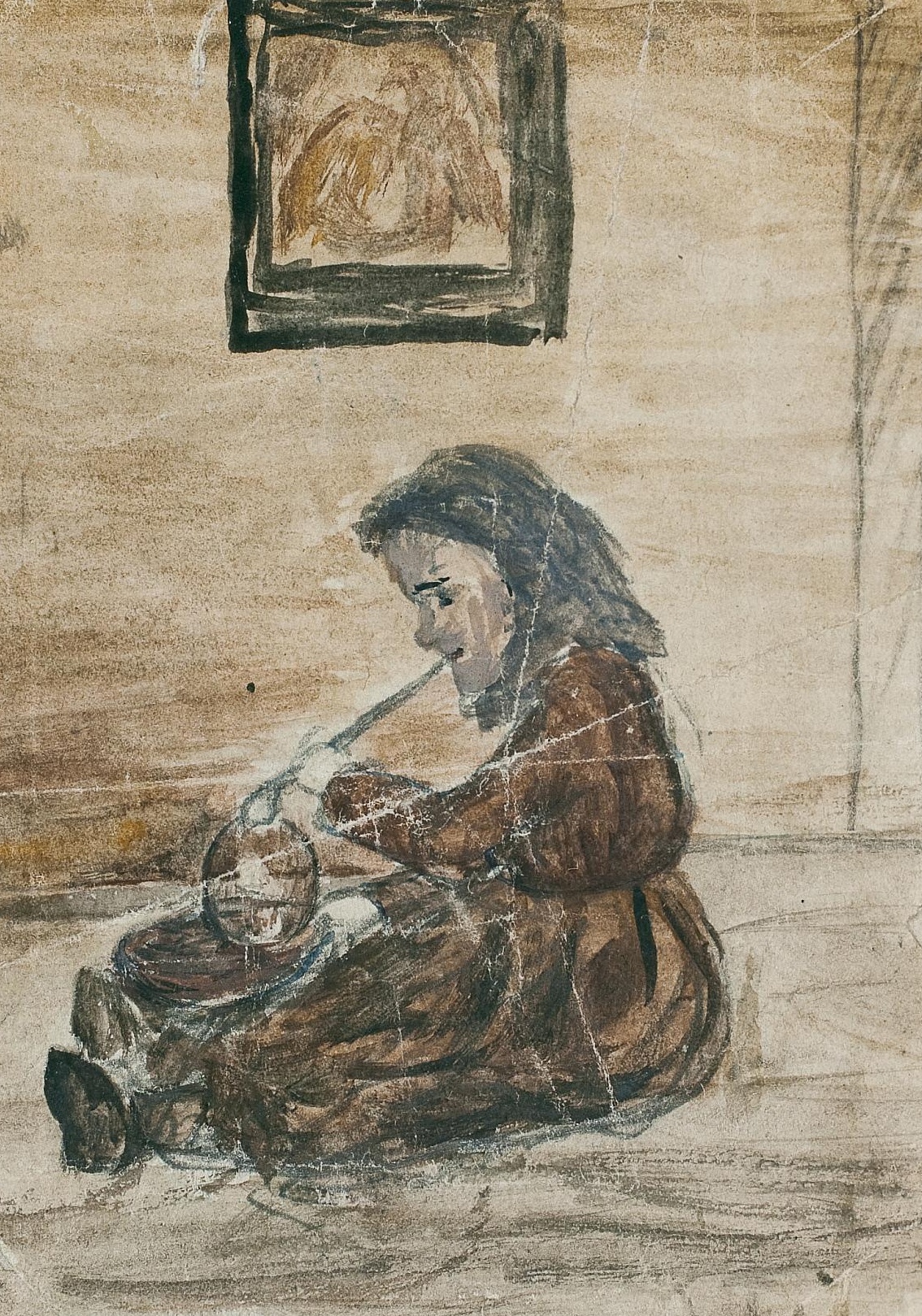}                                         & \includegraphics[align=c,width=2cm,height=2cm]{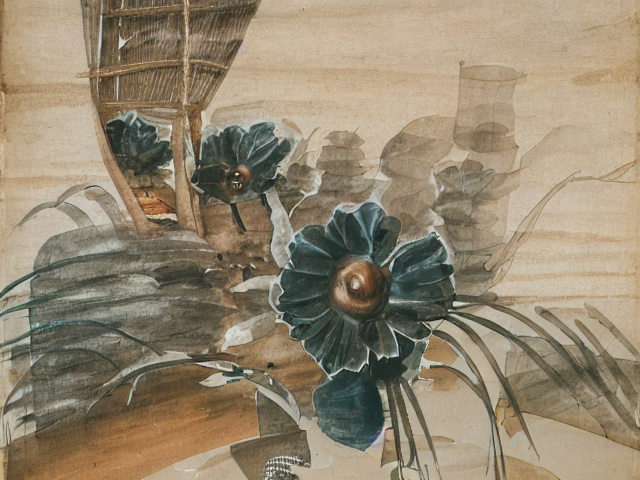} & \includegraphics[align=c,width=2cm,height=2cm]{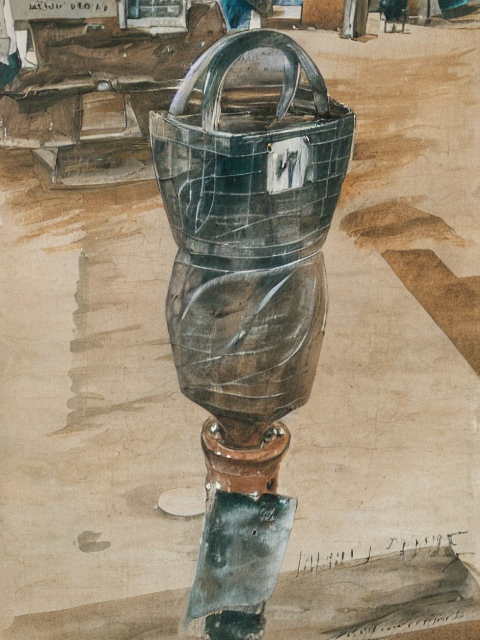} & \includegraphics[align=c,width=2cm,height=2cm]{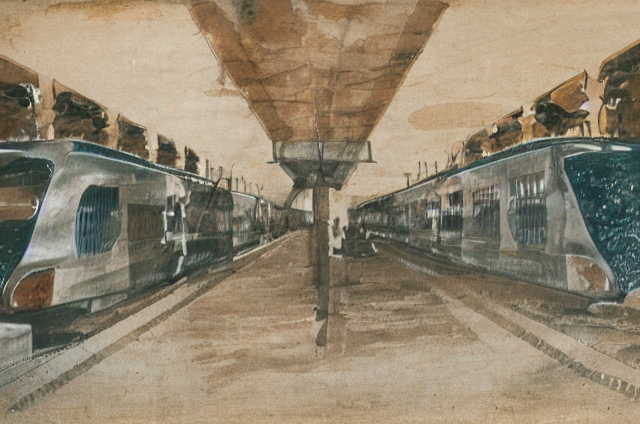}
\end{NiceTabular}%
}
\hfill
\resizebox{\columnwidth}{!}{%
\begin{NiceTabular}{c|c|c|c}
\diagbox{Style \\ Images}{Content \\ Images}
& \includegraphics[align=c,width=2cm,height=2cm]{tables/ground-truth-table/content-images/content-1.jpg} & \includegraphics[align=c,width=2cm,height=2cm]{tables/ground-truth-table/content-images/content-2.jpg} & \includegraphics[align=c,width=2cm,height=2cm]{tables/ground-truth-table/content-images/content-3.jpg}  \\ \hline
 \includegraphics[align=c,width=2cm,height=2cm]{tables/ground-truth-table/style-images/style-1.jpg}                                         & \includegraphics[align=c,width=2cm,height=2cm]{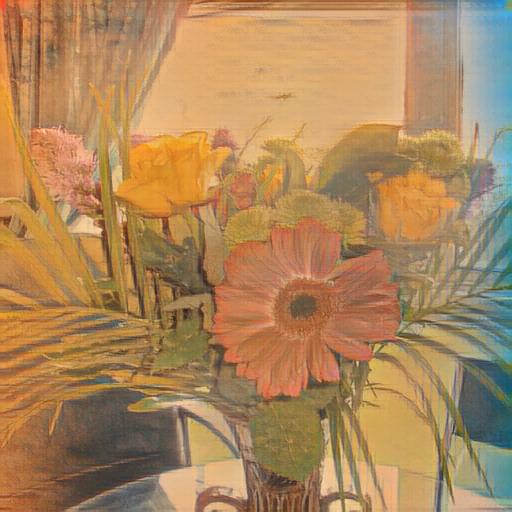} & \includegraphics[align=c,width=2cm,height=2cm]{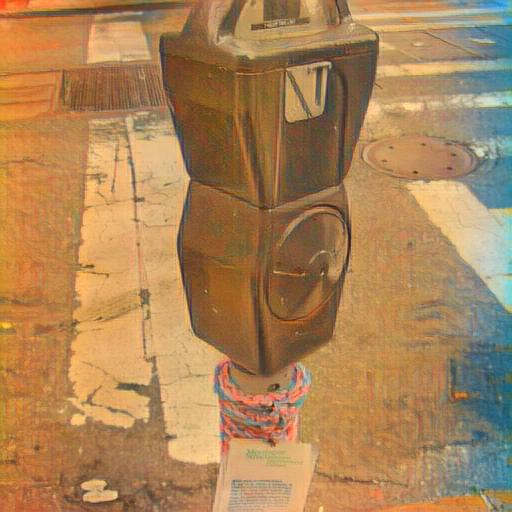} & \includegraphics[align=c,width=2cm,height=2cm]{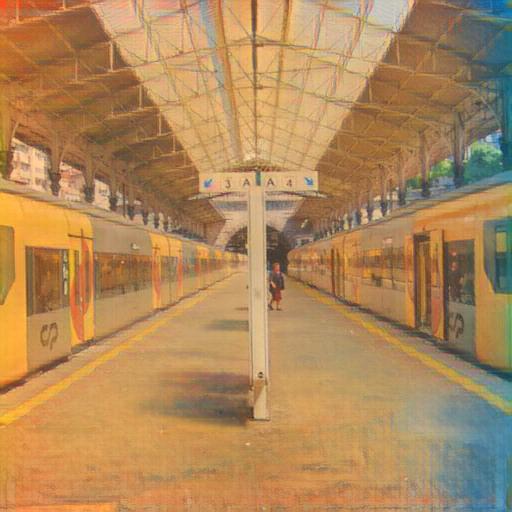} \\ \hline
\includegraphics[align=c,width=2cm,height=2cm]{tables/ground-truth-table/style-images/style-2.jpg}                                         & \includegraphics[align=c,width=2cm,height=2cm]{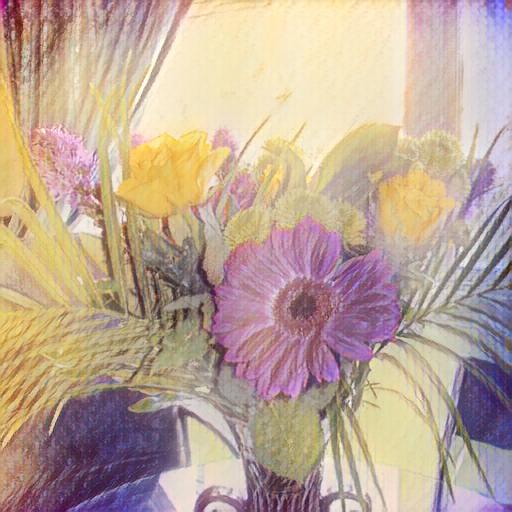} & \includegraphics[align=c,width=2cm,height=2cm]{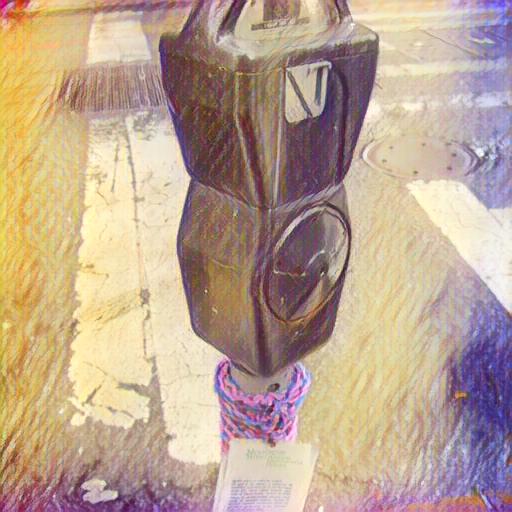} & \includegraphics[align=c,width=2cm,height=2cm]{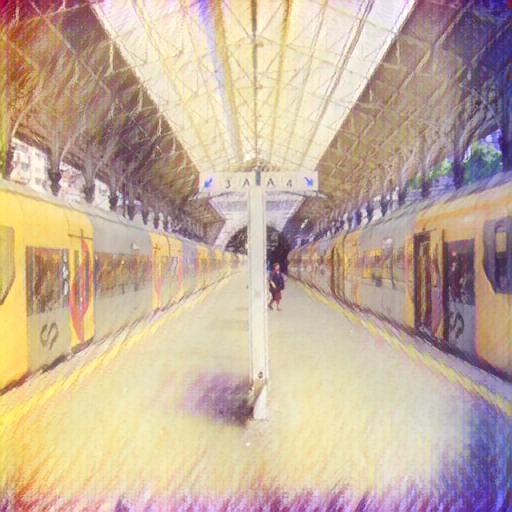}  \\ \hline
\includegraphics[align=c,width=2cm,height=2cm]{tables/ground-truth-table/style-images/style-3.jpg}                                         & \includegraphics[align=c,width=2cm,height=2cm]{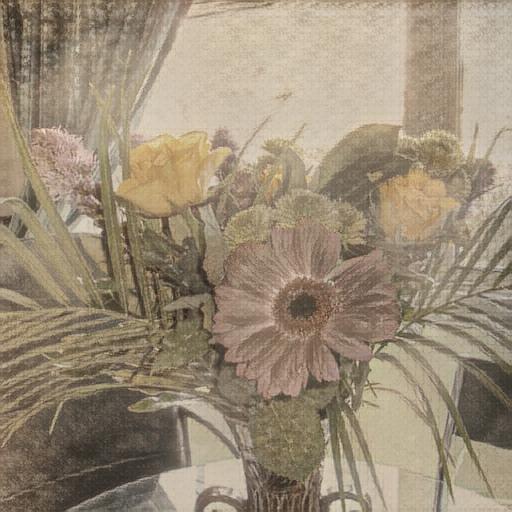} & \includegraphics[align=c,width=2cm,height=2cm]{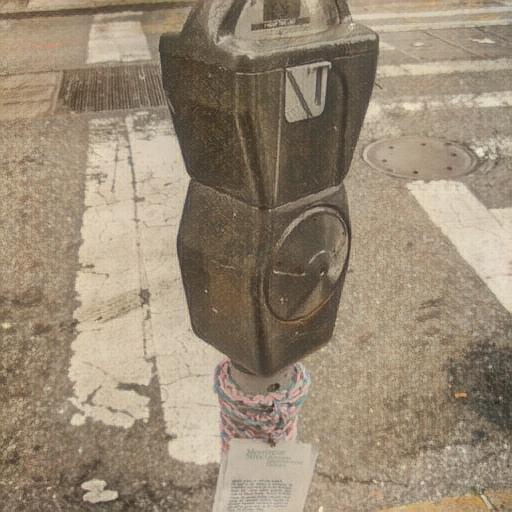} & \includegraphics[align=c,width=2cm,height=2cm]{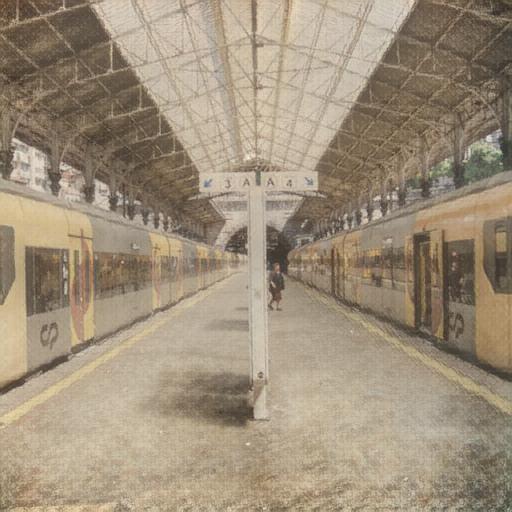}
\end{NiceTabular}%
}
\caption{Examples of content images and style images and their respective style-transfer output images from Styleshot (left) and Mamba-ST (right). The content images were picked from the MS-Coco dataset \cite{mscoco} and the style images were picked from the Edvard Munch Archive dataset \cite{newsnow}. Each row is considered as a single cluster.}
\label{tab:munch_ground_truth}
\end{table*}

\begin{table*}[ht!]
\small
\resizebox{\columnwidth}{!}{%
\begin{NiceTabular}{c|c|c|c}
\diagbox{Style \\ Images}{Content \\ Images}
& \includegraphics[align=c,width=2cm,height=2cm]{tables/ground-truth-table/content-images/content-1.jpg} & \includegraphics[align=c,width=2cm,height=2cm]{tables/ground-truth-table/content-images/content-2.jpg} & \includegraphics[align=c,width=2cm,height=2cm]{tables/ground-truth-table/content-images/content-3.jpg}  \\ \hline
 \includegraphics[align=c,width=2cm,height=2cm]{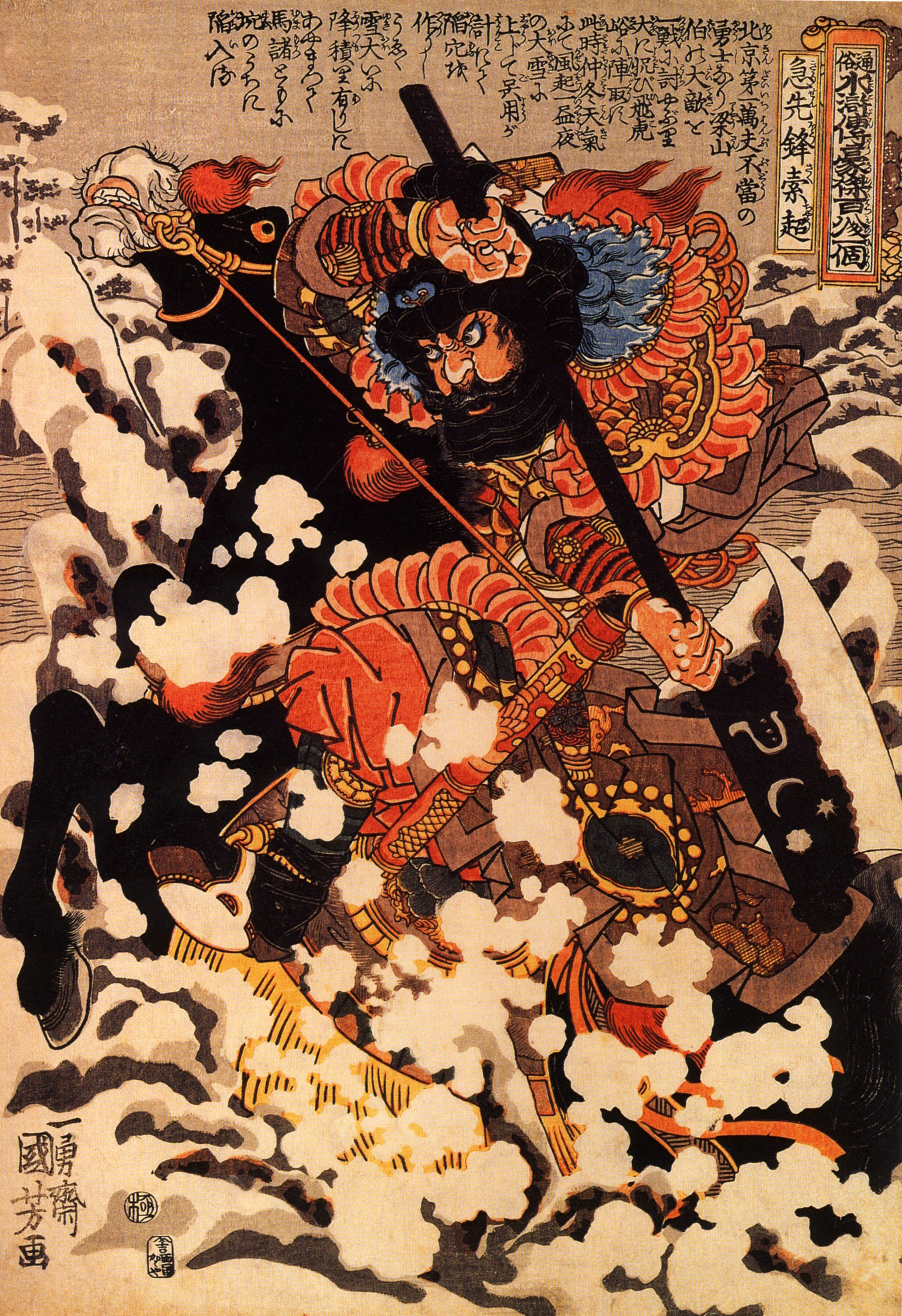}                                         & \includegraphics[align=c,width=2cm,height=2cm]{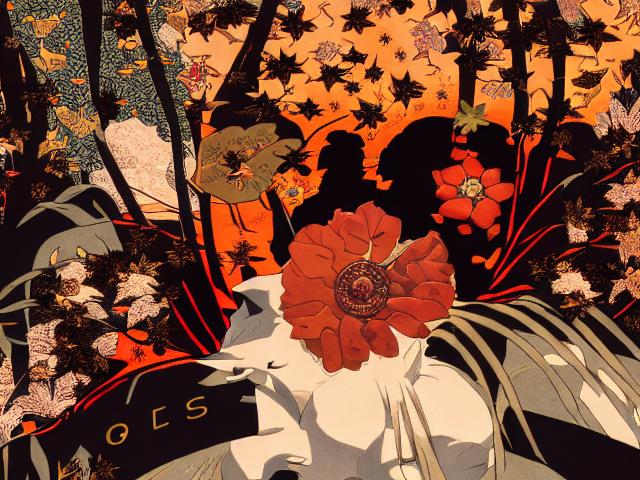} & \includegraphics[align=c,width=2cm,height=2cm]{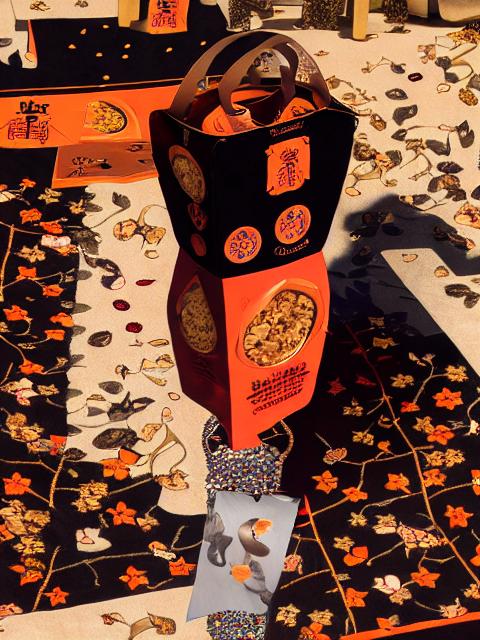} & \includegraphics[align=c,width=2cm,height=2cm]{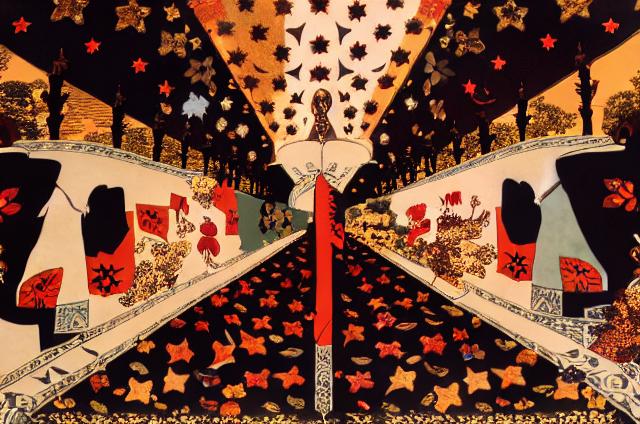} \\ \hline
\includegraphics[align=c,width=2cm,height=2cm]{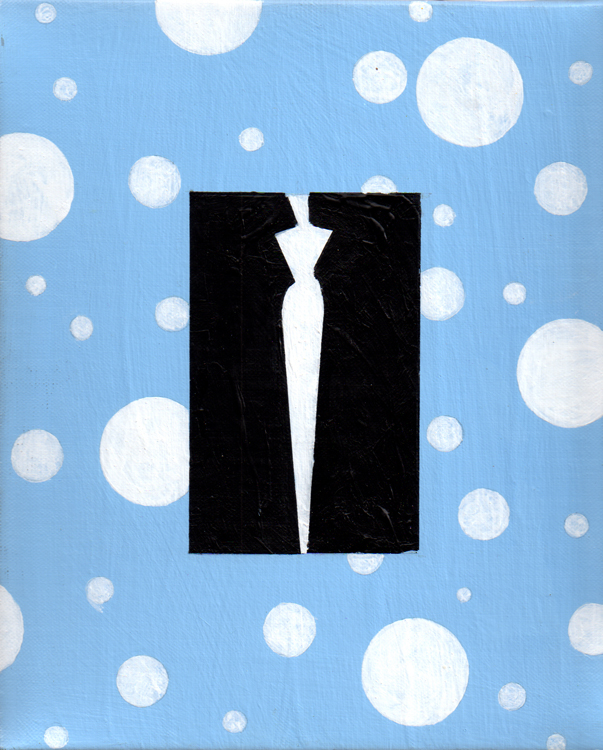}                                         & \includegraphics[align=c,width=2cm,height=2cm]{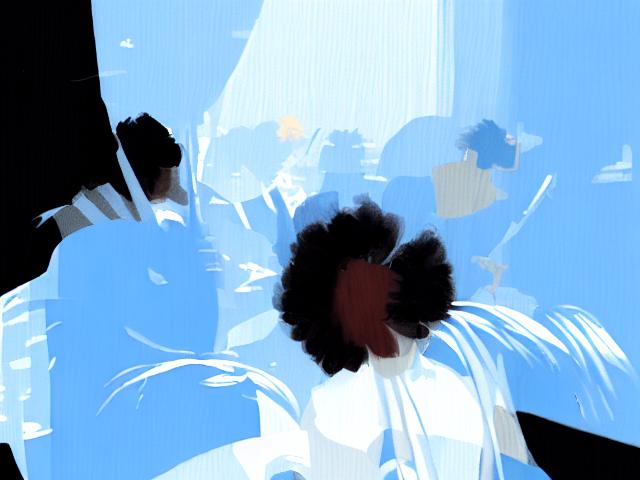} & \includegraphics[align=c,width=2cm,height=2cm]{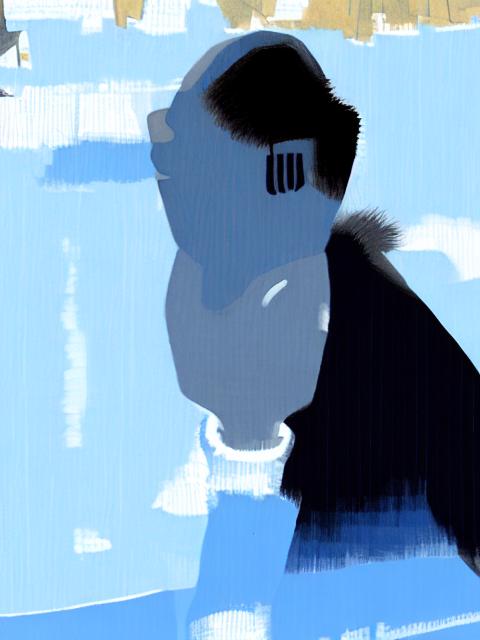} & \includegraphics[align=c,width=2cm,height=2cm]{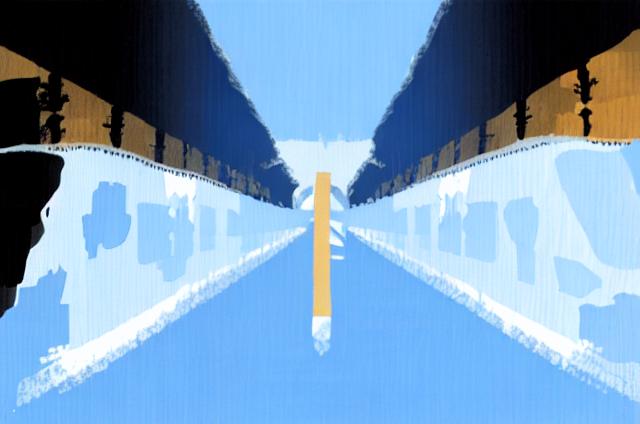}  \\ \hline
\includegraphics[align=c,width=2cm,height=2cm]{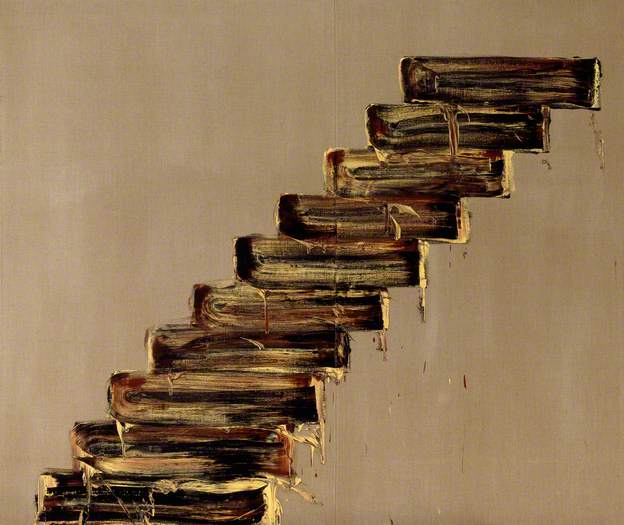}                                         & \includegraphics[align=c,width=2cm,height=2cm]{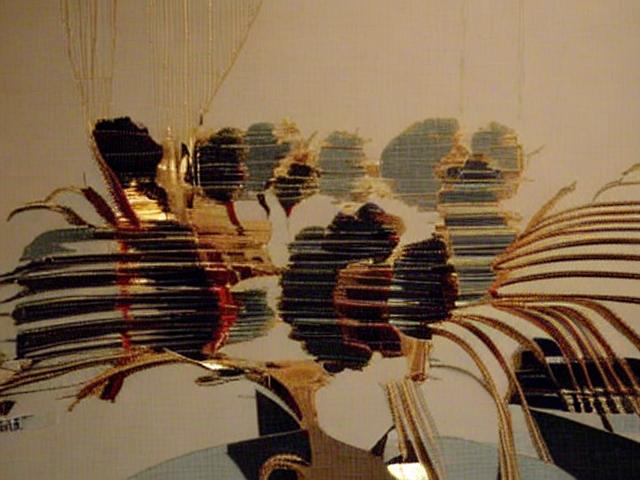} & \includegraphics[align=c,width=2cm,height=2cm]{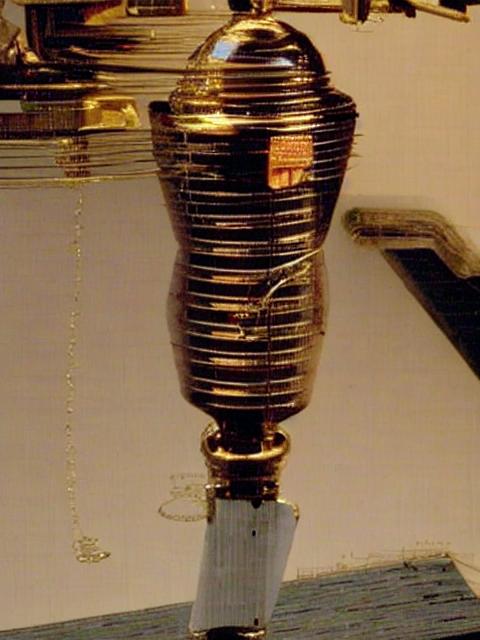} & \includegraphics[align=c,width=2cm,height=2cm]{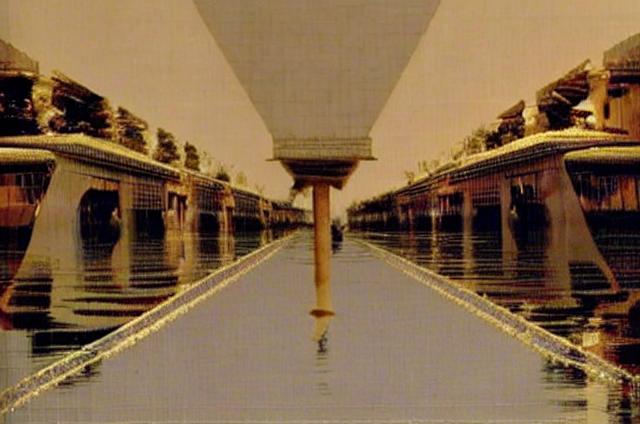}
\end{NiceTabular}%
}
\hfill
\resizebox{\columnwidth}{!}{%
\begin{NiceTabular}{c|c|c|c}
\diagbox{Style \\ Images}{Content \\ Images}
& \includegraphics[align=c,width=2cm,height=2cm]{tables/ground-truth-table/content-images/content-1.jpg} & \includegraphics[align=c,width=2cm,height=2cm]{tables/ground-truth-table/content-images/content-2.jpg} & \includegraphics[align=c,width=2cm,height=2cm]{tables/ground-truth-table/content-images/content-3.jpg}  \\ \hline
 \includegraphics[align=c,width=2cm,height=2cm]{tables/ground-truth-table/wikiart/style-images/style-1.jpg}                                         & \includegraphics[align=c,width=2cm,height=2cm]{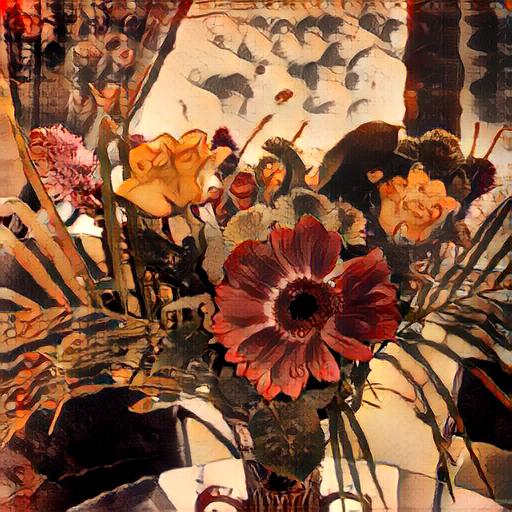} & \includegraphics[align=c,width=2cm,height=2cm]{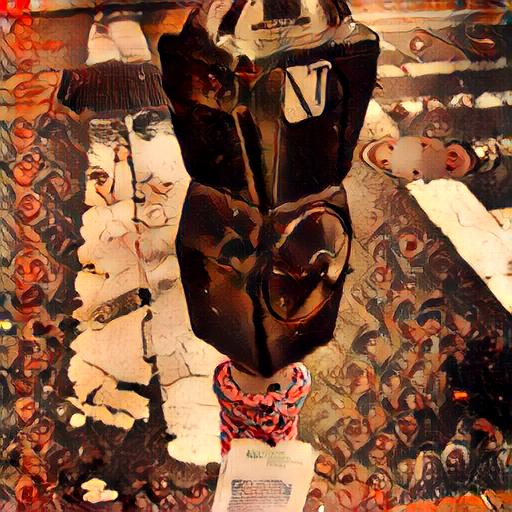} & \includegraphics[align=c,width=2cm,height=2cm]{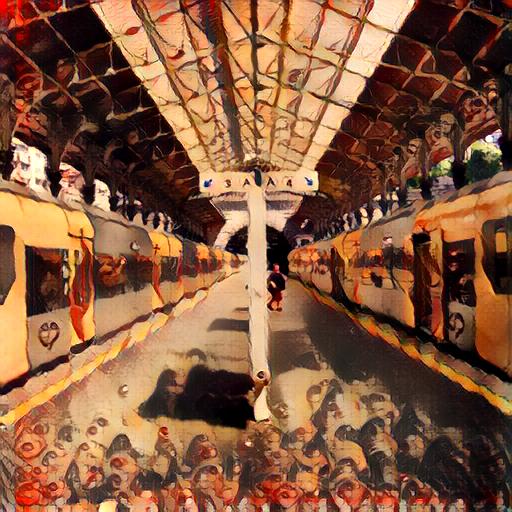} \\ \hline
\includegraphics[align=c,width=2cm,height=2cm]{tables/ground-truth-table/wikiart/style-images/style-2.jpg}                                         & \includegraphics[align=c,width=2cm,height=2cm]{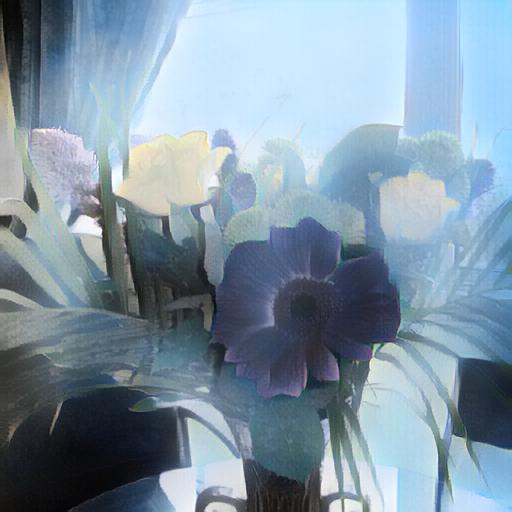} & \includegraphics[align=c,width=2cm,height=2cm]{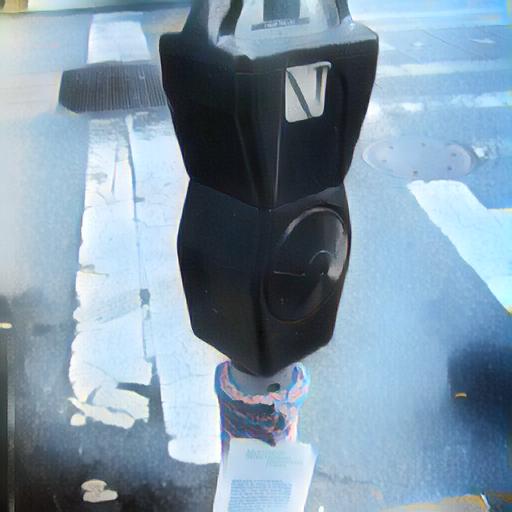} & \includegraphics[align=c,width=2cm,height=2cm]{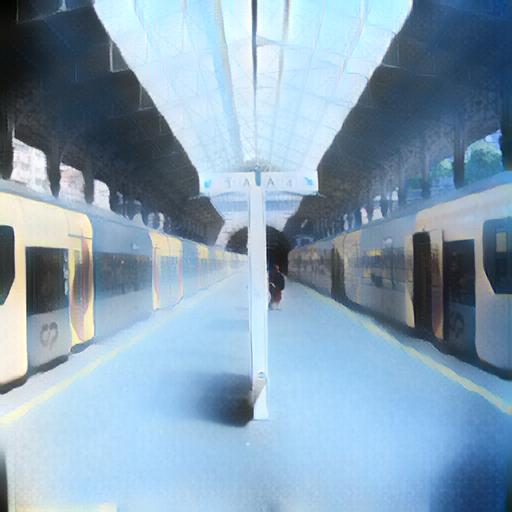}  \\ \hline
\includegraphics[align=c,width=2cm,height=2cm]{tables/ground-truth-table/wikiart/style-images/style-3.jpg}                                         & \includegraphics[align=c,width=2cm,height=2cm]{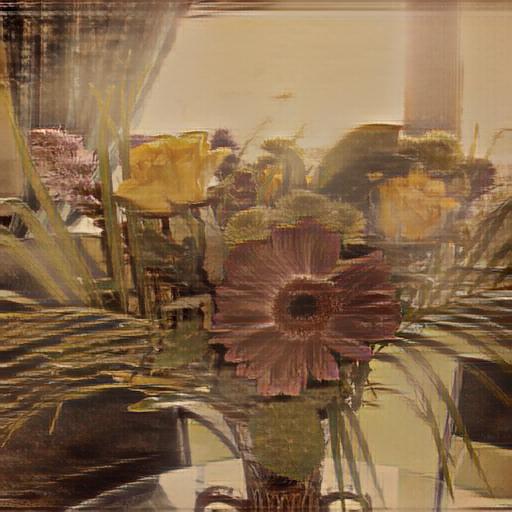} & \includegraphics[align=c,width=2cm,height=2cm]{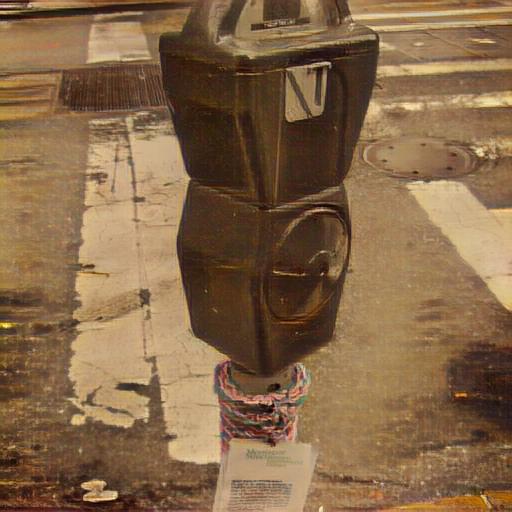} & \includegraphics[align=c,width=2cm,height=2cm]{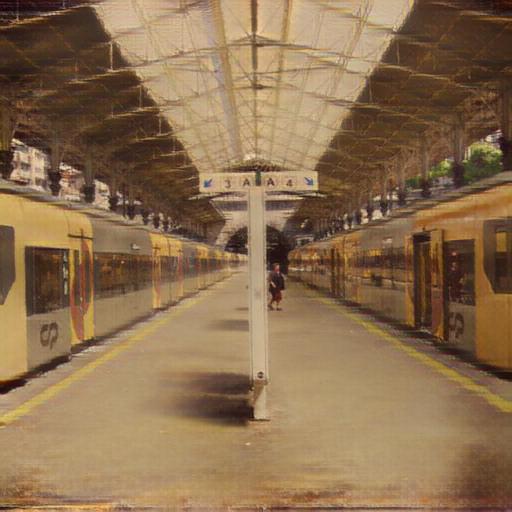}
\end{NiceTabular}%
}
\caption{Examples of content images and style images and their respective style-transfer output images from Styleshot (left) and Mamba-ST (right). The content images were picked from the MS-Coco dataset \cite{mscoco} and the style images were picked from the WikiArt dataset \cite{wikiart-web,tan2016ceci}. Each row is considered as a single cluster.}
\label{tab:wikiart_ground_truth}
\end{table*}

\begin{table*}[ht!]
\small
\resizebox{\columnwidth}{!}{%
\begin{NiceTabular}{c|c|c|c}
\diagbox{Style \\ Images}{Content \\ Images}
& \includegraphics[align=c,width=2cm,height=2cm]{tables/ground-truth-table/content-images/content-1.jpg} & \includegraphics[align=c,width=2cm,height=2cm]{tables/ground-truth-table/content-images/content-2.jpg} & \includegraphics[align=c,width=2cm,height=2cm]{tables/ground-truth-table/content-images/content-3.jpg}  \\ \hline
 \includegraphics[align=c,width=2cm,height=2cm]{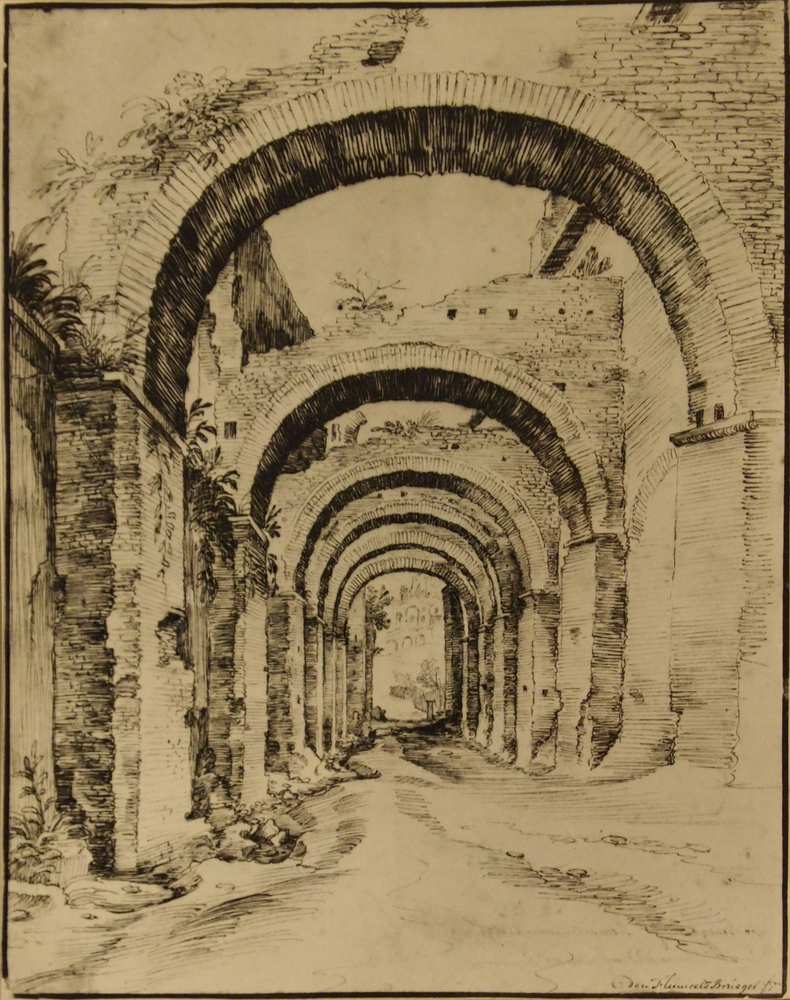}                                         & \includegraphics[align=c,width=2cm,height=2cm]{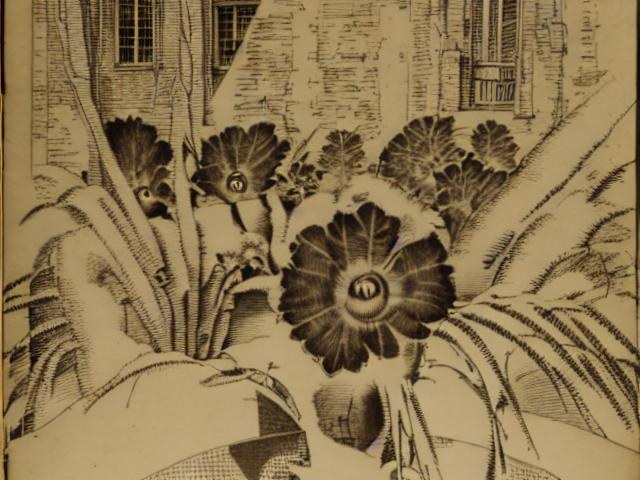} & \includegraphics[align=c,width=2cm,height=2cm]{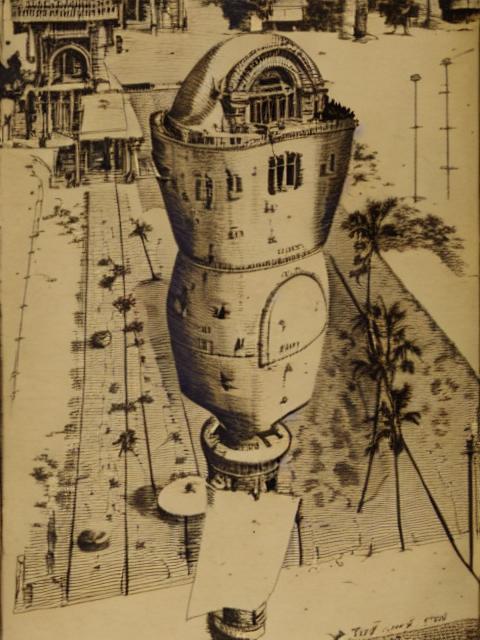} & \includegraphics[align=c,width=2cm,height=2cm]{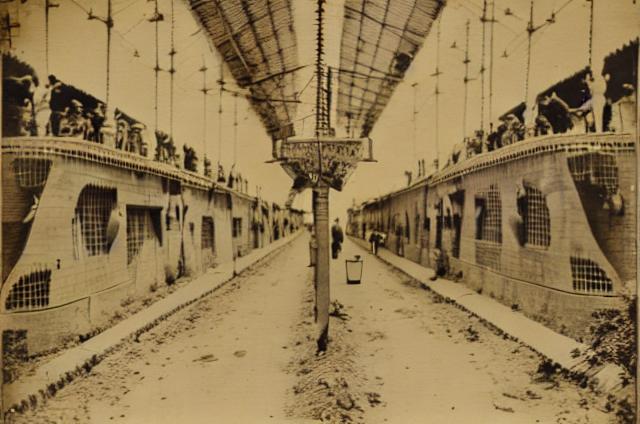} \\ \hline
\includegraphics[align=c,width=2cm,height=2cm]{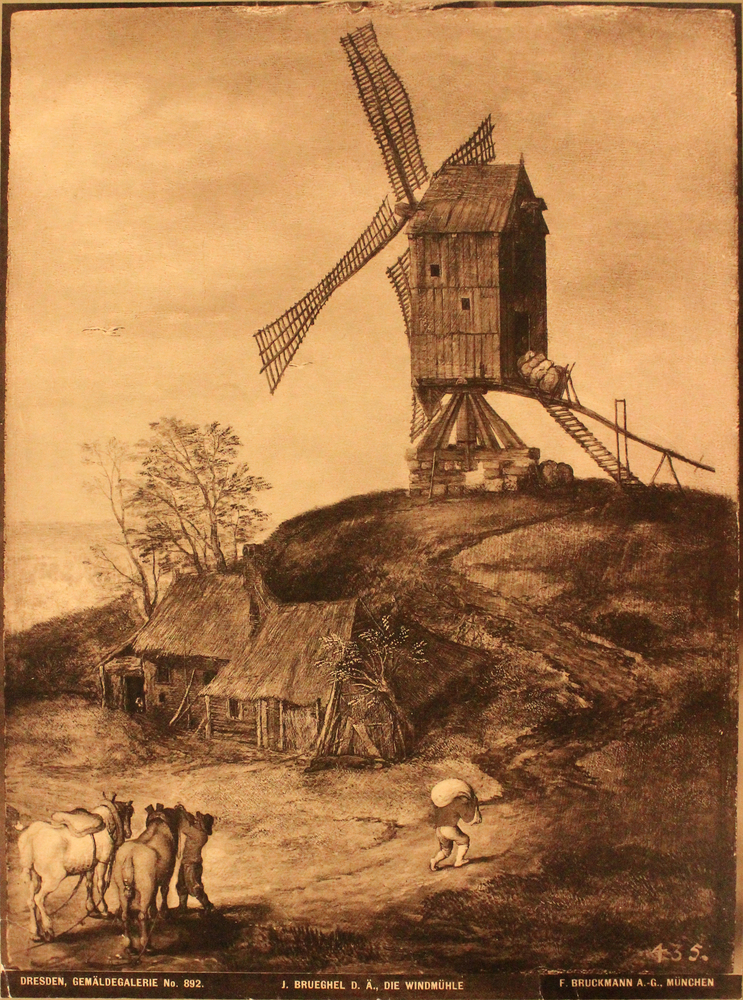}                                         & \includegraphics[align=c,width=2cm,height=2cm]{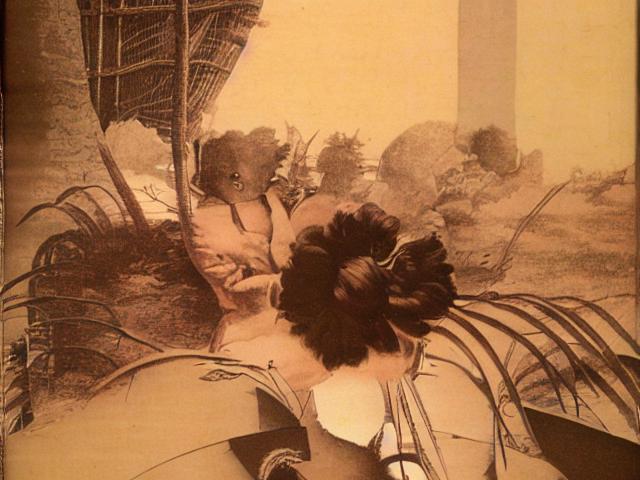} & \includegraphics[align=c,width=2cm,height=2cm]{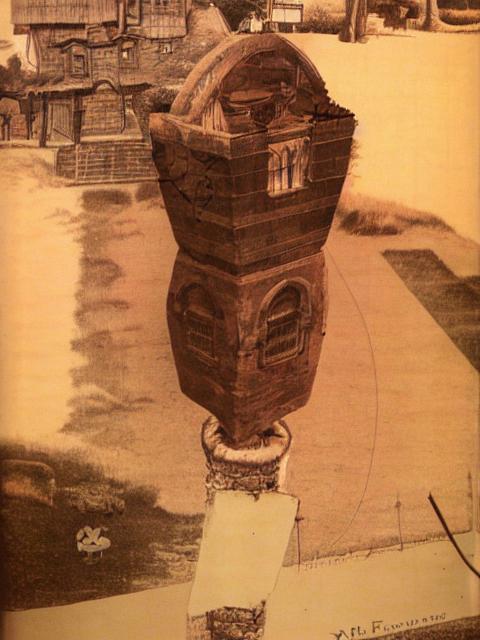} & \includegraphics[align=c,width=2cm,height=2cm]{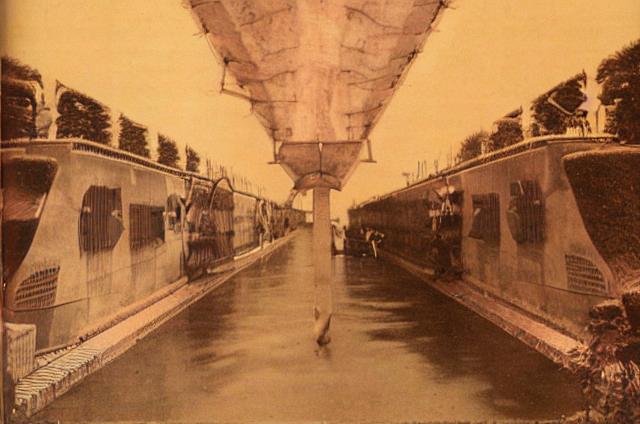}  \\ \hline
\includegraphics[align=c,width=2cm,height=2cm]{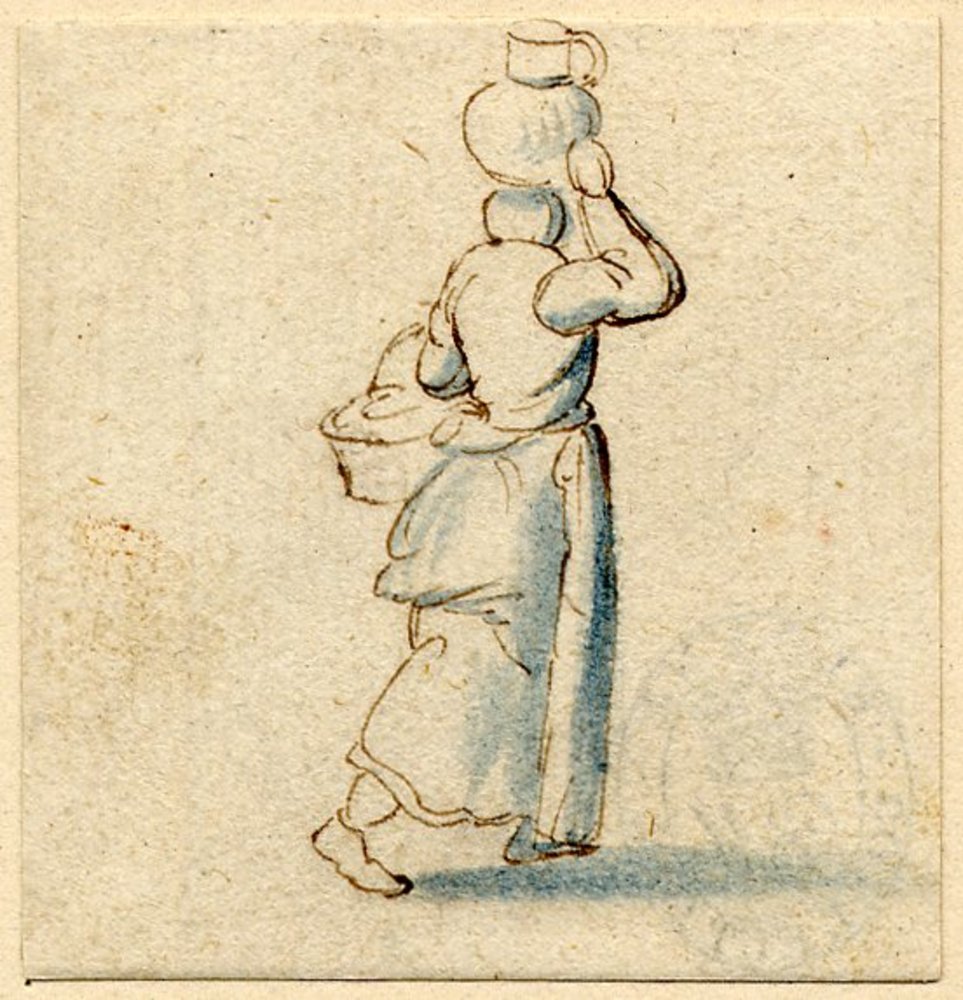}                                         & \includegraphics[align=c,width=2cm,height=2cm]{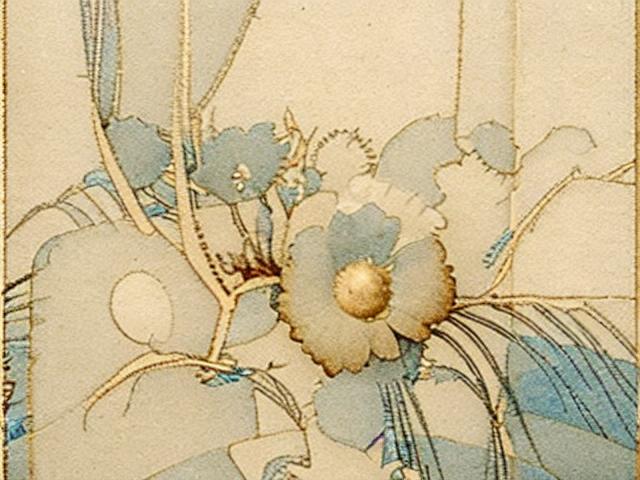} & \includegraphics[align=c,width=2cm,height=2cm]{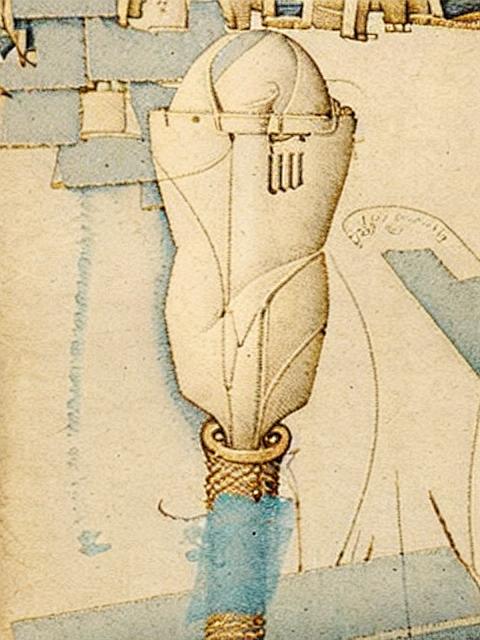} & \includegraphics[align=c,width=2cm,height=2cm]{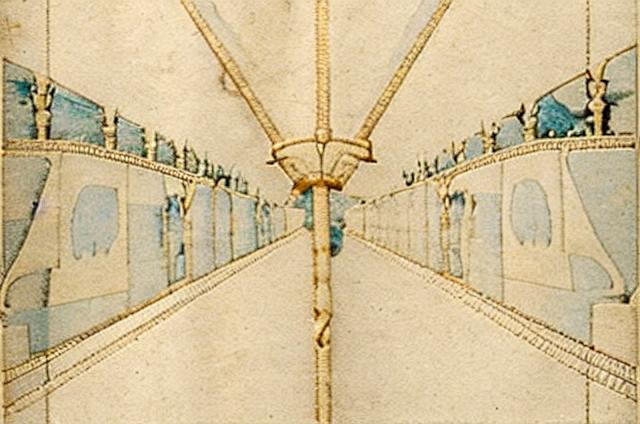}
\end{NiceTabular}%
}
\hfill
\resizebox{\columnwidth}{!}{%
\begin{NiceTabular}{c|c|c|c}
\diagbox{Style \\ Images}{Content \\ Images}
& \includegraphics[align=c,width=2cm,height=2cm]{tables/ground-truth-table/content-images/content-1.jpg} & \includegraphics[align=c,width=2cm,height=2cm]{tables/ground-truth-table/content-images/content-2.jpg} & \includegraphics[align=c,width=2cm,height=2cm]{tables/ground-truth-table/content-images/content-3.jpg}  \\ \hline
 \includegraphics[align=c,width=2cm,height=2cm]{tables/ground-truth-table/brueghel/style-images/style-1.jpg}                                         & \includegraphics[align=c,width=2cm,height=2cm]{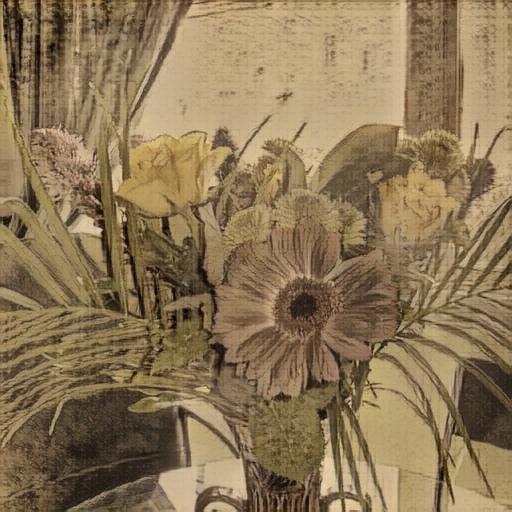} & \includegraphics[align=c,width=2cm,height=2cm]{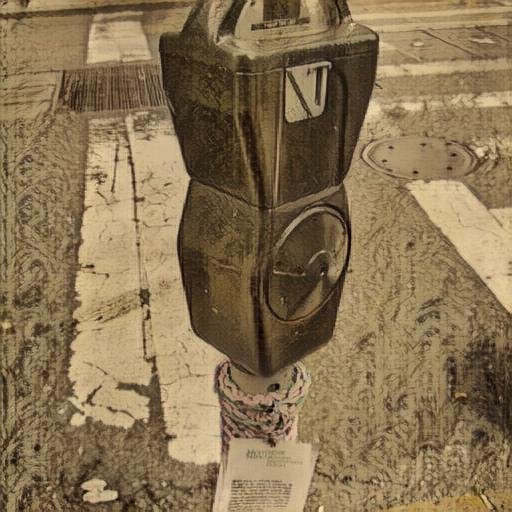} & \includegraphics[align=c,width=2cm,height=2cm]{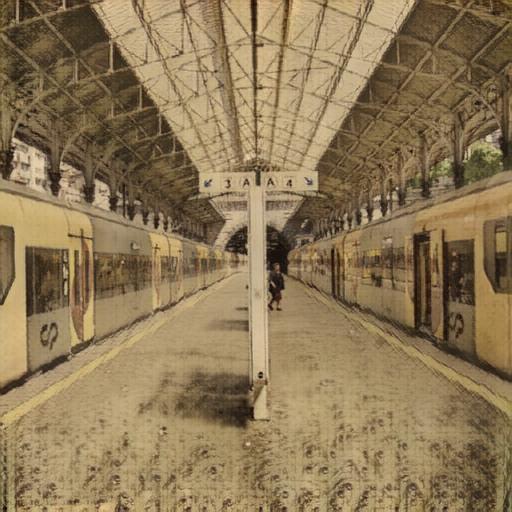} \\ \hline
\includegraphics[align=c,width=2cm,height=2cm]{tables/ground-truth-table/brueghel/style-images/style-2.jpg}                                         & \includegraphics[align=c,width=2cm,height=2cm]{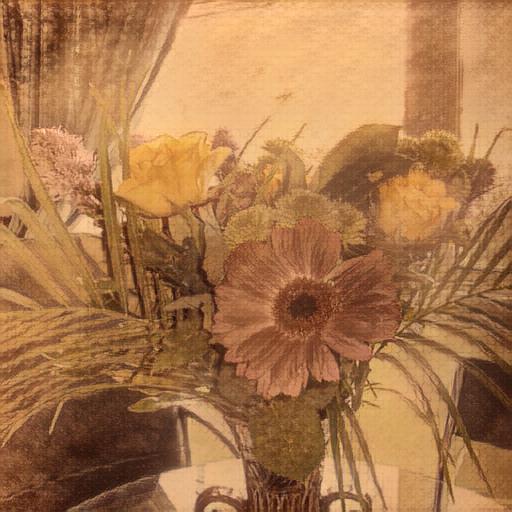} & \includegraphics[align=c,width=2cm,height=2cm]{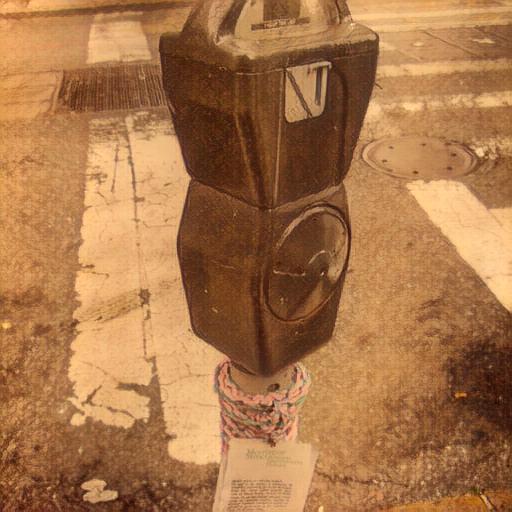} & \includegraphics[align=c,width=2cm,height=2cm]{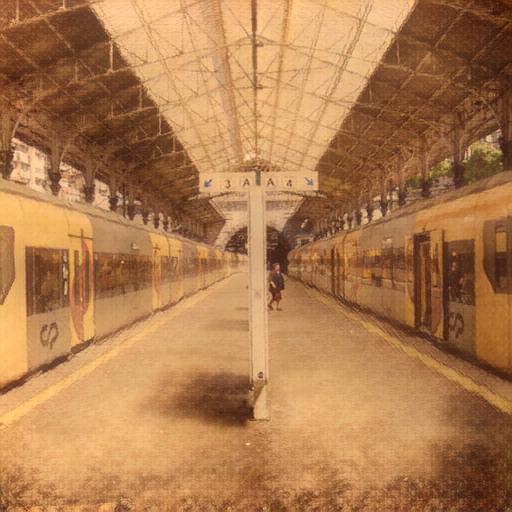}  \\ \hline
\includegraphics[align=c,width=2cm,height=2cm]{tables/ground-truth-table/brueghel/style-images/style-3.jpg}                                         & \includegraphics[align=c,width=2cm,height=2cm]{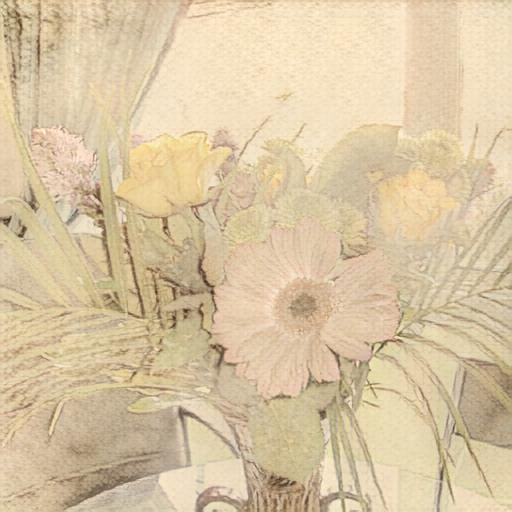} & \includegraphics[align=c,width=2cm,height=2cm]{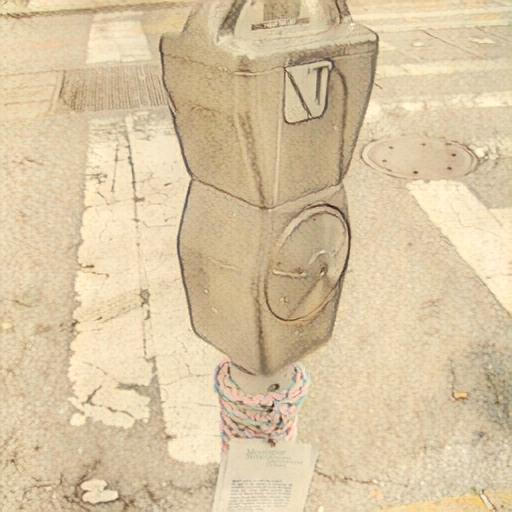} & \includegraphics[align=c,width=2cm,height=2cm]{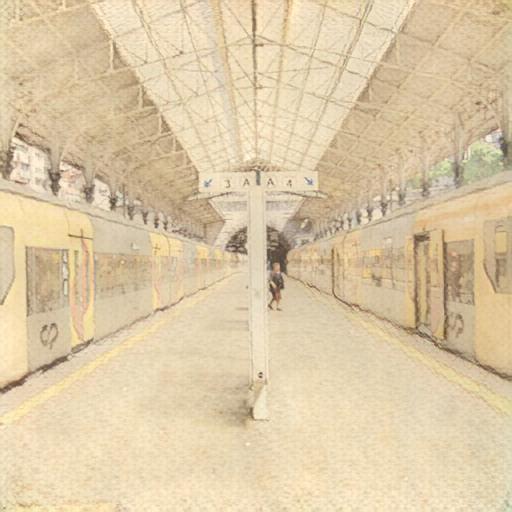}
\end{NiceTabular}%
}
\caption{Examples of content images and style images and their respective style-transfer output images from Styleshot (left) and Mamba-ST (right). The content images were picked from the MS-Coco dataset \cite{mscoco} and the style images were picked from the Brueghel dataset \cite{brueghel-dataset}. Each row is considered as a single cluster.}
\label{tab:brueghel_ground_truth}
\end{table*}

\begin{table*}[ht!]
\small
\resizebox{\columnwidth}{!}{%
\begin{NiceTabular}{c|c|c|c}
\diagbox{Style \\ Images}{Content \\ Images}
& \includegraphics[align=c,width=2cm,height=2cm]{tables/ground-truth-table/content-images/content-1.jpg} & \includegraphics[align=c,width=2cm,height=2cm]{tables/ground-truth-table/content-images/content-2.jpg} & \includegraphics[align=c,width=2cm,height=2cm]{tables/ground-truth-table/content-images/content-3.jpg}  \\ \hline
 \includegraphics[align=c,width=2cm,height=2cm]{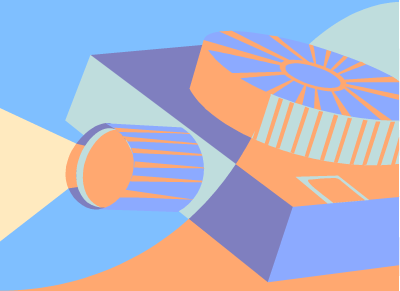}                                         & \includegraphics[align=c,width=2cm,height=2cm]{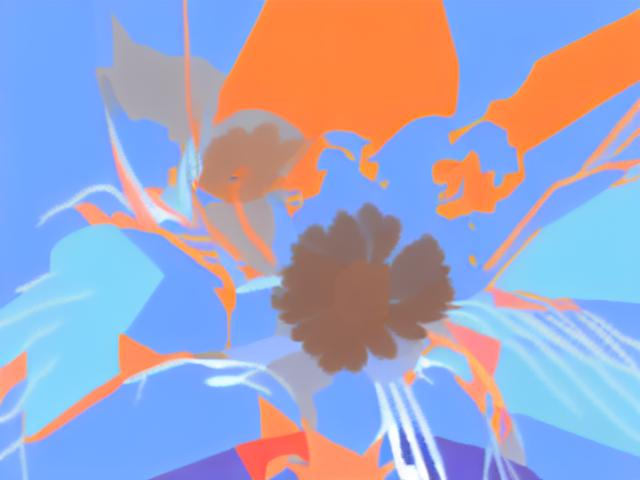} & \includegraphics[align=c,width=2cm,height=2cm]{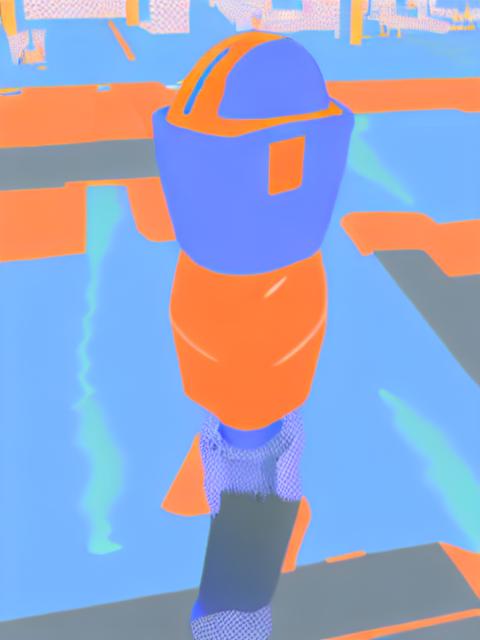} & \includegraphics[align=c,width=2cm,height=2cm]{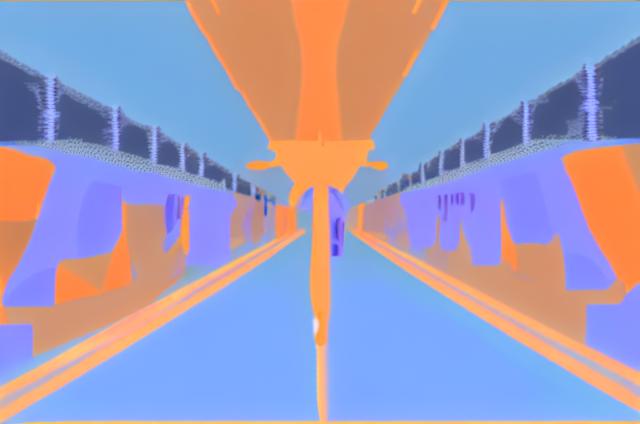} \\ \hline
\includegraphics[align=c,width=2cm,height=2cm]{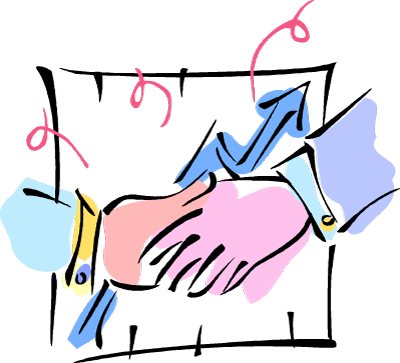}                                         & \includegraphics[align=c,width=2cm,height=2cm]{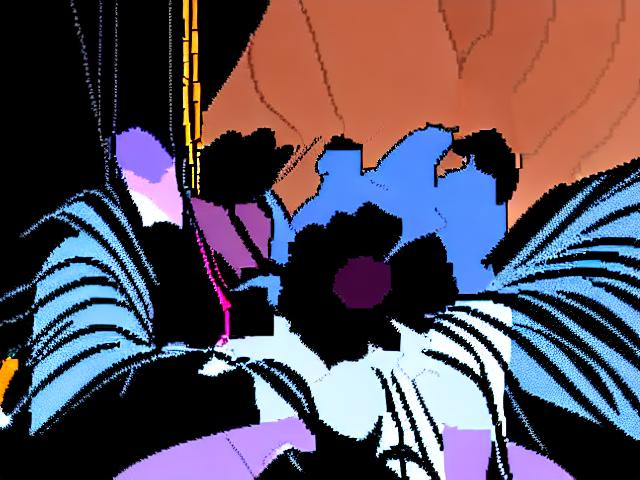} & \includegraphics[align=c,width=2cm,height=2cm]{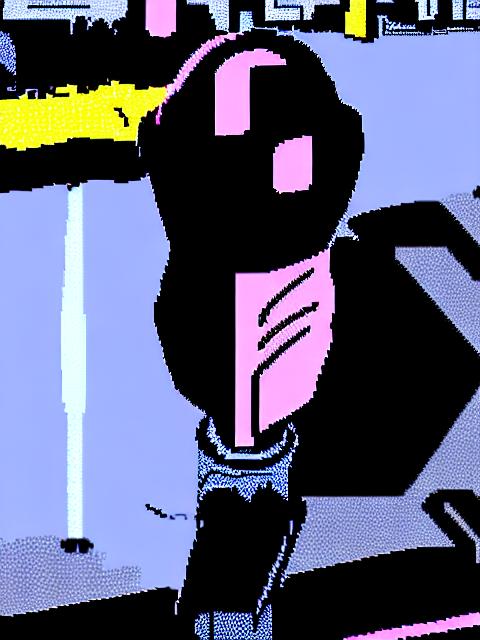} & \includegraphics[align=c,width=2cm,height=2cm]{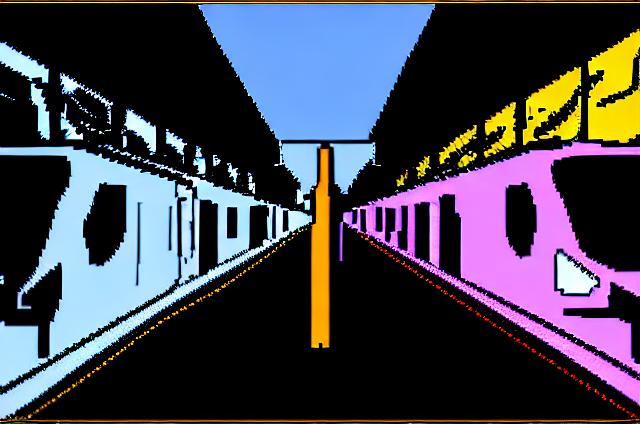}  \\ \hline
\includegraphics[align=c,width=2cm,height=2cm]{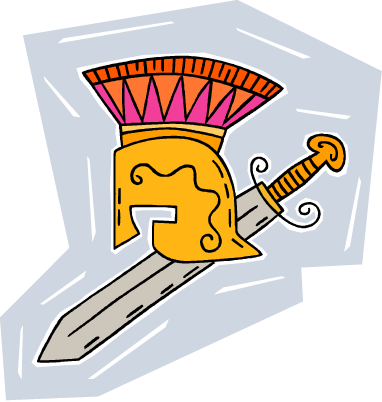}                                         & \includegraphics[align=c,width=2cm,height=2cm]{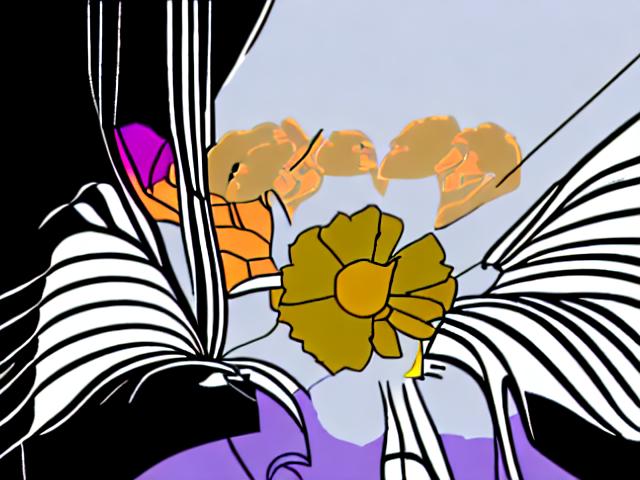} & \includegraphics[align=c,width=2cm,height=2cm]{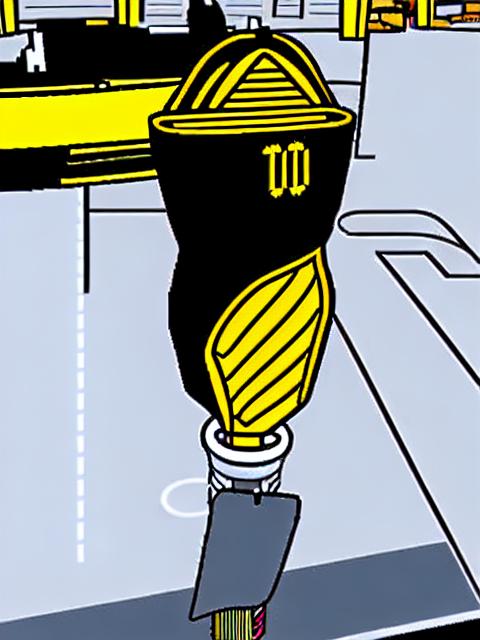} & \includegraphics[align=c,width=2cm,height=2cm]{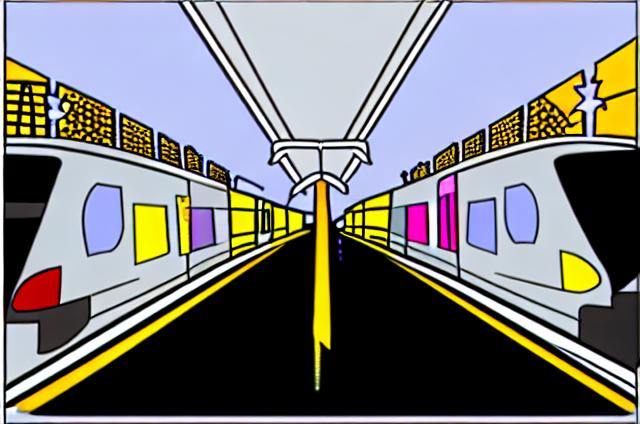}
\end{NiceTabular}%
}
\hfill
\resizebox{\columnwidth}{!}{%
\begin{NiceTabular}{c|c|c|c}
\diagbox{Style \\ Images}{Content \\ Images}
& \includegraphics[align=c,width=2cm,height=2cm]{tables/ground-truth-table/content-images/content-1.jpg} & \includegraphics[align=c,width=2cm,height=2cm]{tables/ground-truth-table/content-images/content-2.jpg} & \includegraphics[align=c,width=2cm,height=2cm]{tables/ground-truth-table/content-images/content-3.jpg}  \\ \hline
 \includegraphics[align=c,width=2cm,height=2cm]{tables/ground-truth-table/clip-art/style-images/style-1.png}                                         & \includegraphics[align=c,width=2cm,height=2cm]{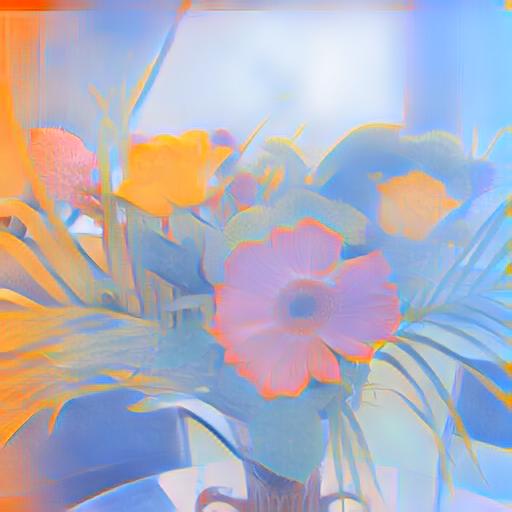} & \includegraphics[align=c,width=2cm,height=2cm]{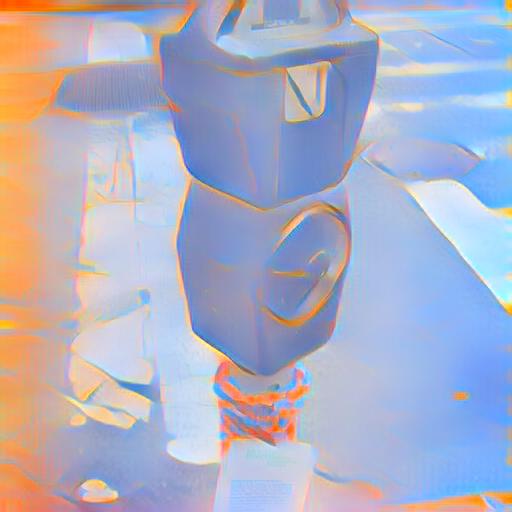} & \includegraphics[align=c,width=2cm,height=2cm]{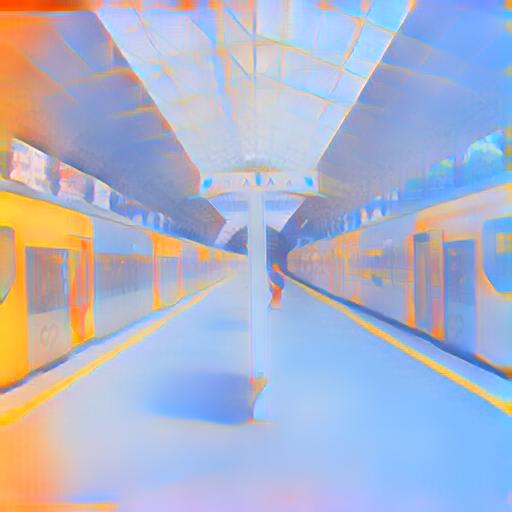} \\ \hline
\includegraphics[align=c,width=2cm,height=2cm]{tables/ground-truth-table/clip-art/style-images/style-2.png}                                         & \includegraphics[align=c,width=2cm,height=2cm]{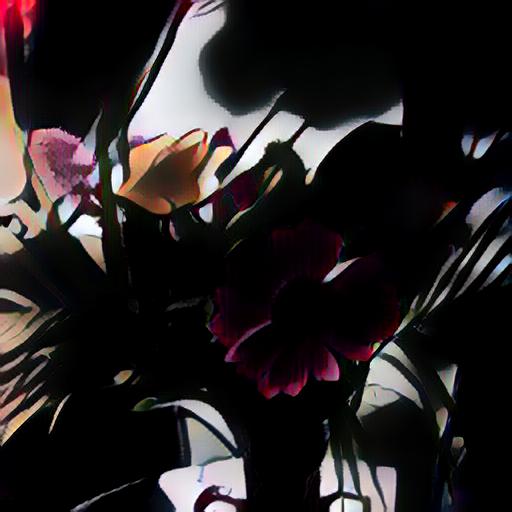} & \includegraphics[align=c,width=2cm,height=2cm]{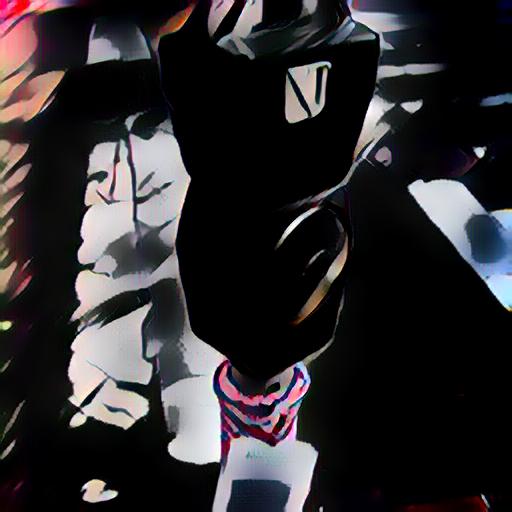} & \includegraphics[align=c,width=2cm,height=2cm]{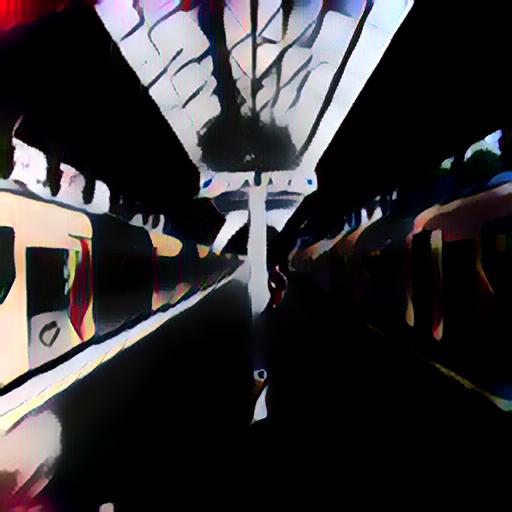}  \\ \hline
\includegraphics[align=c,width=2cm,height=2cm]{tables/ground-truth-table/clip-art/style-images/style-3.png}                                         & \includegraphics[align=c,width=2cm,height=2cm]{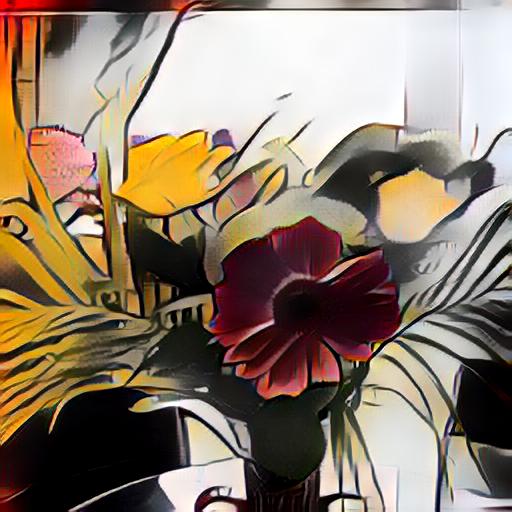} & \includegraphics[align=c,width=2cm,height=2cm]{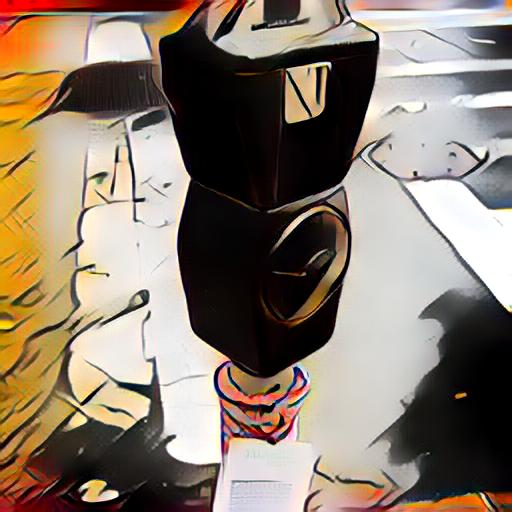} & \includegraphics[align=c,width=2cm,height=2cm]{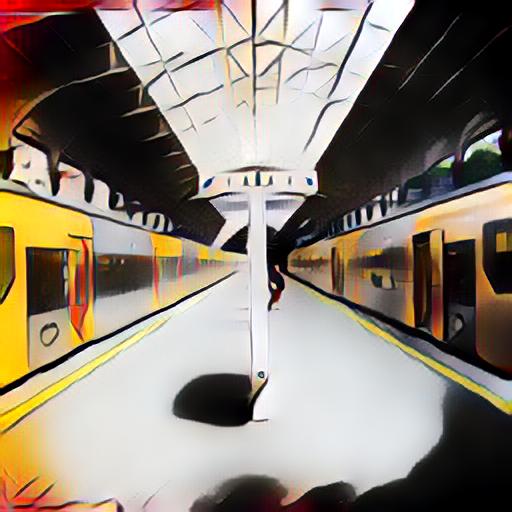}
\end{NiceTabular}%
}
\caption{Examples of content images and style images and their respective style-transfer output images from Styleshot (left) and Mamba-ST (right). The content images were picked from the MS-Coco dataset \cite{mscoco} and the style images were picked from the Clip art illustration dataset \cite{clip-art-dataset, art-explo}. Each row is considered as a single cluster.}
\label{tab:clip-art_ground_truth}
\end{table*}

\begin{figure*}[!t]
 \begin{subfigure}[t]{.48\textwidth}
    \centering
    \includegraphics[width=\textwidth]{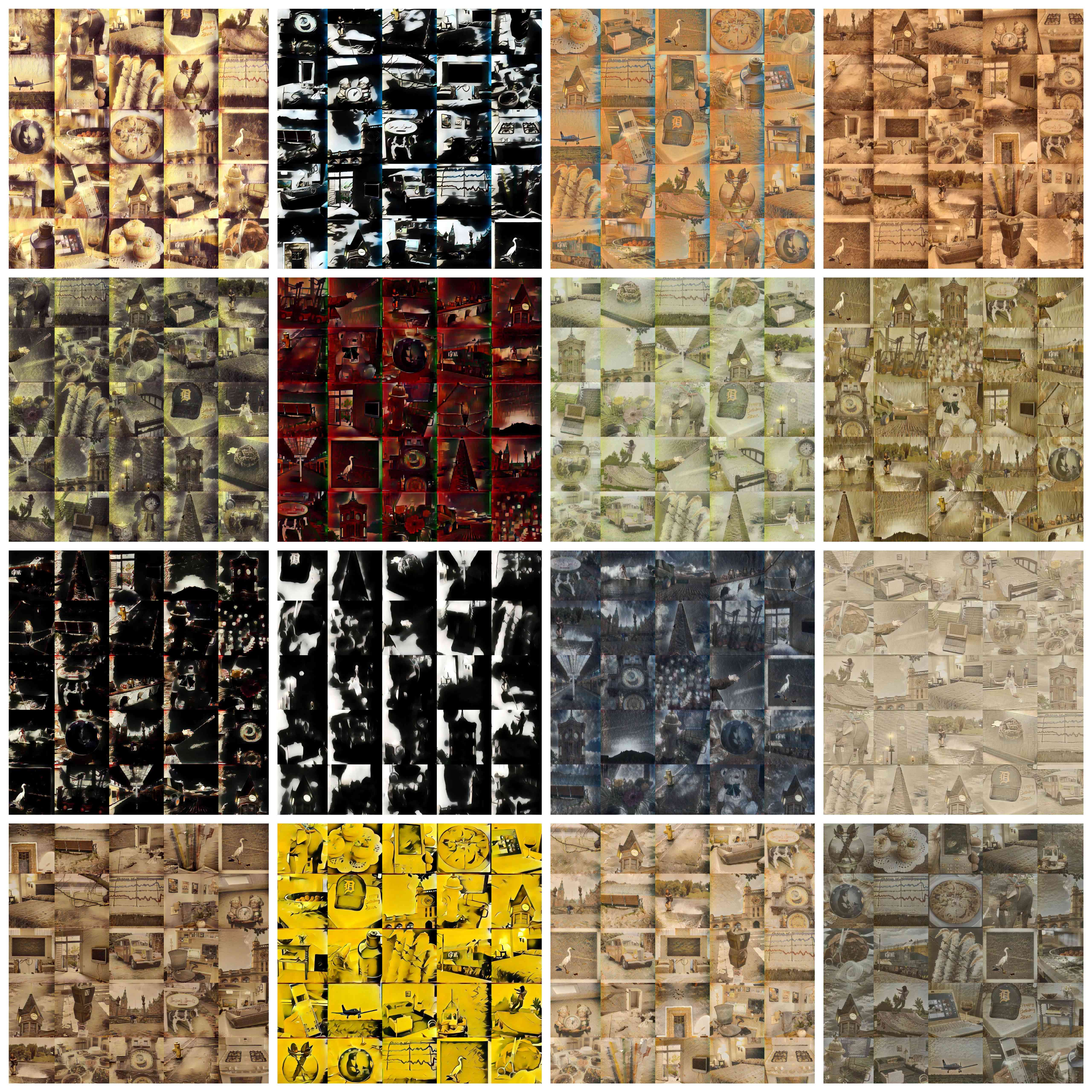}
    \caption{Ground Truth}
  \end{subfigure}
   \hfill
 \begin{subfigure}[t]{0.48\textwidth}
    \centering
    \includegraphics[width=\textwidth]{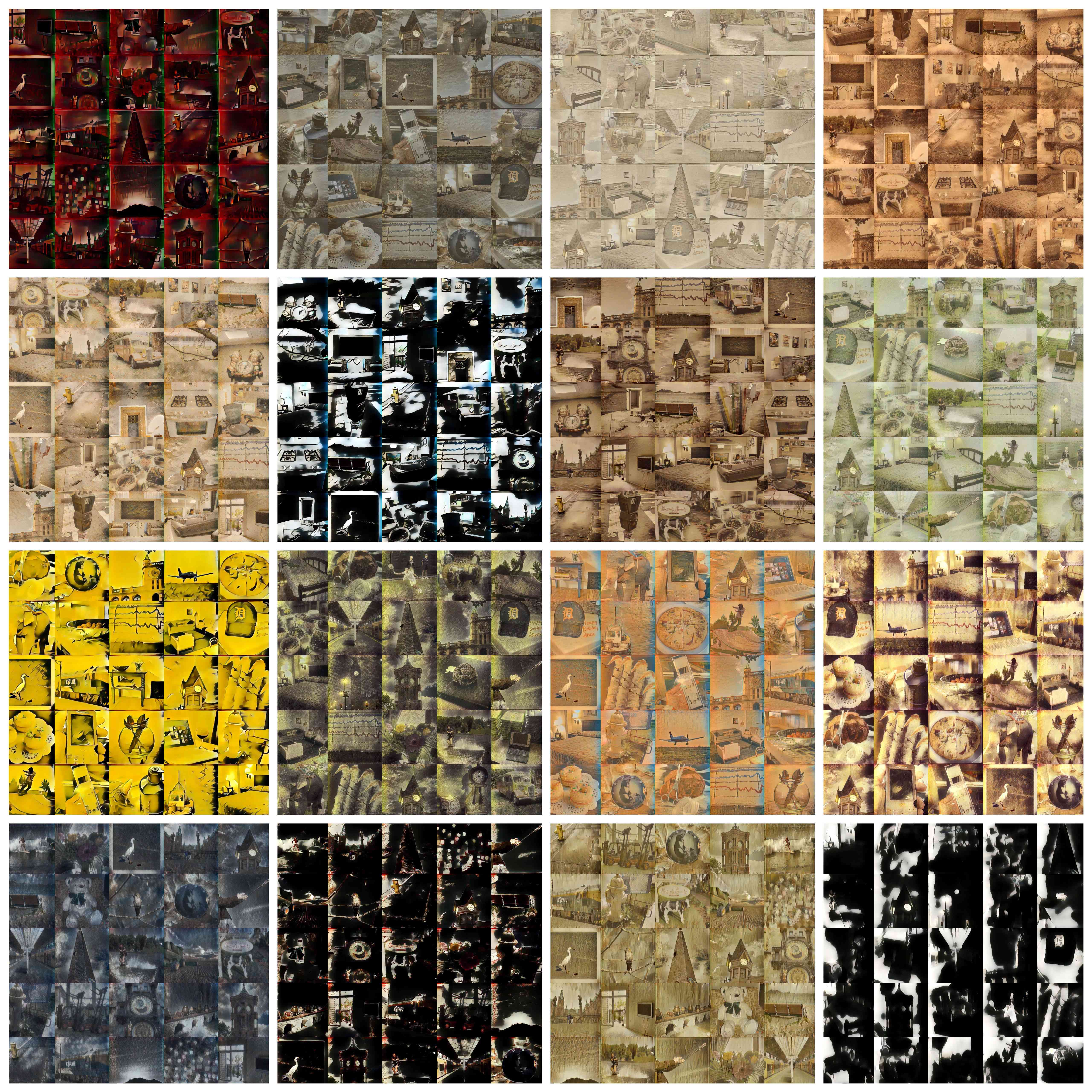}
    \caption {$F_{Stytr2}$ clusters}

  \end{subfigure}
  \begin{subfigure}[t]{.48\textwidth}
    \centering
    \includegraphics[width=\textwidth]{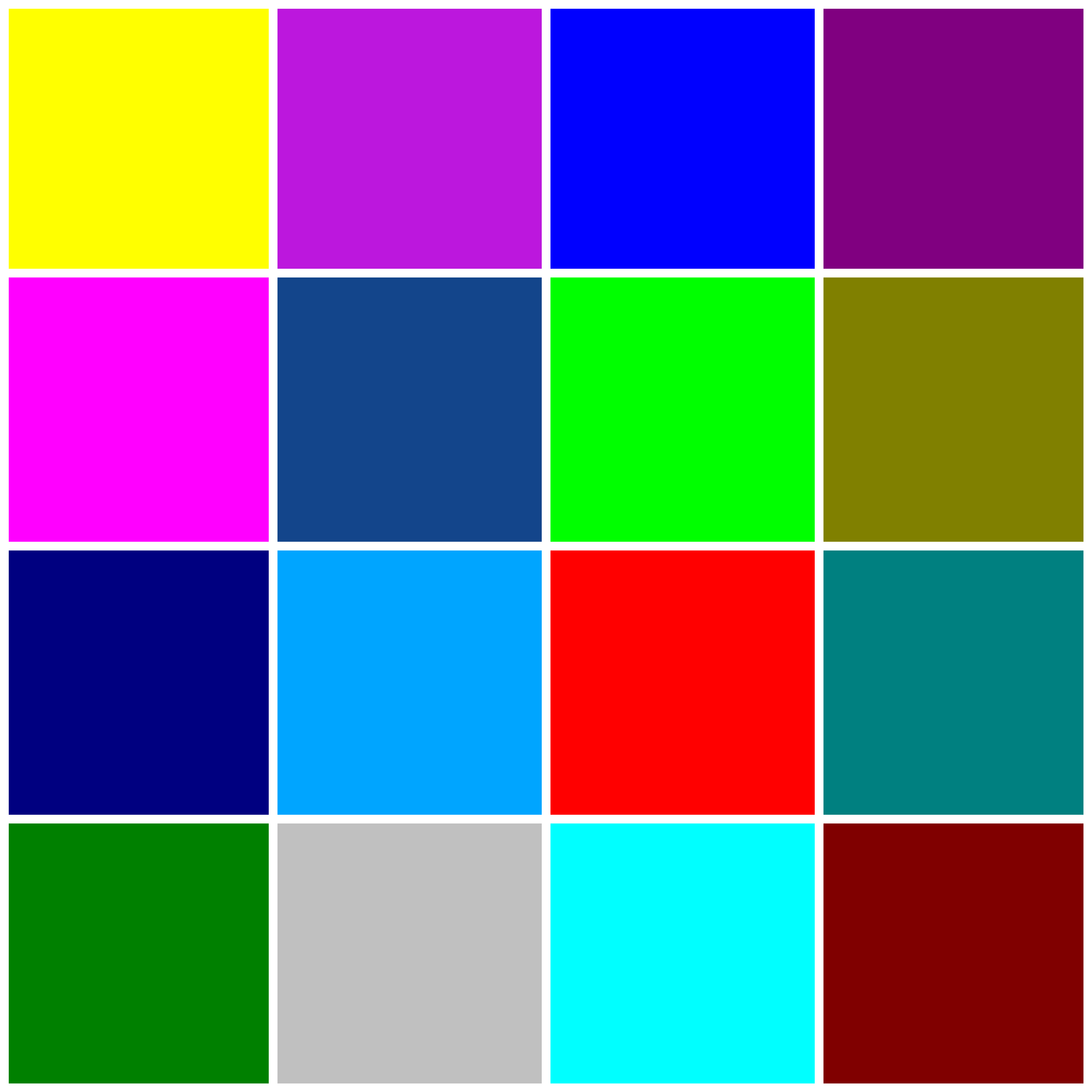}
    \caption{Ground truth color representations per cluster}
  \end{subfigure}
   \hfill
 \begin{subfigure}[t]{.48\textwidth}
    \centering
    \includegraphics[width= .96\textwidth]{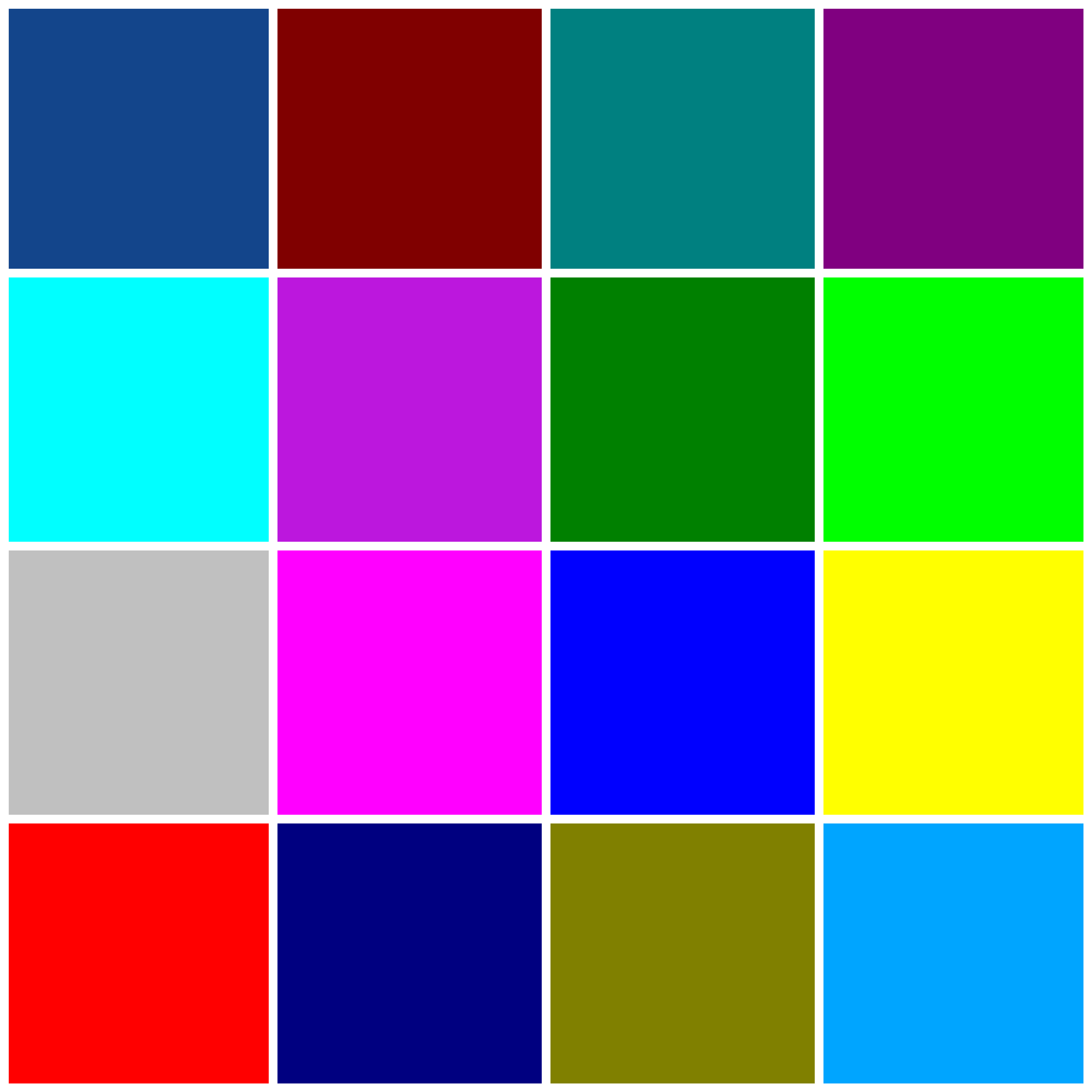}
    \caption {$F_{Stytr2}$ color representations per cluster}
  \end{subfigure}

   \caption{Qualitative comparison of 16 clusters on the Mixed Styleshot Curated (MMC-4k) ground truth and clusters obtained from $F_{stytr2}$. (a) and (b) showcase the artworks present in the two types of clusterings. (c) and (d) show the color representation associated with the ground truth for both clusterings. We obtain a perfect clustering with the Mixed Mamba Curated dataset.}
  \label{fig:qual_fig_mix_mamba}
\end{figure*}

\begin{figure*}[t]
    \centering
    \includegraphics[width=0.9\textwidth]{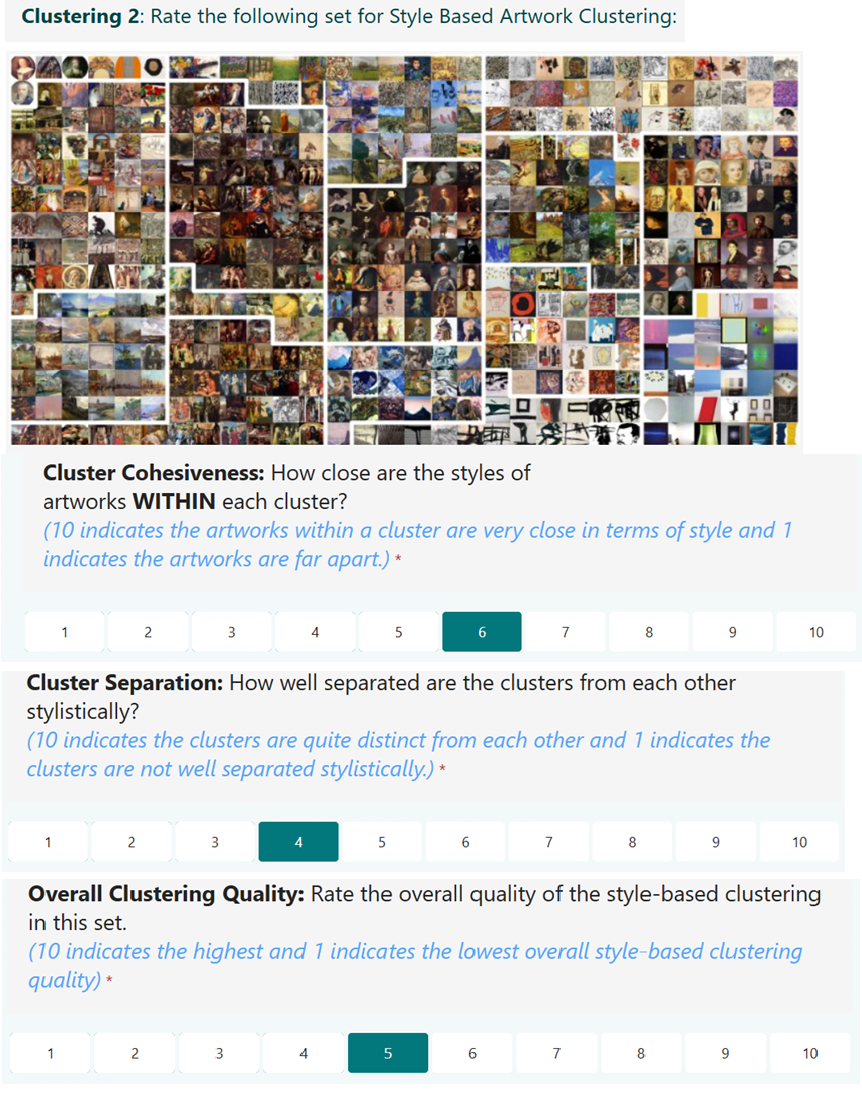}
    \caption{Example figure for the questions asked for a specific clustering in the clustering survey.}
    \label{fig:surv_exmpl}
\end{figure*}
\begin{figure*}[!t]
 \begin{subfigure}[t]{.33\textwidth}
    \centering
    \includegraphics[width=\textwidth]{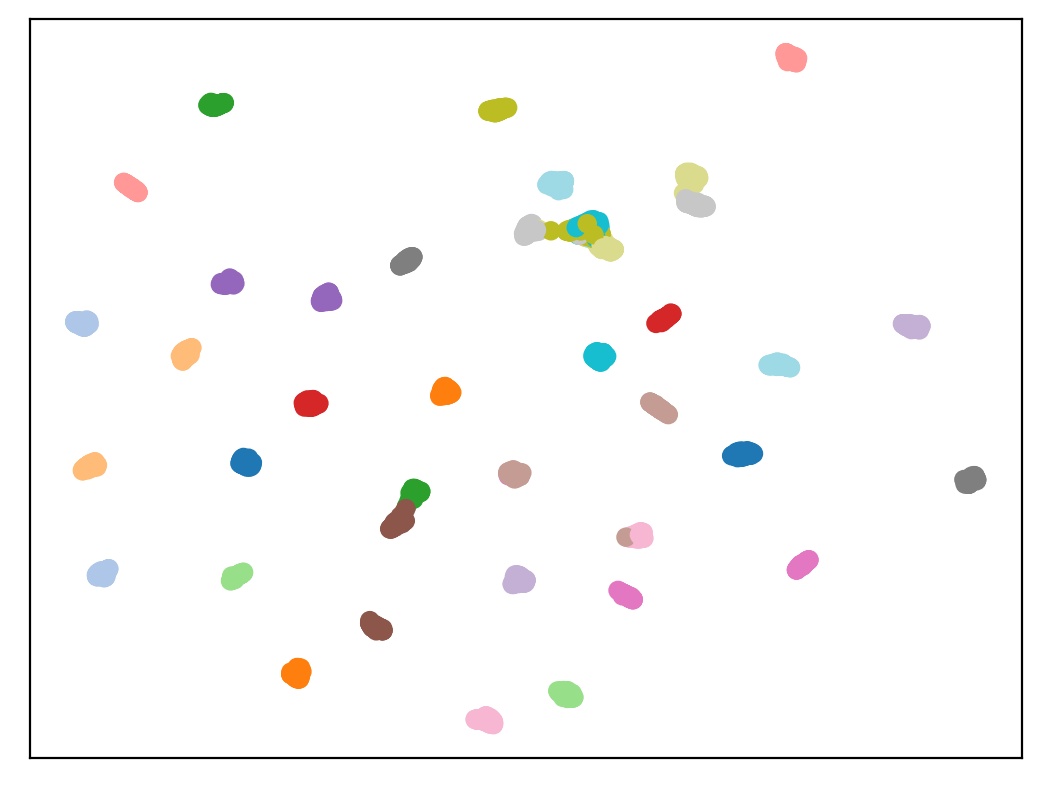}
    \caption{$F_{Gram}$ with K-Means}
  \end{subfigure}
   \hfill
 \begin{subfigure}[t]{0.33\textwidth}
    \centering
    \includegraphics[width=\textwidth]{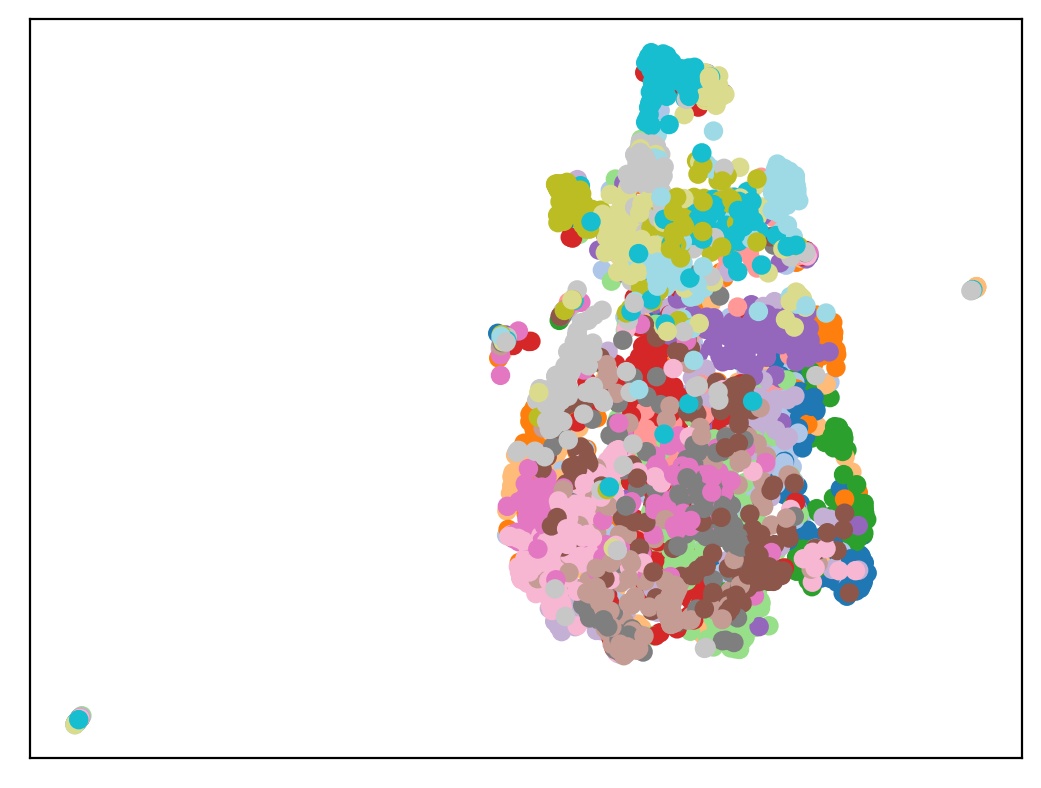}
    \caption {$F_{StyleCap}$ with K-Means}
\hfill
  \end{subfigure}
  \begin{subfigure}[t]{.33\textwidth}
    \centering
    \includegraphics[width=\textwidth]{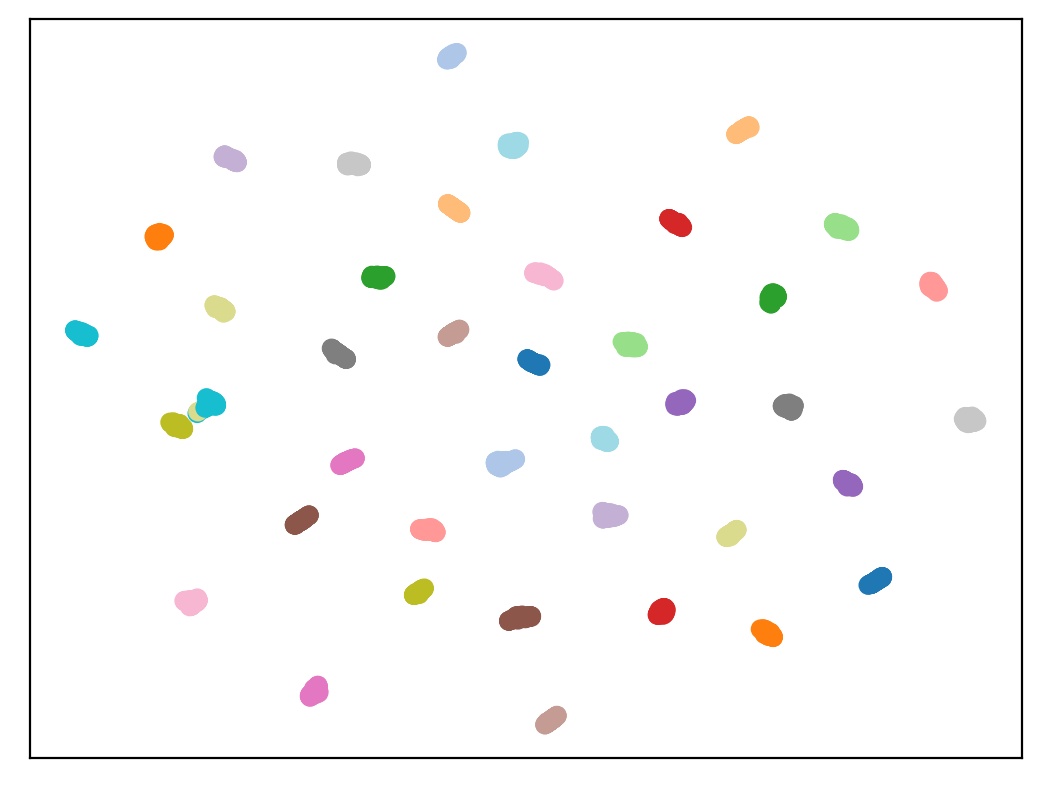}
    \caption{$F_{StyleShot}$ with K-Means}
  \end{subfigure}

 \begin{subfigure}[t]{.33\textwidth}
    \centering
    \includegraphics[width=\textwidth]{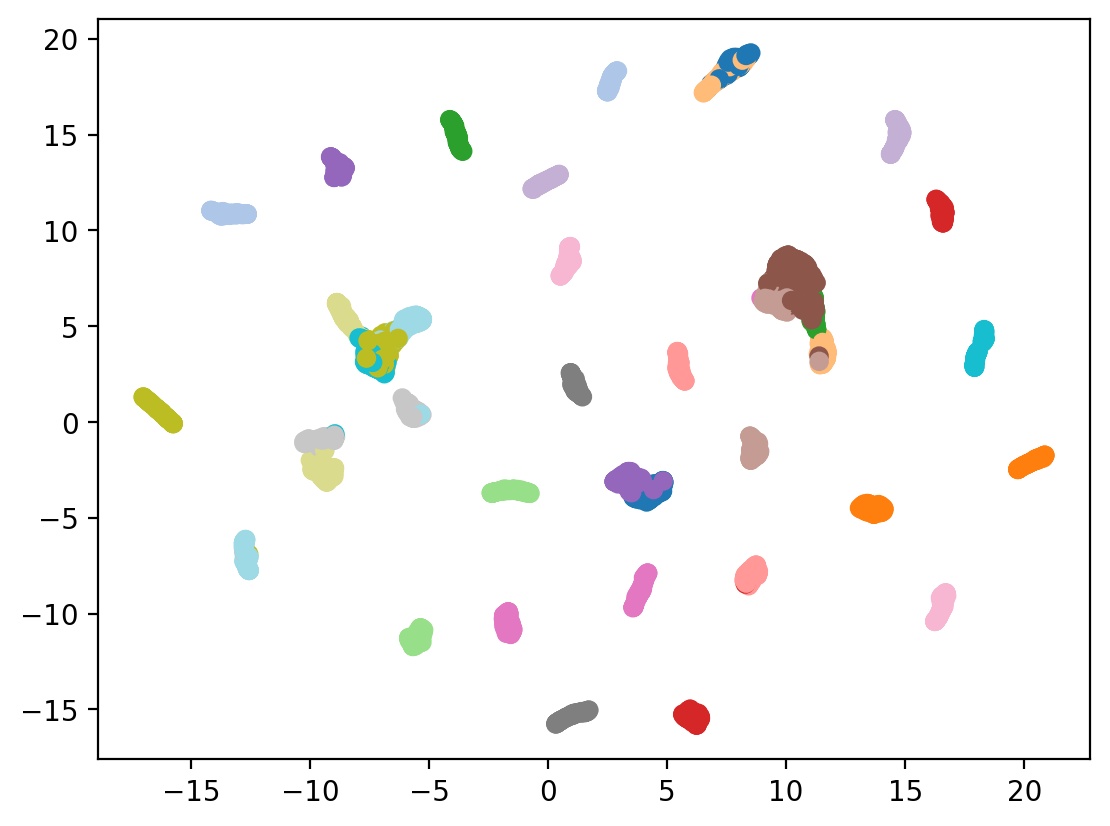}
    \caption{$F_{Gram}$ with DEC}
  \end{subfigure}
   \hfill
 \begin{subfigure}[t]{0.33\textwidth}
    \centering
    \includegraphics[width=\textwidth]{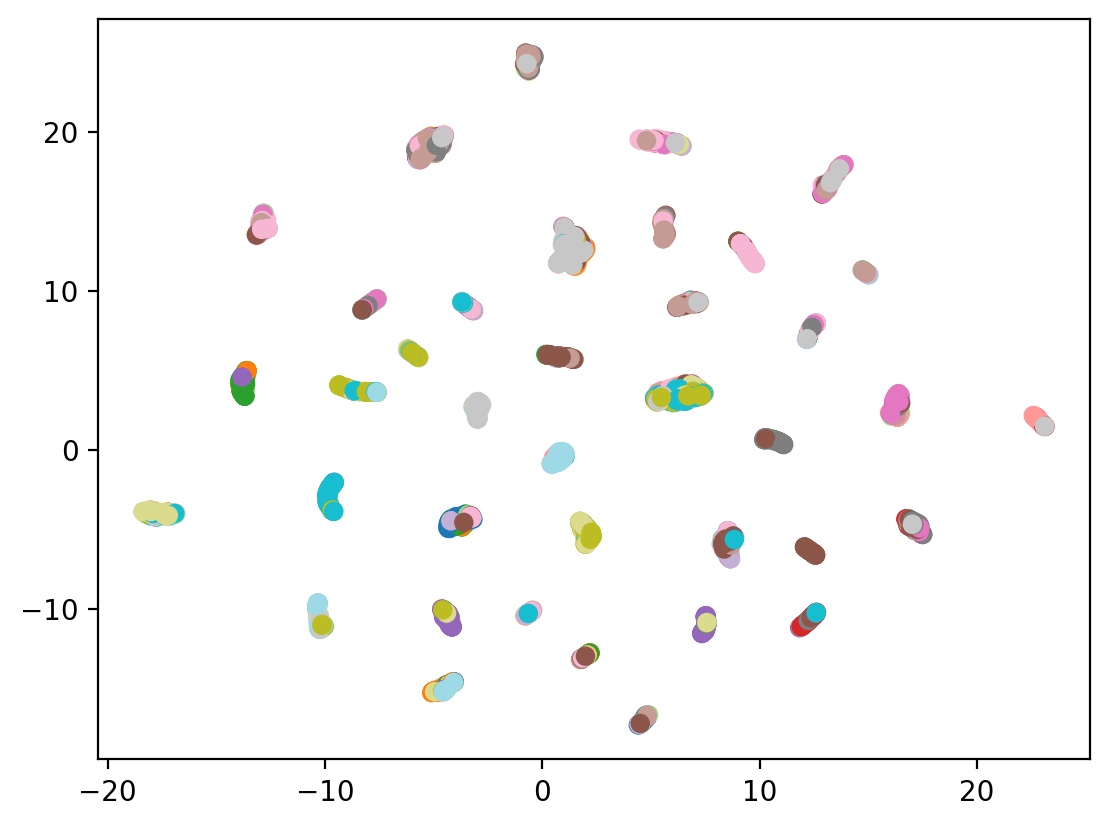}
    \caption {$F_{StyleCap}$ with DEC}
\hfill
  \end{subfigure}
  \begin{subfigure}[t]{.33\textwidth}
    \centering
    \includegraphics[width=\textwidth]{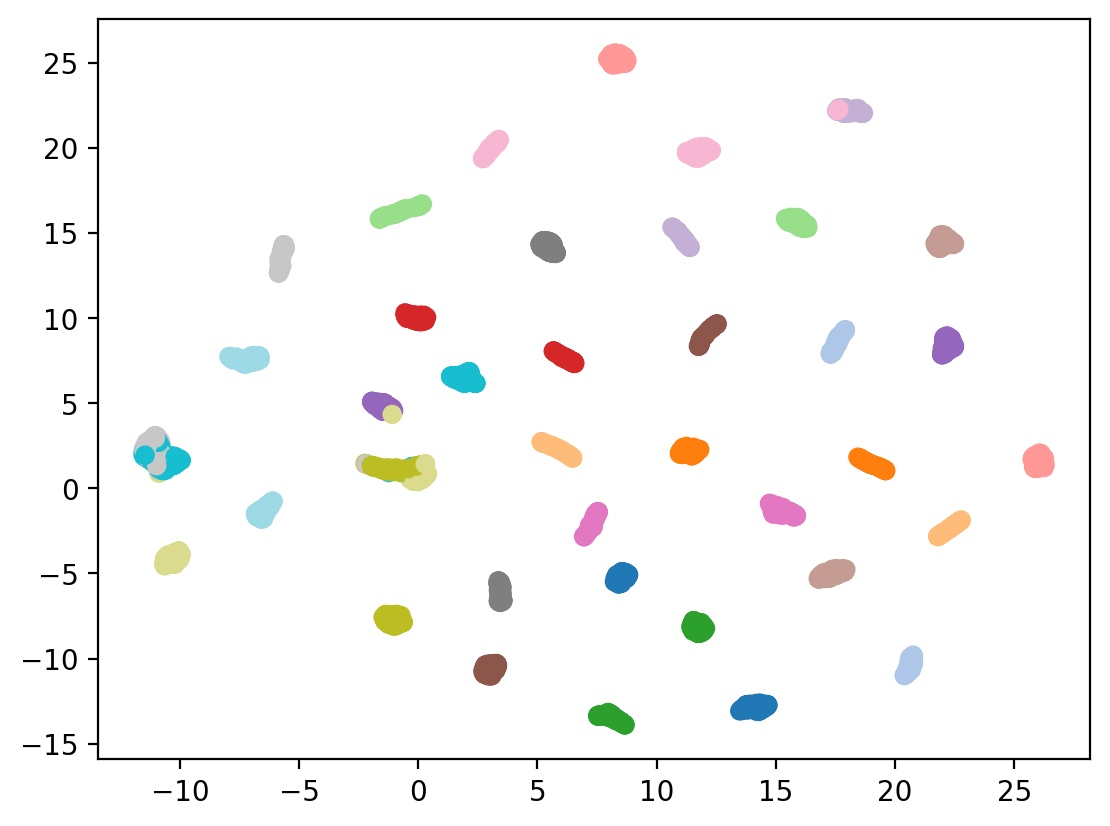}
    \caption{$F_{StyleShot}$ with DEC}
  \end{subfigure}
   \caption{UMAPs for different features with K-Means (top row) and DEC (bottom row) clustering algorithms}
  \label{fig:umap}
\end{figure*}
\begin{figure*}[!t]
 \begin{subfigure}[t]{.33\textwidth}
    \centering
    \includegraphics[width=\textwidth]{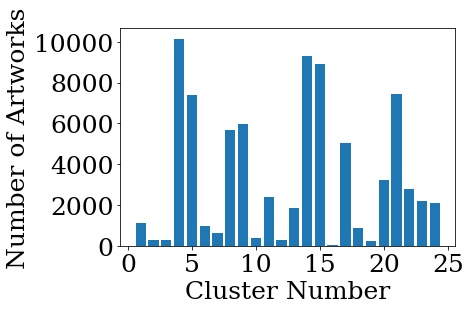}
    \caption{$F_{Dense}$ with WikiArt dataset}
  \end{subfigure}
   \hfill
 \begin{subfigure}[t]{0.33\textwidth}
    \centering
    \includegraphics[width=\textwidth]{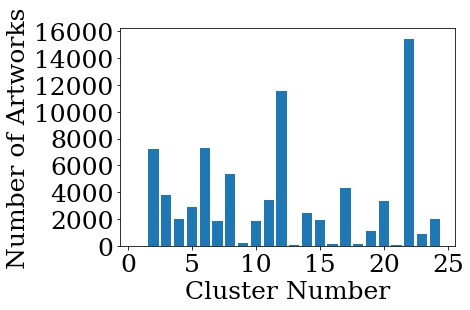}
    \caption {$F_{Gram}$ with WikiArt dataset}
\hfill
  \end{subfigure}
  \begin{subfigure}[t]{.33\textwidth}
    \centering
    \includegraphics[width=\textwidth]{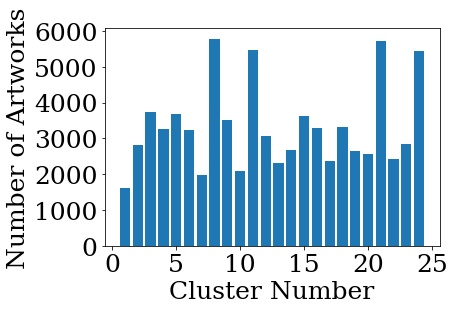}
    \caption{$F_{StyleCap}$ with WikiArt dataset}
  \end{subfigure}

 \begin{subfigure}[t]{.33\textwidth}
    \centering
    \includegraphics[width=\textwidth]{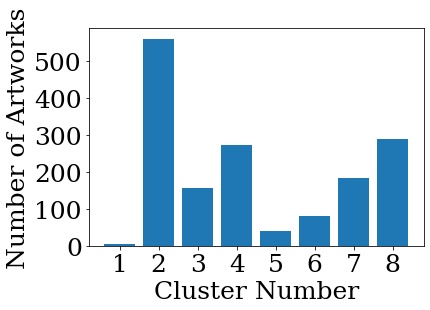}
    \caption{$F_{Dense}$ with Brueghel dataset}
  \end{subfigure}
   \hfill
 \begin{subfigure}[t]{0.33\textwidth}
    \centering
    \includegraphics[width=\textwidth]{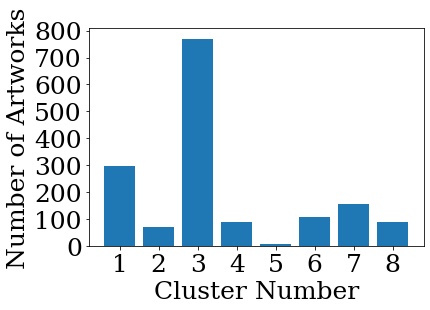}
    \caption {$F_{Gram}$ with Brueghel dataset}
\hfill
  \end{subfigure}
  \begin{subfigure}[t]{.33\textwidth}
    \centering
    \includegraphics[width=\textwidth]{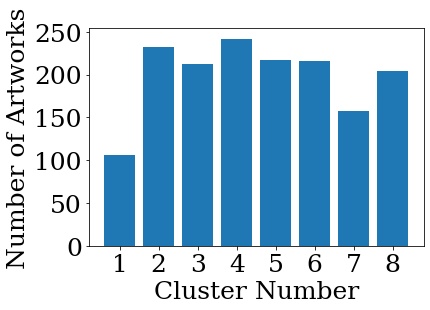}
    \caption{$F_{StyleCap}$ with Brueghel dataset}
  \end{subfigure}
   \caption{Distribution of artworks for different features for the WikiArt dataset (top row) and the Brueghel dataset (bottom row).}
  \label{fig:clust_hist}
\end{figure*}
    \clearpage
}


\end{document}